\documentclass[Afour,sageh,times]{sagej}

\usepackage{moreverb,url}
\usepackage[colorlinks,bookmarksopen,bookmarksnumbered,citecolor=red,urlcolor=red]{hyperref}
\usepackage{amsmath,amssymb,amsfonts}
\usepackage{graphicx}
\usepackage{gensymb}
\usepackage{textcomp}
\usepackage{xcolor}
\usepackage{placeins}
\usepackage{titlesec}
\usepackage{comment}
\usepackage{wrapfig}
\usepackage{placeins}
\usepackage[caption=false, font=footnotesize]{subfig}
\usepackage{multirow}
\usepackage{mathtools}%
\usepackage[nolist,nohyperlinks]{acronym}
\usepackage{colortbl}
\usepackage{pdfpages}
\usepackage{algorithm}
\usepackage{algpseudocode} %
\usepackage[skip=2pt,font=footnotesize]{caption}
\usepackage{adjustbox}

\usepackage{amsmath}

\DeclareMathOperator*{\argmin}{arg\,min}
\usepackage{bm}

\usepackage[capitalise,noabbrev]{cleveref} %

\crefname{algorithm}{algorithm}{algorithms} %
\Crefname{algorithm}{Algorithm}{Algorithms} %

\makeatletter
\let\MYcaption\@makecaption
\makeatother
\usepackage{subcaption}
\makeatletter
\let\@makecaption\MYcaption
\makeatother

\usepackage[T1]{fontenc}
\usepackage{siunitx}
\DeclareSIUnit{\sqrthertz}{\sqrt{\unit{\hertz}}}
\DeclareSIUnit\points{points}

\usepackage{booktabs}
\usepackage{multirow}
\usepackage{makecell}
\usepackage{threeparttable}

\usepackage[nolist,nohyperlinks]{acronym}
\usepackage{fontawesome} %
\usepackage[utf8]{inputenc}

\newcommand{\mat}[1]{\mathbf{#1}}
\newcommand{\rot}[3]{\mat{#1}_\mathtt{#2}^\mathtt{#3}}

\newcommand{\vect}[3]{\bm{#1}_\mathtt{#2}^\mathtt{#3}}

\newcommand{\coord}[1]{$\{\mathtt{#1}\}$}

\DeclareMathOperator*{\Log}{Log}

\usepackage{tikz}
\newcommand{\tikzxmark}[1][red]{%
\tikz[scale=0.23] {
    \draw[color=#1, line width=0.7,line cap=round] (0,0) to [bend left=6] (1,1);
    \draw[color=#1, line width=0.7,line cap=round] (0.2,0.95) to [bend right=3] (0.8,0.05);
}}
\newcommand{\tikzcmark}[1][black]{%
\tikz[scale=0.23] {
    \draw[color=#1, line width=0.7,line cap=round] (0.25,0) to [bend left=10] (1,1);
    \draw[color=#1, line width=0.8,line cap=round] (0,0.35) to [bend right=1] (0.23,0);
}}

\newcommand{\sensorset}{\pazocal{S}}
\newcommand{\robotconstraints}{\mu_R}

\newcommand{\depthsensor}{\pazocal{D}}
\newcommand{\camerasensor}{\pazocal{C}}
\newcommand{\graph}{\mathbb{G}}
\newcommand{\planningpath}{\sigma}
\newcommand{\viewpointset}{\mathbb{V}}
\newcommand{\missionobj}{\mathcal{J}}

\DeclareMathAlphabet{\pazocal}{OMS}{zplm}{m}{n}

\newcommand{\Es}{\pazocal{E}}

\newcommand{\Vs}{\pazocal{V}}

\newcommand{\Ms}{\pazocal{M}}

\newcommand{\Hs}{\pazocal{H}}

\newcommand{\pbf}{\bm{p}}

\newcommand{\hide}[1]{}

\newcommand{\bdmath}{\begin{dmath}}
\newcommand{\edmath}{\end{dmath}}
\newcommand{\beq}{\begin{equation}}
\newcommand{\eeq}{\end{equation}}
\newcommand{\bdm}{\begin{displaymath}}
\newcommand{\edm}{\end{displaymath}}
\newcommand{\bea}{\begin{eqnarray}}
\newcommand{\eea}{\end{eqnarray}}
\newcommand{\beal}{\beq \begin{array}{ll}}
\newcommand{\eeal}{\end{array} \eeq}
\newcommand{\beas}{\begin{eqnarray*}}
\newcommand{\eeas}{\end{eqnarray*}}
\newcommand{\ba}{\begin{array}}
\newcommand{\ea}{\end{array}}
\newcommand{\bit}{\begin{itemize}}
\newcommand{\eit}{\end{itemize}}
\newcommand{\ben}{\begin{enumerate}}
\newcommand{\een}{\end{enumerate}}

\acrodef{ua}[\texttt{UAstack}]{Unified Autonomy stack} %
\acrodef{mpc}[NMPC]{Model Predictive Control}
\acrodef{neuralmpc}[SDF-NMPC]{Neural NMPC through Signed Distance Field Encoding for Collision Avoidance}
\acrodef{cbf}[CBF]{Control Barrier Function}
\acrodef{ccbf}[C-CBF]{Composite Control Barrier Function}
\acrodef{rcbf}[R-CBF]{Robust Control Barrier Function}
\acrodef{ema}[EMA]{Exponential Moving Average}
\acrodef{sdf}[SDF]{Signed Distance Function}
\acrodef{mlp}[MLP]{Multi-Layer Perceptron}
\acrodef{imu}[IMU]{Inertial Measurement Unit}
\acrodef{fmcw}[FMCW]{Frequency-Modulated Continuous Wave}
\acrodef{slam}[SLAM]{Simultaneous Localization And Mapping}
\acrodef{drl}[DRL]{Deep Reinforcement Learning}
\acrodef{dce}[DCE]{Deep Collision Encoder}
\acrodef{exteroceptivedrl}[ExRL]{Exteroceptive DRL}
\acrodef{lm}[LM]{Levenberg–Marquardt}

\acrodef{fov}[FoV]{Field of View}
\acrodef{tof}[ToF]{Time of Flight}
\acrodef{vtol}[VTOL]{Vertical Take-Off and Landing}
\acrodef{ros}[ROS]{Robot Operating System}
\acrodef{lan}[LAN]{Local Area Network}
\acrodef{ptp}[PTP]{Precision Time Protocol}
\acrodef{ntnu}[NTNU]{Norwegian University of Science and Technology}
\acrodef{oasis}[OASIS]{Omnidirectional Autonomy Sensor Integration System}
\acrodef{ve}[VE]{Volumetric Exploration}
\acrodef{gvi}[GVI]{General Visual Inspection}
\acrodef{ei}[EI]{Exploration and Inspection}
\acrodef{tsp}[TSP]{Traveling Salesman Problem}
\acrodef{uart}[UART]{Universal Asynchronous Receiver/Transmitter}
\acrodef{swap}[SWaP]{Size, Weight and Power}
\acrodef{dof}[DoF]{Degrees of Freedom}
\acrodef{fmcw}[FMCW]{Frequency Modulated Continuous Wave}
\acrodef{ugv}[UGV]{Uncrewed Ground Vehicles}
\acrodef{uav}[UAV]{Uncrewed Aerial Vehicles}
\acrodef{sota}[SOTA]{State-of-the-Art}
\acrodef{mipi}[MIPI]{Mobile Industry Processor Interface}
\acrodef{csi2}[CSI-2]{Camera Serial Interface 2}
\acrodef{rmse}[RMSE]{Root Mean Square Error}
\acrodef{ape}[APE]{Absolute Pose Error}
\acrodef{rpe}[RPE]{Relative Pose Error}
\acrodef{ate}[ATE]{Absolute Trajectory Error}
\acrodef{rte}[RTE\textsubscript{10}]{Relative Trajectory Error}
\acrodef{ransac}[RANSAC]{RANdom SAmple Consensus}

\acrodef{gbp3}[OmniPlanner]{OmniPlanner}
\acrodef{ve}[VE]{Volumetric Exploration}
\acrodef{vi}[VI]{Visual Inspection}

\acrodef{qp}[QP]{Quadratic Programming}

\acrodef{gnss}[GNSS]{Global Navigation Satellite System}
\acrodef{vlm}[VLM]{Vision-Language Model}
\acrodef{vla}[VLA]{Vision-Language-Action}
\acrodef{llm}[LLM]{Large Language Model}
\acrodef{ttc}[TTC]{Time-to-Collision}
\acrodef{cnn}[CNN]{Convolutional Neural Network}
\acrodef{gru}[GRU]{Gated Recurrent Unit}
\acrodef{ppo}[PPO]{Proximal Policy Optimization}
\acrodef{elu}[ELU]{Exponential Linear Unit}

\acrodef{snr}[SNR]{Signal-to-Noise Ratio}

\acrodef{ar1}[AR-1]{Aerial Robot 1}
\acrodef{ar2}[AR-2]{Aerial Robot 2}
\acrodef{gr1}[GR-1]{Ground Robot 1}

\newcommand{\perception}{\texttt{perception module}}
\newcommand{\planning}{\texttt{planning module}}
\newcommand{\navigation}{\texttt{navigation module}}

\newcommand\BibTeX{{\rmfamily B\kern-.05em \textsc{i\kern-.025em b}\kern-.08em
T\kern-.1667em\lower.7ex\hbox{E}\kern-.125emX}}

\def\volumeyear{2026}

\begin{document}

\runninghead{Dharmadhikari et al.}

\title{The Unified Autonomy Stack:\\Toward a Blueprint for Generalizable Robot Autonomy}

\author{Mihir Dharmadhikari*, Nikhil Khedekar*, Mihir Kulkarni*, Morten Nissov*, Martin Jacquet, Angelos Zacharia, Marvin Harms, Albert Gassol Puigjaner, Philipp Weiss, Kostas Alexis}

\affiliation{All authors are affiliated with the Autonomous Robots Lab, \ac{ntnu}, Trondheim, Norway}

\corrauth{Kostas Alexis, Department of Engineering Cybernetics, Norwegian University of Science and Technology, H\o{}gskoleringen 1, Trondheim 7034, Norway. * indicates equal contribution.}

\email{konstantinos.alexis@ntnu.no}

\begin{abstract}
We introduce and open-source the \texttt{Unified Autonomy Stack}, a system-level solution that enables resilient autonomy across diverse aerial and ground robot morphologies. The architecture centers on three synergistic modules --multi-modal perception, multi-behavior planning, and multi-layered safe navigation-- that together deliver comprehensive mission autonomy. The stack fuses data from LiDAR, radar, vision, and inertial sensing, enabling (a) robust localization and mapping through factor graph-based fusion, (b) semantic scene understanding, (c) motion and informative path planning through sampling-based techniques adaptive across spatial scales, as well as (d) multi-layered safe navigation both through planning on the online reconstructed map and deep learning-driven exteroceptive policies alongside last-resort safety filters using control barrier functions. The resulting behaviors include safe GNSS-denied navigation into unknown and perceptually-degraded regions, exploration of complex environments, object discovery, and efficient inspection planning. The stack has been field-tested and validated on both aerial (rotorcraft) and ground (legged) robots operating in a host of demanding environments, including self-similar and smoke-filled settings, with complex geometries and high obstacle clutter. These tests demonstrate resilient performance in challenging conditions. To facilitate ease of adoption, we open-source the implementation alongside supporting documentation, validation, and evaluation datasets \url{https://github.com/ntnu-arl/unified_autonomy_stack}. \textcolor{black}{A video giving the overview of the paper and the field experiments is available at \url{https://youtu.be/l8Su8OXsM-E}.}

\end{abstract}


\maketitle

\section{Introduction}

Mobile robots are increasingly deployed to operate in environments where \ac{gnss} is unavailable, perception is degraded, geometry is self-similar, and safe navigation is broadly challenged~\cite{cerberus_science,harlow2024newwave,extremeslam_tro,datar2025m2p2,chung2023into}. Although the robotics community has developed mature components for many individual functionalities such as localization, planning and control, existing autonomy stacks (e.g., the works in~\cite{fernandez2023aerostack2,sanchez2016aerostack,baca2021mrs,mohta2018fast,foehn2022agilicious,goodin2024nature,airstack2025,real_ijars20}) remain largely specialized to a particular robot morphology, sensor suite, or mission class. This specialization limits reuse across platforms, makes systematic field evaluation difficult, slows the accumulation of shared deployment experience, and hinders both consolidation and broader uptake of autonomy across robot categories.

At the same time, limited attention has been given to feature-rich autonomy stacks readily delivering complex behaviors and functionalities such as resilient \ac{gnss}-denied navigation in perceptually-degraded environments, combined with sophisticated informative path planning and assured safety ensuring robust operation across operational environments and conditions. However, recent advances across the ``sense-think-act'' loop -from perception to planning and deep control policies- point toward the potential for a resilient and, to a significant extent, unified autonomy engine. Although research on universal autonomy is still in its early stages, the benefits of unification and the collective need to advance robot capabilities underscore the need for general autonomy solutions.

\begin{figure*}
	\centering
	\includegraphics[width=0.99\linewidth]{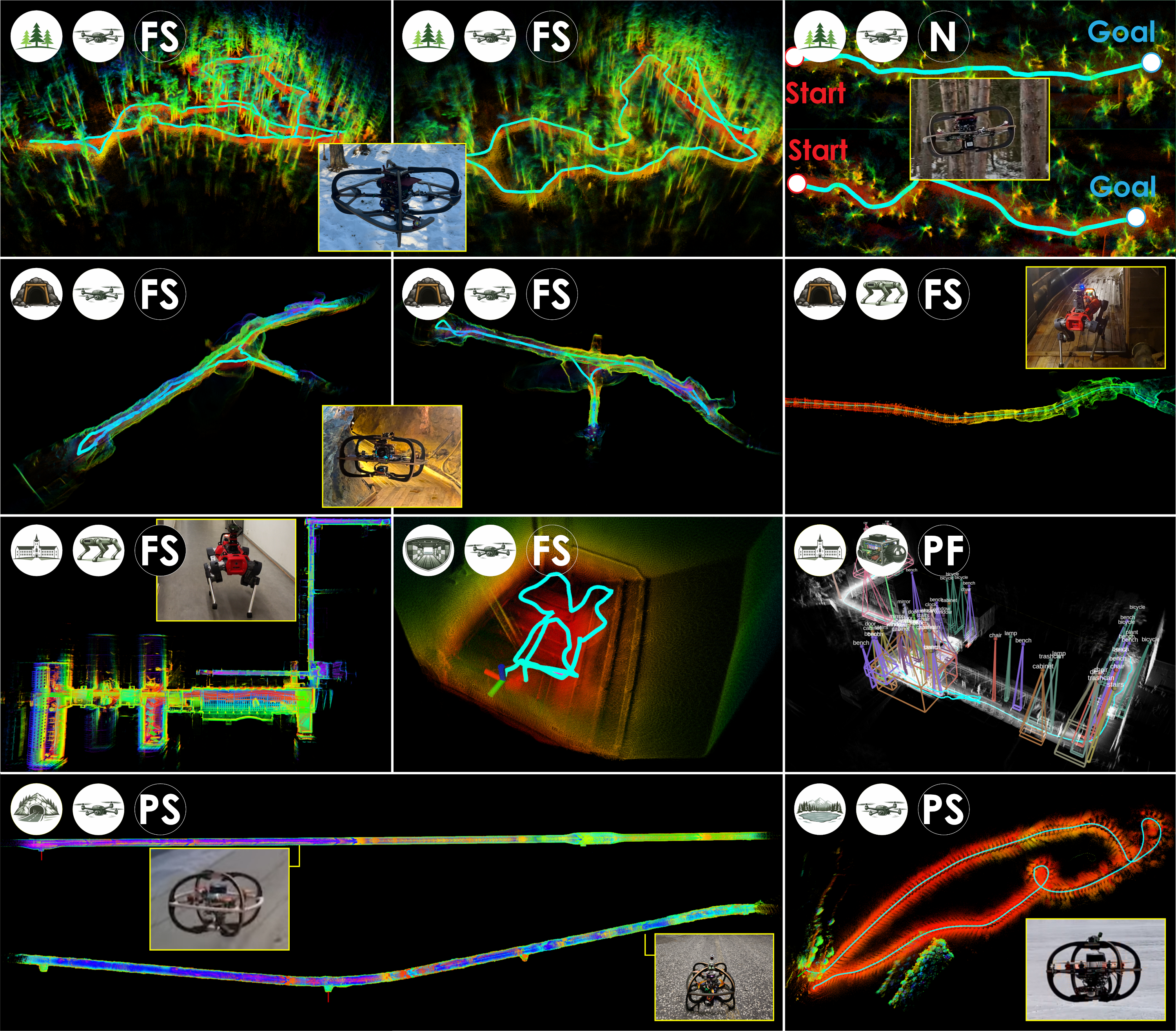}
	\caption{Indicative subset of the evaluation studies conducted to validate and assess the performance of the Unified Autonomy Stack. The tests involve both aerial and ground robots operating in diverse GNSS-denied and at instances perceptually-degraded environments including (a) snow-covered forests, (b) underground mines, (c) road tunnels, (d) a frozen lake, (e) ship cargo holds, as well as (f) the university campus with subsets of it filled with fog. All core modules of the autonomy stack were evaluated independently, alongside full-stack experiments. In this image ``FS'' stands for Full-Stack deployments (involving the synergy of the \perception{}, \planning{}, and \navigation{}), ``N'' for deployments to evaluate the \navigation{}, ``PF'' for testing both the localization and mapping as well as object-level reasoning of the \perception{}, while ``PS'' marks tests focusing only on localization and mapping. The presented instances are from a subset of the experiments as detailed in the paper. The full datasets of the experiments are also released to support verifiability. }
	\label{fig:introfig}
\end{figure*}

Motivated by the above, we present the \ac{ua}, a comprehensive open-source autonomy stack that can support mission-level operation across diverse aerial and ground robot configurations. The \ac{ua} represents a step towards a common autonomy blueprint across diverse robot types -from multirotors and other rotorcrafts to ground robots such as legged systems- delivering mission-complete capabilities for navigation and complex information sampling behaviors (e.g., exploration, inspection, object discovery) in diverse settings, including in strenuous, high-risk natural and industrial environments. Its design emphasizes resilience in that it presents robustness (e.g., against noisy sensor data and mapping imperfections), resourcefulness (e.g., multiple solutions for safety in navigation), and redundancy (e.g., complementary sensor data to handle single-modality failures), enabling it to retain high performance across environments and conditions, including \ac{gnss}-denied, perceptually-degraded, geometrically complex, and potentially adversarial settings that typically challenge safe navigation and mission autonomy.

The \ac{ua} builds upon three core modules and associated contributions. The \perception{} is centered around a novel approach to multi-modal \ac{slam} based on factor graphs, enabling robust fusion of LiDAR, \ac{fmcw} radar, visual perception, and \ac{imu} cues. This supports resilient performance in \ac{gnss}-denied environments with multiple perceptual degradations, including geometric self-similarity, low texture, icy scenes, and dense obscurants (e.g., fog, smoke). Furthermore, the \perception{} integrates scene reasoning through \acp{vlm} enabling object discovery and visual question \& answering (Q\&A). The \planning{} builds upon OmniPlanner~\cite{zacharia2026omniplannerarxiv}, facilitating target reaching, exploration, and inspection path planning through sampling-based methods over an online-derived volumetric map of the environment. It provides a versatile framework that abstracts vehicle configuration and supports diverse mission objectives, which the planner optimizes accordingly. The \navigation{} builds upon the contributions in~\cite{navigation_neuralmpc,navigation_ccbf}, alongside introducing a novel approach to exteroceptive reinforcement learning for navigation. Recognizing that localization errors and mapping imperfections may arise, we adopt a redundant, multi-layered safety approach in which depth-based exteroceptive navigation policies and last-resort control barrier function-based safety filters enhance safety by providing direct and reactive collision avoidance. As a result, the \ac{ua} is tailored to environments that challenge localization and mapping, both by leveraging sensor multi-modality to maintain perception under degraded conditions, and by implementing reactive safety mechanisms to reduce reliance on perfect scene understanding.

To evaluate the performance and assess its resilience, the \ac{ua} is evaluated onboard multiple robot configurations and within a diverse set of environments, an indicative subset of which is shown in Figure~\ref{fig:introfig}. 
First, a detailed quantitative evaluation of the \perception{} is presented, covering $2$ urban tunnels, a frozen lake and a university campus environment characterized by geometric self-similarity, low visibility, and heavy airborne obscurants.
It displays the superior performance of the \perception{} compared to \acl{sota} LiDAR-Inertial, LiDAR-Radar-Inertial, LiDAR-Visual-Inertial, or Visual-Inertial \acs{slam} methods. Object-level reasoning is assessed by building 3D scene graphs with object-level annotations, alongside enabling visual Q\&A on online camera data.
Next, the \navigation{} is evaluated in two real-world deployments. First, in a forest environment for a waypoint-navigation task requiring maneuvering to avoid trees, while map-based path planning is disabled to isolate the reactive layer. Second, we evaluate safety under map discrepancies by introducing previously unmapped obstacles along the path planned by the \planning{} and demonstrate that the \navigation{} layer can handle such unseen obstacles.
These experiments highlight the importance of the multi-layered safety approach and the complementary roles of each safety method.
Finally, the full \ac{ua} is evaluated on aerial and legged robots performing autonomous exploration and inspection missions guided by the \planning. The aerial robot is deployed in a) a low-visibility, multi-branched underground mine, and b) a forest with thin obstacles and local clutter, performing exploration missions, with \navigation{} modalities being tested. Additionally, the aerial robot is deployed in the cargo hold of a ship performing an exploration and inspection mission. On the other hand, the legged robot is deployed in a university campus, and inside the same underground mine as the aerial robot. This demonstrates large-scale missions across heterogeneous platforms and environments.

Importantly, the Unified Autonomy Stack is open to extension, both from the perspective of the robots it readily supports and the missions it enables. Our team is targeting its full-fledged expansion to a diverse set of robot configurations and the extension of the enabled behaviors. To that end, the full implementation is open-sourced (\url{https://github.com/ntnu-arl/unified_autonomy_stack}), alongside documentation and supporting datasets (\url{https://ntnu-arl.github.io/unified_autonomy_stack/}). 

The remainder of this paper is structured as follows. Section~\ref{sec:related_work} overviews related work in autonomy stacks. The \ac{ua} and its modules are detailed in Section~\ref{sec:approach}. Evaluation studies are presented in Section~\ref{sec:evaluation}, followed by conclusions and plans for future work in Section~\ref{sec:conclusion}.

\section{Related Work}
\label{sec:related_work}

\begin{table*}[h]
  \centering
  \caption{Comparison with existing autonomy stacks.}
  \label{tab:related_work:stack_comparison}
  \newcommand{\rotation}{0}
\newcommand{\width}{10pt}
\newcommand{\vmove}{-1.35ex}%
\newcommand{\dash}{---}
\small
\begin{threeparttable}
\begin{tabular}{l@{\hspace{\width}}c@{\hspace{\width}}c@{\hspace{\width}}c@{\hspace{\width}}ccccccccccc}
  \toprule
    \multirow{3}{*}[\vmove]{Stack}  &\multirow{3}{*}[\vmove]{\rotatebox{\rotation}{Modular}}  &\multirow{3}{*}[\vmove]{\rotatebox{\rotation}{Modalities\tnote{1}}} &\multirow{3}{*}[\vmove]{\rotatebox{\rotation}{Embodiments\tnote{2}}} &\multirow{3}{*}[\vmove]{\rotatebox{\rotation}{Verified\tnote{3}}} &\multicolumn{10}{c}{Autonomy Features\tnote{4}} \\ \cmidrule(l){6-15}
     & & & & & \multicolumn{3}{c}{Perception} & \multicolumn{3}{c}{Planning} & \multicolumn{4}{c}{Safety} \\ \cmidrule(lr){6-8} \cmidrule(lr){9-11} \cmidrule(l){12-15}
     & & & & & SLAM & OD & LQ & EP & IP & TP & M & T & Ct & LR \\
    \midrule
    AeroStack      & \tikzcmark & \underline{C}     & A  & SL    & \tikzcmark & \tikzxmark[red] & \tikzxmark[red] & \tikzxmark[red] & \tikzxmark[red] & \tikzcmark & \tikzcmark & \dash        & \tikzxmark[red] & \tikzxmark[red] \\
    AeroStack2     & \tikzcmark & C     & A  & SL    & \tikzxmark[red] & \tikzxmark[red] & \tikzxmark[red] & \tikzxmark[red] & \tikzxmark[red] & \tikzcmark & \tikzcmark & \dash        & \tikzxmark[red] & \tikzxmark[red] \\
    MRS UAV        & \tikzcmark & \underline{C} \underline{L} \underline{G} & A  & SLF   & E                   & \tikzxmark[red] & \tikzxmark[red] & \tikzxmark[red] & \tikzxmark[red] & \tikzcmark & \tikzcmark & \dash        & \tikzxmark[red] & \tikzcmark \\
    Nebula         & \tikzcmark & C\underline{L}    & AG & SLF   & \tikzcmark & \dag & \tikzxmark[red] & \dag & \tikzxmark[red] & \dag & \dag & \tikzcmark & \tikzxmark[red] & \tikzxmark[red] \\
    Agilicious     & \tikzcmark & \underline{C}     & A  &SLF   & \tikzxmark[red] & \tikzxmark[red] & \tikzxmark[red] & \tikzxmark[red] & \tikzxmark[red] & \tikzcmark & \tikzxmark & \dash        & \tikzxmark[red] & \tikzxmark[red] \\
    KR Aut. Flight & \tikzcmark & \underline{CL}    & A  & SLF   & \tikzcmark & \tikzxmark[red] & \tikzxmark[red] & \tikzxmark[red] & \tikzxmark[red] & \tikzcmark & \tikzcmark & \dash        & \tikzxmark[red] & \tikzxmark[red] \\
    NATURE         & \tikzxmark & L     & G  & SLF & \tikzxmark[red] & \tikzxmark[red] & \tikzxmark[red] & \tikzxmark[red] & \tikzxmark[red] & \tikzcmark & \tikzcmark & \tikzcmark & \tikzxmark[red] & \tikzxmark[red] \\
    AirStack       & \tikzcmark & \underline{C}L    & A  & SL    & \tikzcmark & \dag & \tikzxmark[red] & \tikzcmark & \tikzxmark[red] & \tikzcmark & \dag & \dash        & \dag & \tikzxmark[red] \\
    Nav2           & \tikzcmark & \underline{L}     & G  & SLF   & \tikzcmark & \tikzxmark[red] & \tikzxmark[red] & \tikzxmark[red] & \tikzxmark[red] & \tikzcmark & \tikzcmark & \tikzcmark & \tikzxmark[red] & \tikzcmark \\

     Autoware           & \tikzcmark & CR\underline{LG}   & G  & SLF   & \tikzcmark & \tikzcmark & \tikzcmark & \tikzxmark[red] & \tikzxmark[red] & \tikzcmark & \tikzcmark & \tikzcmark & \tikzxmark[red] & \tikzcmark \\

      Apollo           & \tikzcmark & CR\underline{LG}     & G  & SLF   & \tikzcmark & \tikzcmark & \tikzxmark[red] & \tikzxmark[red] & \tikzxmark[red] & \tikzcmark & \tikzcmark & \tikzcmark & \tikzxmark[red] & \tikzcmark \\

    \midrule
    Ours (\ac{ua})           & \tikzcmark & \underline{CRL}   & AG & SLF   & \tikzcmark & \tikzcmark & \tikzcmark & \tikzcmark & \tikzcmark & \tikzcmark & \tikzcmark & \tikzcmark & \tikzcmark & \tikzcmark \\
    \bottomrule

\end{tabular}
\begin{tablenotes}
    \item[1] Supported perception modalities, where C: camera, R: radar, L: LiDAR, G: \ac{gnss}. The underline denotes modalities which are used by \ac{slam}, and breaks indicate non-multi-modal \ac{slam} usage.
    \item[2] Supported embodiments, where A: aerial, G: ground. %
    \item[3] How the autonomy stack was tested, where S: simulation, L: lab, F: field 
    \item[4] Presence of feature, where: \tikzcmark: feature exists and open-sourced, \dag: exists and not open-sourced, \tikzxmark[red]: does not exist, E: external package integrated as is, \dash: not applicable
    \begin{tabular}{l l l}
        OD: Object Detection        & LQ: Language Query            & EP: Exploration Planning \\
        IP: Inspection Planning     & TP: Planning to Target        & M: Map-based safety \\
        T: Traversability checking  & Ct: Safety at control layer & LR: Last resort safety
    \end{tabular}

\end{tablenotes}
\end{threeparttable}

\end{table*}

Research in autonomy is a particularly active field. The robotics community has increasingly released reusable autonomy components, while more complete autonomy stacks remain less common. Driven by the major success of open-source autopilot releases, such as PX4 and ArduPilot~\cite{meier2015px4}, these developments aim to accelerate research on robot autonomy and democratize its adoption through open implementations and standards. In this section, we review relevant literature and position how the \ac{ua} contributes to this landscape.

Typically, major building blocks of autonomy are often documented and released as separate publications and open-source contributions. Indicative examples span \ac{slam}~\cite{xu2022fastlio2,vizzo2023ral,koide2024glim,campos2021orbslam3,Geneva2020ICRA,qin2019generaloptimizationbasedframeworklocal}, motion and path planning~\cite{sucan2012open,wang2022geometrically,mellinger2011minimum,likhachev2003ara}, control~\cite{verschueren2022acados,andersson2018casadi,giftthaler2018control}, robot learning~\cite{petrenko_sample_2020,denys2021rlgames,schwarke2025rslrl,nvidia2025isaaclab,mujoco_playground_2025}, and navigation frameworks~\cite{macenski2020marathon2,ros1navigation}. While these works provide an important foundation for robot autonomy, they are not the focus of this section. Instead, this section assesses and compares existing integrated, open-source autonomy stacks for mobile robots, including aerial and ground systems.

Autonomous driving is among the most mature fields in terms of integrated autonomy stacks. The Autoware Universe~\cite{autoware_universe,kato2018autoware} is a modular ROS2 autonomous driving stack that involves a highly-integrated perception, planning, and control solution. Apollo~\cite{apollo} is another flagship open autonomous driving project, targeting autonomy in structured urban streets. A comparative analysis of major autonomous driving stacks is provided in~\cite{jung2025open}, reflecting the broader community interest in open autonomy frameworks. Off-road autonomy is addressed in NATURE~\cite{goodin2024nature}, which introduces an open-source stack, providing a self-contained pipeline with perception, global and local planning, and control components, with support for both ROS1 and ROS2.

Beyond autonomous driving, aerial robotics has seen some of the most active development of autonomy stacks. The MRS UAV System~\cite{baca2021mrs} is among the most complete open-source autonomy stacks for multirotor aerial robots. It supports LiDAR-based \ac{slam} state estimation, several control methods including SE(3) formulations and model predictive control, trajectory generation, and selected multi-robot capabilities. From a complementary perspective, Aerostack2~\cite{fernandez2023aerostack2} is a ROS2-based autonomy framework for aerial robots that emphasizes standardization, modularity, and reusability. It provides replaceable and dynamically configurable components, enables behavior-based mission planning, and supports multi-robot coordination. Unlike more integrated systems, it does not provide a comprehensive perception-planning-control pipeline and does not include some components required for autonomy (e.g., a \ac{slam} solution). Aerostack2 builds upon Aerostack~\cite{sanchez2016aerostack}, which was an early effort to standardize aerial robot frameworks. Going further, Aerostack2 provides better modularity, a wider set of controller/estimator options, a richer behavior tree, and is implemented in ROS 2. AirStack~\cite{airstack2025} targets modular aerial autonomy and provides a research-to-deployment pipeline with tight support for testing in both simulation and reality. Released as an alpha version at the time of writing, it is not yet fully available; instead, only a subset of its modules is open at the time of writing (e.g., exploration planning, visual-inertial odometry). Following a different direction, the unmanned aerial vehicle abstraction layer (UAL)~\cite{real_ijars20} contributes a layered design providing a standard API for aerial robot control and abstracts specific autopilot implementations. Through a namespaced design, it facilitates multi-robot support, but does not provide higher-level autonomy components. Agilicious~\cite{foehn2022agilicious} differs from these systems by offering a hardware-software stack for agile vision-based quadrotor flight, with an emphasis on tight integration of the perception-action loop. Compared with systems such as the MRS UAV System and Aerostack, it follows a performance-first philosophy, resulting in a tightly coupled but necessarily less modular solution.
The kr\_autonomous\_flight stack~\cite{mohta2018fast} also provides a complete autonomy stack for \ac{gnss}-denied aerial robot missions, integrating vision-based odometry, LiDAR-based \ac{slam}, global planning for target reaching, trajectory generation, while also supporting downstream tasks such as semantic \ac{slam}.
Beyond aerial robotics, Nav2~\cite{macenski2020marathon2} provides a navigation framework for ground robots in predominantly indoor environments. Built around behavior trees, it includes LiDAR-based \ac{slam}, global and local planning, and control, offering a ROS2-compatible navigation stack for ground robots.

Targeting a more unified approach to autonomy, Nebula~\cite{githubNeBulaAutonomy,agha2021nebula} is team CoSTAR's architecture for their participation in the DARPA Subterranean Challenge. The public release includes components such as the multi-robot \ac{slam} method~\cite{chang2022lamp}, while other elements of the perception-action loop and the complete autonomy stack are not open. Beyond autonomy stacks themselves, recent work has also explored how such systems can support higher-level autonomy. For example,~\cite{ravichandran2025deploying} present progress in \ac{llm}-enabled autonomy in field settings with the \ac{llm} orchestrating autonomous operations over several kilometers across aerial and ground systems. This direction aligns with ongoing efforts on \ac{vla} models for autonomy~\cite{zhang2026uav,serpiva2025racevla,xu2026aerialvla,wu2025vla}, building on core modules such as works in \ac{slam} and control. 

Table~\ref{tab:related_work:stack_comparison} summarizes how various autonomy stacks compare in terms of functionalities and features. Specifically, we compare across four aspects: (a) supported sensing modalities and multi-layered safety mechanisms; (b) breadth of capabilities, including supported mission objectives; (c) design modularity and support for different robot embodiments; and (d) open-source availability and field evaluation. 

Compared with existing works, the \ac{ua} presents distinct contributions. First, it represents a functionality-rich autonomy stack experimentally validated on heterogeneous platforms, including multirotors and legged robots. It provides complete implementations of all the core modules (in perception, planning, and navigation) and the overarching architecture for high-performance autonomy. This includes \ac{slam}, \ac{vlm}-based scene reasoning, motion, and informative path planning, and multi-layered safe navigation and control. Second, the \ac{ua} incorporates design choices aimed at enhanced resilience. Aiming to enable resilient operations in strenuous environments, the \ac{ua} builds upon lessons learned from prior works in field autonomy such as~\cite{cerberus_science,extremeslam_tro,agha2021nebula,kottege2024heterogeneous,dharmadhikari2025semantics} and combines multi-modal sensor fusion (combining the complementary benefits of LiDAR, vision, and radar in order to penetrate through \ac{gnss}-denied and perceptually-degraded environments) with multi-layered safety (such that collision-avoidance is assured through three distinct but synergetic methodological pathways). Targeting advanced behaviors, the \ac{ua} supports advanced mission-level behaviors, offering exploration and coverage in complex, large-scale, geometrically complex environments. Finally, the complete implementation of the \ac{ua} is released open-source with supporting datasets and, where applicable, simulation integration, to accelerate testing, verification, adoption, and extension by the community.

\section{Unified Autonomy}
\label{sec:approach}

This section presents the architecture and key modules of the unified autonomy stack.

\subsection{Autonomy Architecture}

The Unified Autonomy Stack is organized around three core modules -- \texttt{perception}, \texttt{planning}, and \texttt{navigation} -- following the principles of the ``sense-think-act'' loop, while targeting generalizability across aerial and ground robot configurations, and resilience in demanding environments. Its overall architecture is presented in Figure~\ref{fig:architecture}. The \ac{ua} consumes diverse sensor data and outputs low-level commands to standard controllers available in most modern robotic systems, for example, on PX4-based drones~\cite{meier2015px4} (or any other MAVLink-compatible autopilot~\cite{koubaa2019micro}) and standard (linear and angular) velocity controllers on ground platforms. Its key features are as follows:

\begin{figure*}
	\centering
	\includegraphics[width=0.99\linewidth]{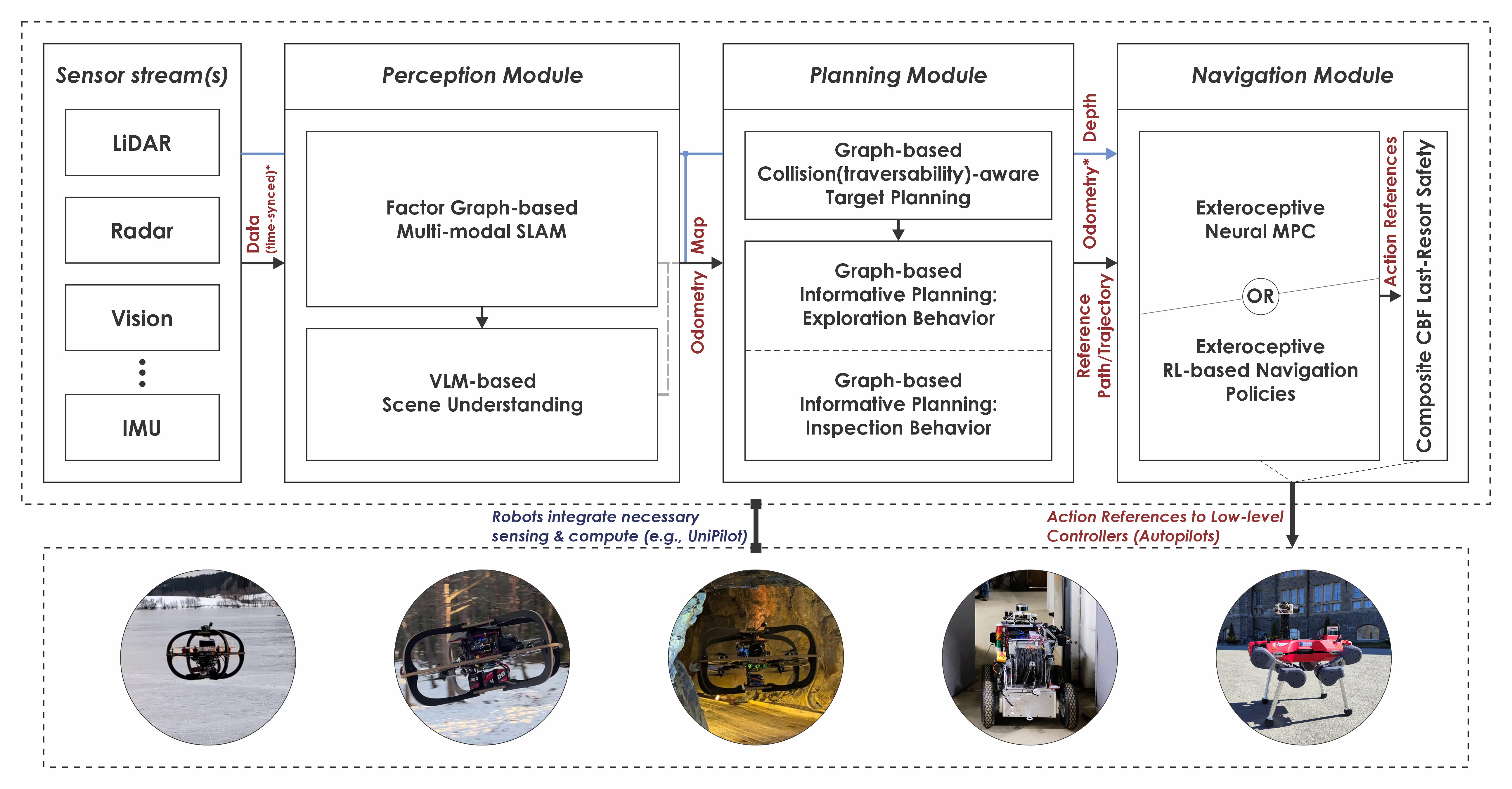}
	\caption{The architecture of the \ac{ua}. The stack involves three core modules, on \texttt{perception}, \texttt{planning}, and \texttt{navigation} that operate in a synergistic fashion. Aiming for operational resilience in diverse \ac{gnss}-denied, perceptually-degraded environments the \ac{ua} emphasizes multi-modal sensor fusion merging data from LiDAR, radar, and camera sensing, alongside \ac{imu} cues. \ac{vlm}-based reasoning builds upon the geometric reconstruction and supports object discovery and visual question/answering. The stack's planning layer offers diverse behaviors, with ready-made implementations for target-reach, unknown area exploration and inspection, across robot morphologies. Even though map-based collision avoidance and traversability analysis are provided within the \planning, the stack's \navigation{} further offers a multi-layered approach to safety involving deep exteroceptive navigation strategies, either through neural model predictive control or reinforcement learning, alongside formal last-resort safety based on control barrier functions. In terms of morphologies, the \ac{ua} currently supports aerial and ground robots, especially rotorcrafts (e.g., multirotors), legged robots and ground rovers. Experimental validation has taken place on different multirotor systems and quadruped legged robots, while simulation examples in the released code include additional morphologies such as ground rovers and helicopters.}
	\label{fig:architecture}
\end{figure*}

    \paragraph{Generalizability:} The \ac{ua} applies with few adjustments to a wide range of robot configurations, offering a consistent user experience for navigation and informative path planning tasks across platforms. Currently out-of-the-box supporting multirotors and other rotorcrafts, alongside several ground systems and especially legged robots, it provides a strong foundation for research in unified embodied AI. In this paper, we present experimental verification with multirotors and quadrupeds, while the associated open-source repository includes examples with additional morphologies such as ground rovers and helicopters. 

    \paragraph{Multi-modality:} The \ac{ua} fuses complementary sensor cues, currently including LiDAR, \ac{fmcw} radar, vision, and \ac{imu}, enabling resilience in perceptually-degraded and \ac{gnss}-denied conditions~\cite{chung2023into}, including settings characterized by self-similar geometries, dark or low-texture scenes, icy regions, and obscurants such as smoke and dust. We emphasize the tight fusion of LiDAR and radar data focusing on their complementary role in numerous perceptually-degraded environments as discussed in the evaluation section. 
    
    \paragraph{Multi-layer Safety:} The \ac{ua} departs from conventional architectures in which safety is ensured by solutions with a single point-of-failure. Most commonly, modern autonomy solutions rely on a cascade of calculations in which collision-free planning takes place only on an online reconstructed map. In practice, it entails that non-trivial localization or mapping errors (e.g., such as those often encountered in perceptually-degraded settings or when encountering thin obstacles) can lead to collisions. The \ac{ua} combines map-based motion planning with deep learning-driven navigation strategies and safety filters that directly consume online exteroceptive depth measurements and, if necessary, adjust the robot's path to re-assert safety. 
    
   \paragraph{Methodological Plurality:} The \ac{ua} integrates both ``conventional'' model-based control, estimation, perception, optimization and planning techniques, as well as deep learning-based methods alongside hybrid techniques. Indicative examples include its factor graph-based multi-modal \ac{slam} and its navigation policies offering options for \ac{drl}-based and Neural \ac{mpc}. 

Subsequently, we outline the key modules of the \ac{ua} and point to prior works as applicable. Furthermore, we discuss the interfaces considered and how \ac{ua} can be extended to new robot configurations.
Importantly, all autonomy modules described hereafter operate subject to the information provided to the stack's \texttt{Robot Abstraction Layer} and the \texttt{Mission Abstraction Layer}.

    \paragraph{\texttt{Robot Abstraction Layer}:} This layer defines the robot type (e.g., multirotor, quadruped, etc.) and its key motion parameters, alongside its sensor suite $\sensorset$ including the sensor types (e.g., LiDAR, Radar, cameras, \ac{imu}) and their configurations (e.g., fields of view, effective ranges, calibration constants). Based on this layer, the modules of the \ac{ua} determine the robot motion constraints $\robotconstraints$ to be applied (e.g., robot size, potentially applicable traversability constraints, kinematic constraints), the sensor data to be fused, sensor intrinsic and extrinsic calibration parameters, time-synchronization information between the sensors, as well as the command interface to be used (e.g., acceleration commands for multirotors, velocity commands to a quadruped legged robot) as detailed in Section~\ref{sec:low_level_interface}. It also sets if certain modules of the \ac{ua} are enabled such as if multi-layered safety will be employed and if yes, which submodules of it will be engaged.
    
    \paragraph{\texttt{Mission Abstraction Layer}:} This layer sets the task for the robot, including if this relates to reaching a target, exploring an unknown area, inspecting a previously explored region, any combination among those, or any other behavior built beyond the existing capabilities of the \ac{ua}. It accordingly defines the objective of the \planning{}. It also sets if certain modules are active such as the \ac{vlm}-based scene reasoning.

\subsection{Perception Module}

The \perception{} includes our solution for multi-modal \ac{slam}, alongside integration with a \ac{vlm}-based reasoning step.

\subsubsection{Multi-Modal \ac{slam}}\label{sec:slam}
The proposed novel multi-modal \ac{slam} system (dubbed MIMOSA-X, where 'X' denotes the modalities used) uses a factor graph estimator to fuse LiDAR, radar, camera, and \ac{imu} measurements using a windowed smoother~\cite{gtsam} for computational efficiency. This architecture, as shown in \cref{fig:architecture}, builds upon ideas proposed in \cite{khedekar_mimosa_2022,perception_dlrio}, with improvements drawn from further developments in \cite{perception_pglio,perception_jplRadar} for enhanced LiDAR and radar integration, alongside vision integration. Unlike loosely coupled approaches~\cite{khedekar_mimosa_2022, khattak_complementary_2020, liosam2020shan}, MIMOSA-X fuses LiDAR registration factors, radar Doppler factors, and preintegrated \ac{imu} factors in a tightly-coupled manner to avoid degenerate optimizations returning partially observable results.
Vision is further optionally fused through between factors.

The estimator considers a state space comprised of the position $\vect{p}{WB}{W}$, velocity $\vect{v}{WB}{W}$, and attitude $\rot{R}{B}{W}$ of the body \ac{imu} frame \coord{B} with respect to an inertial (map) frame \coord{W}. Note that \coord{W} is not perfectly gravity aligned, i.e., gravity in \coord{W} does not point exactly down. Furthermore, calibration states such as the accelerometer $\vect{b}{a}{}$ and gyroscope $\vect{b}{g}{}$ biases and gravity direction $\vect{g}{}{W}$ in the map frame are also included. The online gravity estimation has been shown to improve performance~\cite{nemiroff2023gravity}, as initial uncertainty regarding the platform attitude can result in map-errors which compound over large distances. The state space $\mat{x}$ is thus decomposed into local $\mat{x}_{\text{L}}$ and global $\mat{x}_{\text{G}}$ states such that

\begin{equation}
    \mat{x} = \begin{pmatrix} 
            \smash{\underbrace{\begin{matrix}
                \vect{p}{WB}{W} &\vect{v}{WB}{W} &\rot{R}{B}{W} &\vect{b}{a}{} &\vect{b}{g}{} 
            \end{matrix}}_{\mat{x}_{\text{L}}}} 
            &\smash{\underbrace{\begin{matrix} 
                \vect{g}{}{W} 
            \end{matrix}}_{\mat{x}_{\text{G}}}}
    \end{pmatrix}.
    \vphantom{\underbrace{\vect{p}{WB}{W}}_{\mat{x}_{\text{L}}}}
\end{equation}
By concatenating the states from times $t_{k-l}$ to $t_{k}$, the windowed set of states $\mathcal{X}_{k-l:k}$ is defined as

\begin{equation}
    \mathcal{X}_{k-l:k} = \begin{Bmatrix}
        \mat{x}_{\text{L},k-l} &\mat{x}_{\text{L},k-l+1} &\ldots &\mat{x}_{\text{L},k} &\mat{x}_{\text{G}}
    \end{Bmatrix}.
\end{equation}
The states over this temporal window of size $l+1$ are thus estimated by the iSAM2~\cite{kaess2011isam2} nonlinear optimizer, where the optimal estimate $\mathcal{X}_{k-l:k}^{*}$ is found by minimizing the covariance-weighted ($\Sigma_{\star}$) sum of the residuals $\bm{e}_{\star}$ derived from the \ac{imu}, LiDAR, radar, and vision sensor measurements included in the temporal window, denoted by $\mathcal{I}$, $\mathcal{L}$, $\mathcal{R}$, and $\mathcal{V}$, respectively. This minimization problem can thus be written as follows

\begin{equation}
\begin{aligned}
    \small
    \mathcal{X}_{k-l:k}^{*} = \underset{\mathcal{X}_{k-l:k}}{\arg\min} \Big[ 
        \lVert \bm{e}_{0} \rVert_{\Sigma_{0}}^{2}
        &+ \sum_{i\in\mathcal{F}_{k-l:k}^{\mathcal{I}}} \lVert \bm{e}_{\mathcal{I}_{i}} \rVert_{\Sigma_{\mathcal{I}_i}}^{2}\\
        + \sum_{i\in\mathcal{F}_{k-l:k}^{\mathcal{L}}} \lVert \bm{e}_{\mathcal{L}_{i}} \rVert_{\Sigma_{\mathcal{L}}}^{2}
        &+ \sum_{i\in\mathcal{F}_{k-l:k}^{\mathcal{R}}} \lVert \bm{e}_{\mathcal{R}_{i}} \rVert_{\Sigma_{\mathcal{R}}}^{2}\\
        &+ \sum_{i\in\mathcal{F}_{k-l:k}^{\mathcal{V}}} \lVert \bm{e}_{\mathcal{V}_{i}} \rVert_{\Sigma_{\mathcal{V}_i}}^{2}
    \Big],
\end{aligned}
\end{equation}
\normalsize
for the marginalization prior $\bm{e}_{0}$,~$\Sigma_{0}$, and the \ac{imu}, LiDAR, radar, and vision factors $\mathcal{F}_{k-l:k}^{\star}$ in the window time frame, denoted with the same notation as the residuals and covariance matrices. For each \ac{imu} measurement, a prediction is made from the current graph state, resulting in a high-rate odometry output. Upon receiving a new exteroceptive sensor measurement, the graph is updated with the relevant factors, and the optimal state estimate is calculated, alongside maps updated if the measurement was a LiDAR measurement.

\begin{figure}[h]
    \centering
    \includegraphics[width=\linewidth]{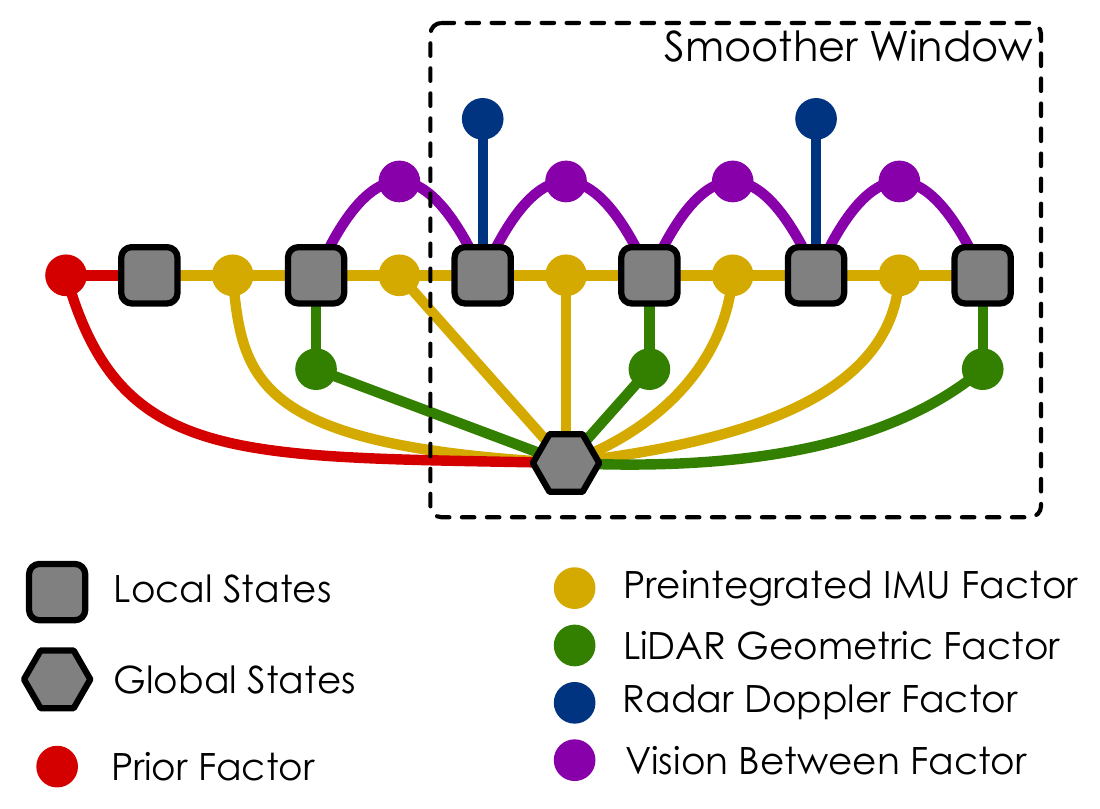}
    \caption{Representative exemplary factor graph constructed by the multi-modal estimator.}
    \label{fig:perception:factor_graph}
\end{figure}

It is not required that all sensors operate at the same frequency, nor that their measurements arrive chronologically by their timestamps. The method does, however, assume that the timestamps of the input measurements are accurate, i.e., some effort from the implementer has been taken to ensure the accuracy of the timestamping of the sensor data in the form of hardware or software-based time synchronization, such that all timestamps are on a common time axis. A representative factor graph constructed by the multi-modal estimator is shown in \cref{fig:perception:factor_graph}. We first give an overarching view of the operation of the system and later detail how each of the sensor measurements is used. 

\paragraph{Initialization}

The method assumes the system is stationary at startup. During this period, \ac{imu} measurements are accumulated over a \SI{1}{\second} window and averaged, yielding the mean accelerometer reading $\vect{\mu}{a}{}$ and mean gyroscope reading $\vect{\mu}{g}{}$. Since the system is at rest, the gyroscope mean is a direct estimate of the gyroscope bias, so we set $\vect{b}{g,0}{} = \vect{\mu}{g}{}$. The accelerometer mean satisfies $\vect{\mu}{a}{} = \vect{b}{a,0}{} + (\rot{R}{B,0}{W})^\top \vect{g}{W}{}$,
under the assumption that $\vect{b}{a,0}{}$ is constant over the initialization window and that measurement noise is negligible. Because the accelerometer bias is not yet known, we approximate $\rot{R}{B,0}{W}$ by aligning $\vect{\mu}{a}{}$ with the gravity direction $\begin{bmatrix} 0 &0 &1\end{bmatrix}^\top$, which determines the roll and pitch components. The yaw component is unobservable from accelerometry alone and is set to zero. This approximation introduces an attitude error proportional to the magnitude of $\vect{b}{a,0}{}$; for instance, a bias of \SI{1}{\meter\per\second\squared} yields an initial attitude error of approximately \SI{5}{\degree}, following~\cite{farell2008aidednavigation}. The gravity direction $\vect{g}{W}{}$ is estimated, instead of being fixed, to account for this error.  The initial position $\vect{p}{WB,0}{W}$ and velocity $\vect{v}{WB,0}{W}
$ are both set to zero.
 
Initialization is triggered upon arrival of the first exteroceptive sensor measurement, provided the \ac{imu} buffer spans at least $\SI{1}{\second}$. In the case of LiDAR, this step additionally initializes the global map, as described in the following section.

\paragraph{Propagation}

Upon receiving a new measurement from the \ac{imu}, the measurement is added to a buffer for future use in the main thread. In a separate thread, this measurement is used to propagate the latest state in the graph, which is then published to provide high-rate odometry for feedback control.

Upon receiving a new measurement from an exteroceptive sensor, typically, the method (a) creates a new state at the timestamp derived from the measurement, (b) connects the new state with the remaining graph with a preintegrated \ac{imu} factor derived from the \ac{imu} buffer, (c) adds a factor derived from the measurement, (d) optimizes the graph, and (e) publishes the new state. However, depending upon the actual sensor stream and system, the method may deviate slightly from this. We now detail these deviations. If the timestamp of the incoming measurement is older than the lag window, it is discarded prioritizing low-latency data. If the timestamp of the new measurement is very close (i.e., there are no \ac{imu} measurements in between) to the timestamp of any of the states in the window, then we treat the measurement as having the same timestamp as that state, i.e., we do not create a new state and use the matched state for adding a measurement-derived factor. This has the added advantage that late measurements arriving to the graph out-of-order can seamlessly be integrated, assuming they are within the smoother window. If the timestamp is older than the newest state in the window (i.e., it arrived with high latency), then we identify the correct location that it should have been added as well as the preintegrated \ac{imu} factor that currently connects the states straddling this timestamp. The preintegrated \ac{imu} factor is then replaced by a new state at the new timestamp, along with the measurement-derived factor and two new preintegrated \ac{imu} factors. 

The next paragraphs detail the specifics on the handling of each sensor modality. Note, some factor residuals are constructed on a per-point basis, e.g., in the case of the LiDAR and radar factors. For computational savings, these are implemented as a single Hessian factor by summing individual contributions. For brevity, the factor residuals will be defined on a per-point basis, and the summation implied.

\paragraph{Inertial Measurement Unit}
\ac{imu} measurements are stored in a buffer on arrival, for easy use upon receiving measurements from one of the aiding sensors. At that point, the corresponding exteroceptive factors are created and connected to the graph by an \ac{imu} preintegration factor, following~\cite{forster2017manifold}, with the following residuals that preintegrate the measurements between times $i$ and $j$

\begin{equation}
    \bm{e}_{\mathcal{I}} = \begin{bmatrix} \bm{e}_{\mathcal{I},\mat{R}}^\top &\bm{e}_{\mathcal{I},\bm{p}}^\top   &\bm{e}_{\mathcal{I},\bm{v}}^\top \end{bmatrix}^\top,
\end{equation}
including attitude $\bm{e}_{\mathcal{I},\mat{R}}$, position $\bm{e}_{\mathcal{I},\bm{p}}$, and velocity $\bm{e}_{\mathcal{I},\bm{v}}$ error terms. The reader is referred to~\cite{forster2017manifold,gtsam} for a detailed description of each term.

\paragraph{LiDAR}
The point cloud measurement obtained from a LiDAR typically contains several thousand points that were sampled at various instants between $[t,t+t_s]$ where $t$ is the timestamp of the measurement and $t_s$ is the duration of the sweep. Since the LiDAR is likely to have moved during this time, it is necessary to deskew the point cloud to account for this motion. As the \ac{imu} measurements for this duration are available, we use them to propagate the latest state in the graph up to $t+t_s$, storing the intermediate pose $\mathbf{\tilde{T}}_{\mathtt{B, t+t_k}}^{\mathtt{W}}$ for every unique timestamp $t+t_k$ in the point cloud. Using these intermediate poses, we iterate over the points $\vect{r}{}{L, t+t_k}$ at each unique timestamp and transform them to the body frame at the timestamp of the last point $t+t_s$ as given by 
\begin{equation}
    \vect{r}{}{B, t+t_s} = \mathbf{\tilde{T}}_{\mathtt{W}}^{\mathtt{B, t+t_s}} \mathbf{\tilde{T}}_{\mathtt{B, t + t_k}}^{\mathtt{W}} \mathbf{T}_{\mathtt{L}}^{\mathtt{B}} \circ \vect{r}{}{L, t+t_k},
\end{equation}
where $\circ$ denotes the homogeneous transformation action on a vector in $\mathbb{R}^{3}$.

For further processing, assuming that the IMU propagation was correct, the point cloud is considered to have been sampled instantaneously at $t + t_s$.
Afterwards, the point cloud is downsampled, for computational efficiency, first by removing three out of four points, and second by organizing the point cloud into a voxel grid and subsampling, ensuring a maximum of $n_p$ points per voxel and a minimum distance of $\eta_p$ between any two points in a voxel. Afterwards, the correspondences are found by relating points in the current cloud with planes fit in the map; these correspondences are added to the graph in the form of point-to-plane residuals. This per-point residual $\epsilon_{\mathcal{L}}$ is calculated as follows

\begin{equation}
    \epsilon_{\mathcal{L}} = \vect{n}{}{W} \cdot \left( \rot{R}{B}{W} \vect{\tilde{r}}{\mathcal{L}}{B} + \vect{p}{WB}{W} - \vect{r}{0}{W} \right),
\end{equation}
for a plane defined by the normal $\vect{n}{}{W}$ and point $\vect{r}{0}{W}$ and the corresponding transformed point $\vect{\tilde{r}}{\mathcal{L}}{B}$ from the downsampled LiDAR point cloud. The per-point residuals are whitened using the point noise covariance $\sigma_{\epsilon_{\mathcal{L}}}^{2}$ and assembled into a single dense hessian factor.
For outlier rejection, these residuals are augmented with Huber M-estimators~\cite{huber1964hubernorm}. Post-optimization, the pose is compared with previous key frames, and if a significant difference in position or attitude is detected, a new key frame is created, and the current point cloud is added to maintain a monolithic map.

\paragraph{Radar}
In the context of this method, \ac{fmcw} radars are assumed to return point cloud measurements, where each point is defined by its 3D position $\vect{r}{\mathcal{R}}{}$ and radial speed $v_{r}$.
Unlike the estimator proposed in~\cite{perception_dlrio}, here the \ac{ransac} least-squares calculation of linear velocity from the radar point cloud is omitted. Instead, the individual points from the radar point cloud are integrated into the graph. By directly integrating the radial speed measurements, we avoid the potential limitations associated with first estimating linear velocity independently. Namely, these are the minimum number and diversity of points required for fully resolving the 3 axes of linear velocity. Even with sufficient points, low point cloud sizes can still result in poor estimation of either the velocity or the covariance matrix, leading to degraded performance.

Thus, the per-point residual $\epsilon_{\mathcal{R}}$ for the radar Doppler factor is

\begin{equation}
    \epsilon_{\mathcal{R}} = -\vect{\hat{v}}{WR}{R} \cdot \frac{\vect{\tilde{r}}{\mathcal{R}}{}}{\lVert \vect{\tilde{r}}{\mathcal{R}}{} \rVert} - \tilde{v}_{r},
\end{equation}
where $\vect{\hat{v}}{WR}{R}$ is the radar-frame velocity estimate and $\tilde{\bm{r}}$, $\tilde{v}_{r}$ are the radar point position and radial speed measurements. The radar-frame velocity is composed from the state estimates as

\begin{equation}
    \vect{\hat{v}}{WR}{R} = \rot{R}{B}{R} \left( (\rot{\hat{R}}{B}{W})^\top \vect{\hat{v}}{WB}{W} + \left(\vect{\bar{\omega}}{WB}{B} \times \vect{p}{BR}{B} \right) \right),
\end{equation}
assuming the extrinsic translation $\vect{p}{BR}{B}$ and rotation $\rot{R}{B}{R}$ between \coord{B} and \coord{R} is known a priori, and that the angular rate during a given radar chirp period $\vect{\bar{\omega}}{WB}{B}$ can be accurately estimated by averaging the \ac{imu} gyroscope measurements. 
As the radar sensor is known to generate spurious points~\cite{harlow2024newwave}, the residual is augmented with a Cauchy M-estimator for outlier rejection. This has the added benefit of improved resilience against dynamic objects by suppressing the influence of such outliers. The choice of Cauchy is motivated by the desire for an M-estimator that more rapidly nullifies the impact of significantly large outliers.

\paragraph{Vision}
Vision factors are added in a loosely-coupled manner, taking advantage of the wealth of capable estimators that exist in the vision community. Specifically, a visual-inertial estimator based on~\cite{bloesch2017Rovio} processes the camera and \ac{imu} measurements, creating odometry estimates as a result. The pose estimates $\rot{\tilde{T}}{B}{W}$ from this external method are stored in a buffer. This information is incorporated, if it passes a D-Optimality pose quality check~\cite{carrillo2012dopt}, into the factor graph with a relative transform factor, which compares the relative transform between pose estimates $\rot{\hat{T}}{B}{W}$ and the aforementioned measurements across the same time interval. Assuming vision pose measurements are available for times $t_{i}$ and $t_{j}$, the factor residual can be calculated as

\begin{equation}
    \bm{e}_{\mathcal{V}} = \Log \left( \left( (\mathbf{\tilde{T}}_{\mathtt{B},i}^{\mathtt{W}})^{-1} \mathbf{\tilde{T}}_{\mathtt{B},j}^{\mathtt{W}} \right)^{-1} \left( (\mathbf{\hat{T}}_{\mathtt{B},i}^{\mathtt{W}})^{-1} \mathbf{\hat{T}}_{\mathtt{B},j}^{\mathtt{W}} \right) \right),
\end{equation}
where $\Log$ denotes the logarithmic map from the $\textit{SE}(3)$ manifold to its Lie algebra $\mathfrak{se}(3)$.
This approach draws inspiration from \cite{khattak_complementary_2020,khedekar_mimosa_2022}.

\subsubsection{Vision-Language Reasoning}

\begin{figure}[h]
    \centering
    \includegraphics[width=\linewidth]{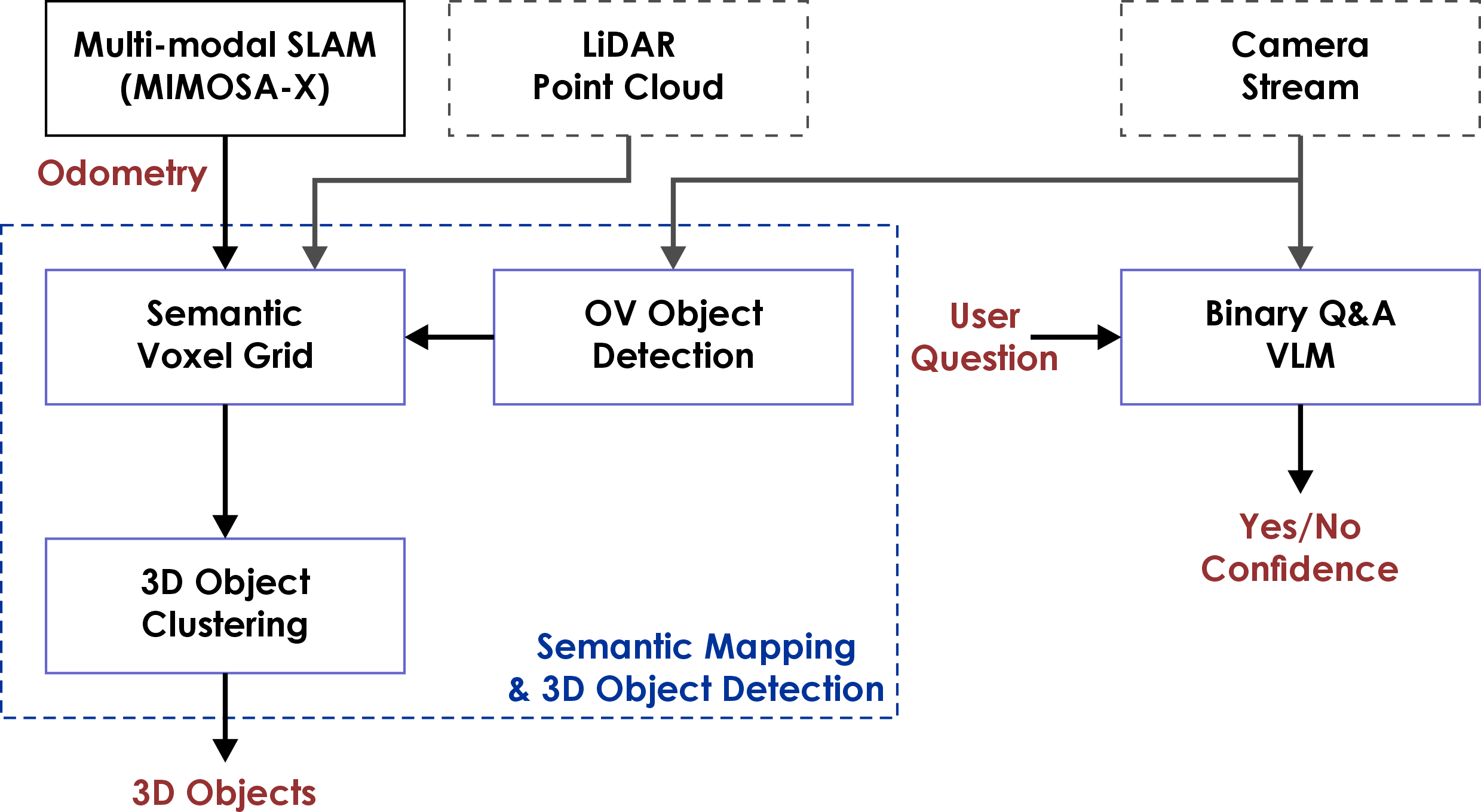}
    \caption{Illustration of the current VLM-based functionality of the \ac{ua} for semantic scene mapping and visual Q\&A.}
    \label{fig:perception:vlmreasoning}
\end{figure}

Our semantic reasoning system integrates two complementary vision-language model (VLM) capabilities: (i) open-vocabulary object perception with semantic 3D mapping, and (ii) binary visual question-answering (Yes/No) for high-level scene reasoning. Together, these capabilities collectively enable semantic scene understanding and contextual judgment and decision making from online visual data. The \ac{vlm}-based functionality is illustrated in Figure~\ref{fig:perception:vlmreasoning}. Even though our implementation employs YOLOe~\cite{Wang2025YOLOE} and GPT-5, the proposed semantic reasoning system is designed to be compatible with other open-source and proprietary \ac{vlm}s that support open-vocabulary object detection and visual question-answering tasks.

\paragraph{Open-Vocabulary Object Detection and Semantic 3D Mapping}

3D object detection is formulated as a semantic mapping problem. Objects are detected on the camera image using an open-vocabulary detector (YOLOe) or a \ac{vlm}-based detector (GPT-5) initialized with a set of labels. These models produce labeled 2D bounding boxes and associated detection confidences. In parallel, a 3D voxel grid is maintained using LiDAR measurements and pose estimates from our \ac{slam} solution.

To integrate semantic detections into the 3D representation, voxel grid points are projected into the camera frame using the camera extrinsics and the current odometry estimate. For each 2D detection, the subset of projected points that fall within the corresponding bounding box is extracted and clustered to remove outliers. The resulting points are used to update the voxels' semantic values via Bayesian fusion, which uses the object detector's confidences.

Finally, Euclidean clustering is performed for each semantic class present in the voxel grid to extract 3D object instances. These objects are represented as 3D bounding boxes, enabling semantic and spatial reasoning.

\paragraph{Binary Visual Question-Answering}

For high-level semantic assessment, a \ac{vlm} (GPT-5) processes the front-camera image together with a binary ``Yes/No'' question. These visual question-answering tasks typically focus on safety- or navigation-related properties of the scene (e.g., ``is an object blocking a door?''), which can be challenging to infer from geometric information alone. The model produces a binary answer, alongside its response confidence (ranging from 0 to 1) and a brief explanation of its reasoning.

\subsection{Planning Module}
\begin{figure*}
	\centering
	\includegraphics[clip, trim = 0cm 0cm 0cm 0cm, width=1\linewidth]{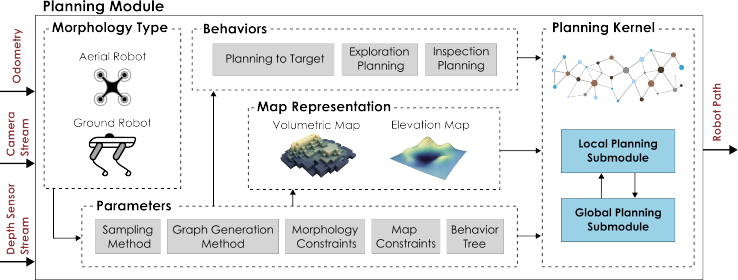}
	\caption{Planning module architecture. The \planning{} is facilitated through \acs{gbp3} designed to work universally across aerial, ground, and underwater robot morphologies. At the core, the planner integrates a domain- and morphology-agnostic planning kernel that utilizes the dual map representation to build local and global graphs, both satisfying the robot motion constraints set by the \texttt{Robot Abstraction Layer}. The local graph is used for local information gathering and collision-free navigation, whereas the global graph is used for fast but coarse planning in the entire known space. Utilizing these, the appropriate behavior (Exploration, Inspection, Planning to Target) is executed as per the task set by \texttt{Mission Abstraction Layer}.}
	\label{fig:planning}
\end{figure*}

The \planning{} in the \ac{ua} is facilitated through \acs{gbp3}~\cite{zacharia2026omniplannerarxiv}, a graph-based planner designed to work across diverse aerial, ground, and underwater robot morphologies. The planner is currently applicable to systems for which graph-based planning is a viable option (e.g., various thrust-controlled aerial and underwater systems such as multirotors and ROVs, legged robots, and differential drive ground rovers). At the core, the planner utilizes a unified planning kernel that is agnostic to robot morphology, environment type, and mission objective and provides both target navigation as well as informative planning behaviors, such as exploration and inspection, based on the task objective $\missionobj$ ($\missionobj_{TP}$: Planning to a target, $\missionobj_{EP}$: Exploration Planning, $\missionobj_{IP}$: Inspection Planning) set by the \texttt{Mission Abstraction Layer}. Algorithm~\ref{alg:planning_module} gives a high-level overview of the \planning, while Figure~\ref{fig:planning} illustrates the different parts of the module and their interactions. The reader is referred to~\cite{zacharia2026omniplannerarxiv} for the algorithmic details regarding this module.

\begin{algorithm}[t]
    \caption{\planning}
    \begin{algorithmic}[1]
        \State $\robotconstraints \gets \textbf{GetRobotConstraints()} $ \Comment{\texttt{Robot Abstraction Layer}}
        \State $\{ \depthsensor, \camerasensor \} \gets \textbf{GetSensorParams()} $ \Comment{\texttt{Robot Abstraction Layer}}
        \State $\missionobj \gets \textbf{GetMissionObjective()} $ \Comment{\texttt{Mission Abstraction Layer}}
        \Repeat
            \State $\graph_L \gets \textbf{BuildLocalGraph(} \robotconstraints, \Ms, \Hs \textbf{)} $ \Comment{Planning Kernel}
            \State $\graph_G \gets \textbf{UpdateGlobalGraph(} \robotconstraints, \graph_L \textbf{)} $ \Comment{Planning Kernel}
            \If{$\missionobj = \missionobj_{TP}$}
                \State $\textbf{PlanningToTarget(} \robotconstraints, \pbf_t, \depthsensor, \graph_L, \graph_G \textbf{)} $
            \ElsIf{$\missionobj = \missionobj_{EP}$}
                \State $\textbf{ExplorationPlanning(} \robotconstraints, \depthsensor, \graph_L, \graph_G, \Ms \textbf{)} $
            \ElsIf{$\missionobj = \missionobj_{IP}$}
                \State $\textbf{InspectionPlanning(} \robotconstraints, \depthsensor,\camerasensor, \graph_L, \Ms \textbf{)} $
            \EndIf
        \Until{\textrm{Robot Endurance Critical}}
        \State $\textbf{ReturnToHome(} \graph_L, \graph_G, \robotconstraints \textbf{)} $ \Comment{Planning Kernel}
    \end{algorithmic}
    \label{alg:planning_module}
\end{algorithm}

\subsubsection{Planning Kernel}\label{subsub:planning_kernel}
The Planning Kernel of \acs{gbp3} serves as the backbone of the methodology, providing the data structures and functions for searching the robot's admissible configuration space to enable the desired behaviors. All operations take place on a dual environment representation consisting of a) a Volumetric Map $\Ms$ (in this work Voxblox~\cite{planning_voxblox}), with voxel size $v_m$, and, when applicable, b) an elevation map $\Hs$ for ground robots (utilizing the work from~\cite{fankhauser2018probabilistic,Fankhauser2014RobotCentricElevationMapping}), with grid size $v_h$. The kernel employs a bifurcated local/global planning architecture. 
The \textit{Local Planning Submodule} operates in a local volume $\bm{b}_L$ and constructs a bounded, sampling-based, dense graph $\graph_L$, with its vertex and edge sets $\Vs_L, \Es_L$ inside $\bm{b}_L$ spanning the locally reachable configuration space. All vertices and edges are sampled such that they lie entirely in collision-free space and respect the robot motion constraints $\robotconstraints$ defined by the \texttt{Robot Abstraction Layer} (e.g., traversability, robot size $\bm{b}_R$, etc.). The graph $\graph_L$ is used by the \planning~ for local information gathering and collision-free navigation as described in the subsequent subsections.
The \textit{Global Planning Submodule} maintains a sparse global graph $\graph_G = \{\Vs_G, \Es_G\} $ built by aggregating the sparsified local planning graphs across the mission. This graph is used to represent the entire known space, providing fast global planning functionality. As $\graph_G$ is built from $\graph_L$, all vertices and edges in $\Vs_G, \Es_G$ are in free space and satisfy the robot motion constraints.
The Global Planning Submodule also keeps track of the robot's remaining endurance (or otherwise-defined remaining mission time). At each planning iteration, the Planning Kernel checks if the remaining time is sufficient to execute the path given by the specific behavior and return to the start location. If yes, that path is executed, else, the Kernel triggers a homing manuever to guide the robot back to the starting location.

\subsubsection{Planning to a Target}\label{subsub:planner_tr}
\ac{ua} facilitates planning to a desired waypoint $\pbf_t$ both within the already explored space, as well as in the unknown, as long as this is iteratively found to be possible. In each planning iteration, first, a guiding path $\planningpath_t$ is calculated. If $\pbf_t$ lies in the known (explored) space, $\planningpath_t$ is simply calculated as the shortest path along $\graph_G$ to the vertex in $\Vs_G$ closest to $\pbf_t$. Otherwise, $\planningpath_t$ is calculated as the path towards the vertex that is on the frontier of the explored space and is closest to $\pbf_t$.
Next, $\graph_L$ is used to calculate a local path $\planningpath_{TP}$ guiding the robot along $\planningpath_t$ which is then commanded to the subsequent modules to track. When the robot reaches within a distance $d_{path}$ from the last point in $\planningpath_{TP}$, the next iteration of the planner is triggered. This entire process is repeated until the robot reaches $\pbf_t$ or no progress can be made towards it, at which point the planner declares that the waypoint is unreachable. It is highlighted that the homing maneuver mentioned in \cref{subsub:planning_kernel} uses the Planning to Target behavior with the starting location as $\pbf_t$.

\subsubsection{Exploration Planning}

The first informative planning behavior supported by OmniPlanner is the \ac{ve}, where the robot is tasked to iteratively uncover the unknown volume using an \ac{fov}- and range-constrained depth sensor $\depthsensor$ (whose parameters are given by the \texttt{Robot Abstraction Layer}). In each planning iteration, $\graph_L$ is used to find the path that leads to uncovering the largest amount of unknown space. First, shortest paths from the current robot location to each vertex in $\Vs_L$ are calculated. An information gain, called Volume Gain, related to the amount of unknown volume mapped by $\depthsensor$ from a robot configuration, is calculated for each vertex in $\Vs_L$. The path $\sigma_{EP}$ with the highest aggregated Volume Gain is selected as the next exploration path. When the robot reaches within a distance $d_{path}$ from the last point in $\planningpath_{EP}$, the next iteration of the planner is triggered and the process is repeated.
When no informative path is found in $\graph_L$, the $\graph_G$ is utilized to reposition the robot to a frontier of the explored space. The planner tracks vertices in $\graph_G$ having high Volume Gain (called frontier vertices), and repositions the robot to the frontier vertex having the highest gain using the Planning to Target behavior described in \cref{subsub:planner_tr}. Upon reaching the frontier, local exploration continues.

\subsubsection{Inspection Planning}

In the \ac{vi} behavior, the planner is tasked to inspect a subset of the occupied surface in the mapped volume using an \ac{fov}- and range-constrained camera sensor $\camerasensor$ (whose parameters are given by the \texttt{Robot Abstraction Layer}) at the desired viewing distance $d_{view}$ (given by the \texttt{Mission Abstraction Layer}). A set $\viewpointset$ of viewpoints, at a distance $d_{view}$ from the occupied surface, is built. A graph $\graph_{VI}$ is built using the Local Planning Submodule to connect the viewpoints in $\viewpointset$. The minimal viewpoint set $\viewpointset_{best}$ viewing the entire surface is selected, and the order to visit them is calculated by solving the \ac{tsp} problem. The shortest paths along $\graph_{VI}$ connecting the subsequent viewpoint in the tour are concatenated to form the inspection path.

\subsection{Navigation Module}

\begin{figure*}
	\centering
	\includegraphics[width=0.99\textwidth]{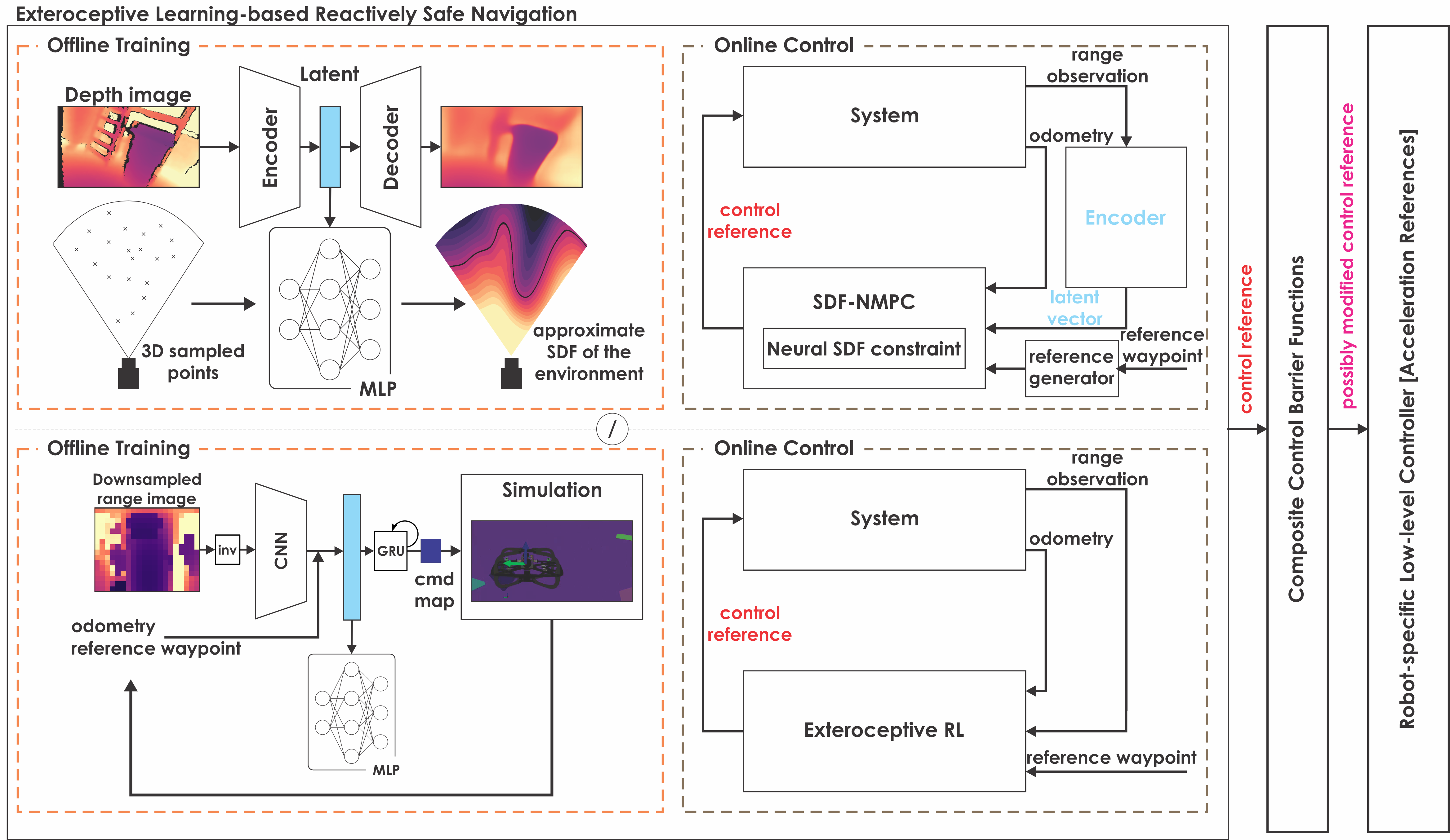}
    \caption{Navigation module architecture. Two swappable local navigation modalities are offered: (\textit{top}) \acs{neuralmpc}, which encodes depth images into a latent \acs{sdf} representation online and embeds it as a constraint in an \acs{mpc} controller; and (\textit{bottom}) \acs{exteroceptivedrl}, which combines inverted range images with proprioceptive state through a policy trained using \acs{ppo} in simulation to directly output acceleration commands for waypoint-directed collision-free navigation. In both cases, the resulting acceleration control reference is passed through the \acs{ccbf} safety filter, which composes range measurements into a composite barrier function to reactively modify the reference acceleration and guarantee collision-free operation as a last resort, before forwarding to the robot-specific low-level controller.}
	\label{fig:navigation}
\end{figure*}

The \ac{ua} takes a multi-layered approach to safety, illustrated in Figure~\ref{fig:navigation}. Conventional modern safe navigation and collision avoidance are based on the planning of paths subject to map constraints, which are then blindly followed by an onboard controller. Despite the major success of this paradigm in many environments and mission profiles, as discussed in~\cite{navigation_neuralmpc,navigation_drl} and demonstrated in field experience~\cite{extremeslam_tro} it represents a single point of failure which can lead robots to collisions due to odometry errors or erroneous/incomplete mapping. Although such errors are not common, experience shows that they do manifest and are often catastrophic, at least regarding a robot's ability to continue its mission. Furthermore, odometry and mapping challenges manifest more frequently in perceptually-degraded and geometrically complex environments~\cite{extremeslam_tro}. While the \ac{ua} does perform map-based avoidance through volumetric mapping for collisions and traversability analysis for ground systems, it further adds two redundant layers of safety: depth sensor-based trajectory tracking and a last-resort safety filtering based on \acp{cbf}.
With respect to depth sensor-driven navigation policies, two approaches are offered within the stack, owing to their distinct benefits in certain conditions: (i) a \ac{sdf}-based neural \ac{mpc}~\cite{navigation_neuralmpc} or (ii) a novel \ac{drl}-based policy trained for safe navigation and smooth collision avoidance.
These methods enable local deviations from the reference trajectory if and when necessary, to ensure collision avoidance. Both methods provide swappable local navigation policies with distinct performance characteristics that will be discussed in Section~\ref{sec:ablation}. To further assert safety with formal guarantees, a composite \ac{cbf}-based formulation building upon~\cite{navigation_ccbf} is introduced as a last-resort safety filter allowing to modify the control references in the unlikely situation that all other collision-avoidance methods in the stack fail.
Importantly, this additional safety layer within the \ac{ua} is currently implemented for obstacle avoidance and not for traversability analysis, predominantly targeting flying robots. This architecture extends upon recent ideas and developments in the research community, such as the perceptive locomotion in~\cite{takahiro2022perceptive}, which indeed may also be used as an alternative to the exteroceptive depth navigation strategies discussed below when it comes to quadruped robots. 

\subsubsection{Neural SDF-NMPC}\label{sec:ua_nmpc}
Detailed in~\cite{navigation_neuralmpc}, the \acs{neuralmpc} enables collision-free navigation in unknown environments relying only on depth sensing and (possibly drifting) odometry. In its current implementation, it emphasizes rotorcraft navigation.

The \acs{neuralmpc} represents the visible environment as a Euclidean \ac{sdf}, defined to be positive in visible free space and negative in occluded regions (i.e., behind obstacles). For efficient computation and to provide a representation compatible with gradient-based \ac{mpc}, this \ac{sdf} is approximated by a neural network, online, from the latest depth measurement. As a result, the environment is described purely locally, improving robustness to potentially drifting odometry.
To keep the representation compatible with a compact neural model, the \ac{sdf} is saturated beyond a threshold $T_\text{SDF}$. This \ac{sdf} is constructed online from the latest depth measurement. As a result, the environment is described purely locally, improving robustness to potentially drifting odometry.

A two-stage neural network is used. First, the input depth image is clamped at a distance $d_\text{max}$, since only short-range surroundings are relevant for short-horizon collision avoidance. Compression into a low-dimensional latent space $\mathbf{z}$ is achieved via a convolutional encoder, trained jointly with a decoder to reconstruct the input, ensuring that $\mathbf{z}$ captures a reliable latent representation of the depth data. Notably, the encoding is biased to place greater emphasis on obstacles close to the robot, encouraging accurate encoding of nearby geometry. The decoder is used only during training, while the encoder provides input to a downstream \ac{mlp} network that reconstructs the \ac{sdf}.

Specifically, this \ac{mlp} takes as input the latent vector and a $3$D position, and approximates the corresponding \ac{sdf} value evaluated in that point. The regression task is trained in a supervised manner, including losses that enforce consistency of the \ac{sdf} gradient. In this way, the trained \ac{mlp} represents the following parametric function:

\begin{equation}
    \begin{split}
        \mathbb{R}^3 &\rightarrow \mathbb{R} \\
        \bm{p} &\mapsto \text{SDF}_{\bm{\theta},\mathbf{z}}(\bm{p})~,
    \end{split}
\end{equation}
where $\bm{\theta}$ are the neural network weights, and $\mathbf{z}$ is the latent code corresponding to the depth measurement.

Finally, the neural \ac{sdf} is embedded into the nonlinear \ac{mpc} controller as an explicit position constraint. The following constraint is enforced over the receding horizon:

\begin{equation}
    \text{SDF}_{\bm{\theta},\mathbf{z}}(\vect{p}{S}{B}) \ge r,
\label{eq:mpc:const_sdf}%
\end{equation}
where $r$ is a user-defined threshold accounting for the robot radius and a possible safety margin, and $\vect{p}{S}{B}$ denotes the position of the robot expressed in the frame \coord{S} in which the depth measurement was captured.
Note that additional constraints further ensure that the robot remains within the sensor frustum, i.e., within the region that is currently observable and where the neural \ac{sdf} is defined. This follows the intuitive principle of ``look where you move'', effectively restricting motion to visible free space.

Critically, the method enforces feasibility and stability-type criteria (under fixed sensor observations), with a terminal condition ensuring that the terminal state allows a collision-free braking maneuver to hover. This is assessed by computing the minimum braking distance given the input bounds, and evaluating the \ac{sdf} at the predicted hovering position. Under this condition, recursive feasibility is ensured, and with a suitable quadratic terminal cost, the optimal value is shown to be non-increasing over time. The \ac{mpc} generates acceleration commands from a velocity reference trajectory, since velocity tracking prevents the accumulation of position errors when collision constraints prevent accurate tracking of a nominal trajectory. Accordingly, the interface with the \planning{} provides such velocity references derived from planned paths.

\subsubsection{Exteroceptive Deep Reinforcement Learning}\label{sec:ua_drl}

The \ac{ua} further offers exteroceptive \ac{drl}-based navigation (dubbed \acs{exteroceptivedrl}). The proposed novel \ac{exteroceptivedrl} approach is formulated as a waypoint navigation problem and considers as input the vector to goal location, the robot orientation, velocity, and angular rates alongside the instantaneous depth image from an exteroceptive sensor (including stereo or RGB-D cameras, Time-of-Flight camera sensors, or LiDARs). We employ end-to-end learning to train a navigation policy to generate commands directly from the robot's current state and range measurement. The policy is trained using the Aerial Gym Simulator~\cite{navigation_aerialgym} to command acceleration and yaw-rate setpoint commands (as commonly provided by most autopilots such as PX4~\cite{meier2015px4} and ArduPilot~\cite{ardupilot}). Relevant open-source examples for training are provided in Aerial Gym, while the policy can be trained on any compatible simulation tool.
Similar to the \acs{neuralmpc}, \ac{exteroceptivedrl} ensures safe collision-free navigation without a map and thus contributes to multi-layered safety. %
This work is distinct from prior work of the authors in~\cite{kulkarni_semantically-enhanced_2023, navigation_dce}, by a) departing from a modularized two-step approach and introducing an end-to-end methodology, b) introducing a novel reward function incorporating \acf{ttc}, and c) commanding acceleration and yaw-rate setpoints instead of velocity references.%

\subsubsection*{Observations}

The observation vector for training the policy consists of both proprioceptive and exteroceptive components. The proprioceptive components are expressed in a yaw-aligned, roll and pitch stabilized coordinate frame \coord{V} sharing the same origin as the robot body IMU frame \coord{B}. Let $\vect{p}{}{W},\vect{p}{*}{W}\!\in\mathbb{R}^{3}$ be the robot and target
positions, $\psi, \psi^{*}$ be the robot and target yaw respectively, the yaw error be $\psi_{e} = ((\psi^{*} - \psi + \pi) \bmod 2\pi) - \pi$, and $\vect{\delta}{}{W} = \vect{p}{*}{W} - \vect{p}{}{W}$ be the position error with $\delta=\rVert \boldsymbol{\delta} \lVert$ the vector to the goal. During training, the goal direction $\vect{\delta}{}{V}/{\delta}$, roll $\phi$ and pitch $\theta$ measurements are perturbed by adding noise sampled from a uniform distribution before populating the observation tensor. The linear $\vect{v}{WB}{B}$ and angular velocities $\vect{\omega}{WB}{B}$ are not perturbed. The range measurements from the exteroceptive sensor are min-pooled to $16 \times 20$ pixels and inverted as a $2$D inverse range-image $\mathbf{I_r}$.

\begin{table}[ht]
\centering
\caption{Observation vector $\bm{o}_{t}\in\mathbb{R}^{337}$.}
\label{tab:obs}
\begin{tabular}{@{}clp{4.5cm}@{}}
\toprule
Index & Symbol & Description \\
\midrule
$0$:$2$    & $\vect{\delta}{}{V}/{\delta}$ & Goal direction in \coord{V} \\
$3$        & $\delta$                       & Distance to goal (\si{\meter}) \\
$4$        & ${\phi}$            & Roll (\si{\radian}) \\
$5$        & ${\theta}$          & Pitch (\si{\radian}) \\
$6$        & $\psi_{e}$                & Yaw error (\si{\radian}) \\
$7$:$9$   & $\vect{v}{WB}{B}$          & Linear velocity in \coord{B} (\si{\meter\per\second}) \\
$10$:$12$  & $\vect{\omega}{WB}{B}$ & Angular velocity in \coord{B} (\si{\radian\per\second})\\
$13$:$16$  & $\bm{u}_{t-1}^{acc}$          & Previous setpoint (\si{\meter\per\second\squared}, \si{\radian\per\second})  \\
$17$:$336$ & $\mat{I_r}$              & Inverse-range image (\si{\meter}$^{-1}$) \\
\bottomrule
\end{tabular}
\end{table}

\subsubsection*{Actions}

The \ac{drl} policy outputs a normalized action command $\bm{a}_{t} = \{a_x, a_y, a_z, a_{\dot{\psi}}\} \in[-1,1]^{4}$ that is scaled and mapped to linear acceleration and yaw-rate setpoint commands at time $t$ as $\bm{u}^{acc}_{t}=(u_n^xa_{x},\,u_n^ya_{y},\,u_n^za_{z},\,u_n^{\dot{\psi}}a_{\dot\psi})$, where $u_n^x,\,u_n^y,\,u_n^z,\,u_n^{\dot{\psi}}$ are tunable scaling parameters, with their values described in Table~\ref{tab:evaluation:params}. %

\subsubsection*{\acf{ttc}}

We propose the usage of an expected ``time-to-collision'' metric at each timestep that provides a dense reward signal to the robot. This metric is distinct compared to approaches that directly reward based on raw range data~\cite{navigation_drl}, thus prioritizing the relationship between velocity and distance to measured obstacles, only penalizing rapid approaches to obstacles. This metric is calculated only in simulation and is treated as privileged information that is not available to the policy, yet influences the reward signal. For each point $i$ in the full-resolution point cloud expressed in a sensor frame \coord{S}, the vector from the sensor is calculated as ${\vect{r}{}{S}}_i$. The linear component of velocity of the robot along the direction to each point is computed as the projection of $\vect{v}{WB}{S}$ onto the unit direction vector of ${\vect{r}{}{S}}_i$. This is subsequently used to calculate the expected time to collision $\tau_i$ using the distance of the point from the robot:

\begin{equation}
    v^{\perp}_{i} = \vect{v}{WB}{S}\cdot \frac{{\vect{r}{}{S}}_i}{\lVert {\vect{r}{}{S}}_i \rVert}, \qquad
    \tau_{i} =
    \begin{cases}
        \lVert {\vect{r}{}{S}}_i \rVert/v^{\perp}_{i} & v^{\perp}_{i}>0, \\
        10\,\mathrm{s}                 & v^{\perp}_{i}\leq 0.
    \end{cases}
\end{equation}

Positive time-to-collision values are clamped between \SI{0}{\second} and \SI{10}{\second}. 
Negative values indicate that the robot is moving away from an obstacle, hence they are set to \SI{10}{\second} to make their effect negligible. The minimum time-to-collision $\tau_{min} = \min_i \tau_i$ across all points is considered and used to penalize the robot, emphasizing imminent collisions while largely ignoring well-separated obstacles.

\subsubsection*{Policy Architecture}

The observation vector $\bm{o}_t$ is partitioned into a proprioceptive and an exteroceptive component, as detailed in Table~\ref{tab:obs}. The latter is treated as a single-channel $2$D image and processed by a \ac{cnn}-based encoder $f_{\text{CNN}}$ with three convolutional blocks. Each block consists of a $3\times3$ convolution with padding of width $1$, an \ac{elu} activation, and a $3\times3$ max-pooling layer. The resulting feature map is flattened into a 128-dimensional range embedding, which is concatenated with the proprioceptive state vector. This combined representation is then passed through an MLP with hidden layer sizes $[256, 128, 64]$ and ELU activations, followed by a 128-dimensional \ac{gru} layer. The policy is trained using \ac{ppo}~\cite{schulman_proximal_2017-1} implementation provided by Sample Factory~\cite{petrenko_sample_2020}. The training time is about \SI{60}{\minute} on a consumer grade laptop with an NVIDIA RTX 3080 Ti GPU.

\subsubsection*{Environment Setup and Curriculum}

Each simulated environment is composed of a rectangular room with dimensions ranging from $10 \times 10 \times 6$ to $15 \times 15 \times 10$ m. A simulated robot is initialized on one side of the environment while the goal position is sampled on the opposite side with an arbitrary yaw setpoint. The acceleration and yaw-rate setpoint is tracked using a controller derived from the work in~\cite{lee_geometric_2010}, whose parameters are randomized at each episode to increase robustness and improve sim2real performance. Within each environment, 25 to 70 cuboidal obstacles of various sizes are sampled. An episode is marked as a \textit{success} if the robot reaches within \SI{1}{\metre} of the goal after a predefined number of time steps. If the robot remains collision-free but does not reach the goal, the episode is marked as a \textit{timeout}. Collisions with obstacles are detected by the physics engine, terminate the episode, and are recorded as \textit{crashes}. To encourage stable learning across various environment complexities, a curriculum adjusts the number of obstacles $n_{\textrm{obs}}$ in each environment. The number is increased or decreased when the average success rate $\zeta_s$ over $2048$ episodes crosses the upper or lower thresholds, $\zeta_s^{+}=0.70$ and $\zeta_s^{-}=0.60$, respectively:

\begin{equation}
    n_{\textrm{obs}} \leftarrow
    \begin{cases}
        \min(n_{\textrm{obs}}+2,\;70) & r_{s} > \zeta_s^{+}, \\
        \max(n_{\textrm{obs}}-1,\;25) & r_{s} < \zeta_s^{-}, \\
        n_{\textrm{obs}}              & \text{otherwise.}
    \end{cases}
\end{equation}

The normalized progress fraction $\wp=(n_{\textrm{obs}}-25)/(70 - 25)$ is calculated and used to scale all non-terminal
reward terms by $K(\wp)=1+2\wp\in[1,3]$, amplifying training signal as
task difficulty increases.

\subsubsection*{Reward Function}
The reward function is designed to encourage the robot to navigate efficiently 
to the goal while maintaining safe separation from obstacles and smooth control 
behavior. It contains three groups of terms: (i) goal-directed terms that 
reward proximity and velocity alignment toward the goal, (ii) stabilization 
terms that encourage low velocity, correct heading, and low angular rate when 
in the vicinity of the goal, and (iii) penalty terms that discourage excessive 
speed, large control increments, and proximity to obstacles as captured by the 
\ac{ttc} metric.

Let the speed $v = \rVert \vect{v}{WB}{B} \lVert$, $\bm{u}_{t-1}^{acc}\in\mathbb{R}^{4}$ the previous command;
$\tau_{min} \in [0,10]\,\mathrm{s}$ the minimum time-to-collision across all rays;
and $\wp\in[0,1]$ the curriculum progress fraction. Two kernel functions compose all reward and penalty terms:
\begin{align}
    \mathfrak{R}(\mathfrak{m},\mathfrak{a},\mathfrak{v})&=\mathfrak{m}\exp(-\mathfrak{a} \mathfrak{v}^{2}), \label{eq:erf}\\
    \mathfrak{P}(\mathfrak{m},\mathfrak{a},\mathfrak{v})&=\mathfrak{m}\bigl(\exp(-\mathfrak{a} \mathfrak{v}^{2})-1\bigr). \label{eq:epf}
\end{align}

The following intermediate quantities are defined for compactness and 
summarized in Table~\ref{tab:intermediate}, while all reward terms are presented in Table~\ref{tab:reward}.

\begin{table}[ht]
\centering
\caption{Intermediate quantities used in the reward function.}
\label{tab:intermediate}
\begin{tabular}{@{}lll@{}}
\toprule
Symbol & Definition & Description \\
\midrule
$w(\psi_e)$ & $\mathfrak{R}(1,2,\psi_{e})$ & Heading gate \\
$s$ & $\mathfrak{R}(2,2,v-2)$ & Speed-saturation gate \\
$\xi$ & $\vect{v}{WV}{V}\!\cdot \vect{\delta}{}{V} / (v \cdot \delta) $ & Cosine alignment \\
$w(\delta)$ & $1-\mathfrak{R}(1,2,\delta)$ & ${\approx}0$ near, ${\approx}1$ far \\
$\Delta\bm{u}$ & $\bm{u}_{t}^{acc}-\bm{u}_{t-1}^{acc}$ & Control increment \\
\bottomrule
\end{tabular}
\end{table}

\begin{table}[ht]
\centering
\caption{Reward terms.}
\label{tab:reward}
\begin{tabular}{@{}lp{5.0cm}@{}}
\toprule
Term & Formula \\
\midrule
$r_{\mathrm{pos}}$  & $\mathfrak{R}(3,1,\delta)$ \\
$r_{\mathrm{prox}}$ & $\mathfrak{R}(5,8,\delta)\cdot w(\psi_e)$ \\
$r_{\mathrm{lin}}$  & $(20-\delta)/20$ \\
$r_{\mathrm{vel}}$  & $[\xi s\;\text{if}\;\xi{>}0,\;\text{else}\;{-}0.2]{\cdot}\min(\delta/3,1)$ \\
\cmidrule{1-2}
$r_{\mathrm{spd}}$  & $\mathfrak{R}(1.5,10,v)+\mathfrak{R}(1.5,0.5,v)$ \\
$r_{\mathrm{hdg}}$  & $\mathfrak{R}(2,0.2,\psi_{e})+\mathfrak{R}(4,15,\psi_{e})$ \\
$r_{\omega}$        & $\mathfrak{R}(1.5,5,{\vect{\omega}{WB}{B}}_{z}) \cdot w(\psi_e)$ \\
$r_{\mathrm{stab}}$ & $(r_{\mathrm{spd}}+r_{\mathrm{hdg}}+r_{\omega})\cdot\mathbf{1}_{\delta<1\,\mathrm{m}}$ \\
\cmidrule{1-2}
$P_{\mathrm{spd}}$  & $\mathfrak{P}(2,2,\max(v-3,0))$ \\
$P_{+x}$            & $\mathfrak{P}(2,8,\max({\vect{v}{WV}{V}}_{x},0))\cdot w(\delta)$ \\
$P_{\Delta \bm{u}}$      & $\sum_{i\in\{x,y,z,\dot\psi\}}\mathfrak{P}(0.3,5,\Delta u_n^{i})$ \\
$P_{|\bm{u}|}$           & $\mathfrak{P}(0.1,0.3,u_n^{x})+\mathfrak{P}(0.1,0.3,u_n^{y})$
                       $+\mathfrak{P}(0.15,1,u_n^{z})+\mathfrak{P}(0.15,2,u_n^{\dot\psi})$ \\
$P_{\mathrm{ctrl}}$ & $P_{\Delta \bm{u}}+P_{|\bm{u}|}$ \\ 
$P_{\mathrm{TTC}}$  & $\mathfrak{R}(-3,2,\tau_{min}^{2})$ \\
\bottomrule
\end{tabular}
\end{table}

The total reward at each timestep takes the form:

\begin{equation}
    r_{t} =
    \begin{cases}
        -10 & \text{collision,}\\[3pt]
        \begin{aligned}
            K(\wp)\bigl(
                &r_{\mathrm{pos}}+r_{\mathrm{prox}} + r_{\mathrm{vel}}\\ &+ r_{\mathrm{lin}}
                +r_{\mathrm{stab}}
                +P_{\mathrm{spd}}\\ &+ P_{+x} + P_{\mathrm{ctrl}}+P_{\mathrm{TTC}}
            \bigr)
        \end{aligned}   & \text{otherwise.}
    \end{cases}
    \label{eq:reward}
\end{equation}

\subsubsection{Composite CBF-based Safety Filter}\label{sec:ua_ccbf}

Beyond the aforementioned navigation approaches --which combine map-based safety of the \planning{} with reactive collision avoidance control-- the \ac{ua} further provides a last-resort safety filter. The rationale for adding this final layer is twofold. On the one hand, both the \ac{neuralmpc} and the \ac{drl} navigation strategies involve deep neural network processing, which, despite training to consider noise and other imperfections, is treated as a source of possible (albeit unlikely) error. On the other hand, using fundamentally different collision-checking at different spatiotemporal scales --spanning map-based planning, the navigation strategies, and the safety filter-- offers resourcefulness. It thus reflects a conservative but meaningful choice to safeguard the robot from a collision which represents one of the most problematic events during a mission. 

Based on \acp{ccbf} formalism for safe navigation ~\cite{navigation_ccbf},  we compose a \ac{ccbf} directly from recent range measurements as in~\cite{embedded_ccbf} to modify the acceleration setpoint when an unexpected impending collision is detected. 
It is a key module to fully and formally safeguard autonomous robots. %
The \ac{ccbf} is described hereafter considering the case of a flying robot. First, the system model is approximated to be of degree $2$ and takes the form:

\begin{equation}
    \vect{x}{}{} = \begin{bmatrix} \vect{p}{WB}{W}\\ \vect{v}{WB}{W} \end{bmatrix},\quad \vect{u}{}{}=\begin{bmatrix}\vect{a}{}{W} \end{bmatrix}.
\end{equation}
For a rotorcraft such as a multirotor aerial robot, we here use the simplified linear system model

\begin{equation}    
\dot{\vect{x}{}{}}
= \underbrace{\begin{bmatrix}
    \mathbf{0}_{3} & \mathbf{I}_{3} \\
    \mathbf{0}_{3} & \mathbf{0}_{3}
\end{bmatrix}
\vect{x}{}{}}_{f(\vect{x}{}{})}
+
\underbrace{\begin{bmatrix}
    \mathbf{0}_{3} \\
    \mathbf{I}_{3}
\end{bmatrix}}_{g(\vect{x}{}{})}
\vect{u}{}{},
\end{equation}
where $\vect{a}{WB}{W}$ is the linear acceleration. Here $f(\vect{x}{}{})$ and $g(\vect{x}{}{}) \vect{u}{}{}$ denote the state system dynamics vectors.

\begin{figure}
    \centering
    \includegraphics[width=\linewidth]{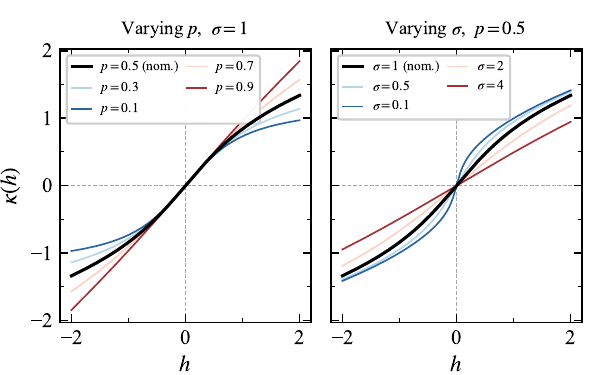}
    \caption{Visualization of the used kappa function with nominal values $\lambda=1$, $p=0.5$, $\sigma=1$.}
    \label{fig:kappa function}
\end{figure}

Considering the above, the method then represents the ``free-space'' set as $\mathcal{C}_{\vect{x}{}{}} = \bigcap_{i=1}^N  \{ \vect{x}{}{}: \| \vect{p}{WB}{W}-\bm{p}_{\mathtt{WO}_{i}}^\mathtt{W} \|^2 \ge \varepsilon \}$ with $\varepsilon > 0$ representing safety radius around each obstacle point $\bm{p}_{\mathtt{WO}_{i}}^\mathtt{W}$. A condensed set of $N$ points is obtained by means of the most recent, sparsified depth observation. This is then associated with equivalent scalar ``distance-squared'' functions $\mathbf{\nu}_{i,0}(\vect{x}{}{}) := \| \vect{p}{WB}{W}-\bm{p}_{\mathtt{WO}_{i}}^\mathtt{W} \|^2 - \varepsilon^2$ with the safety for obstacle $i$ being $\mathbf{\nu}_{i,0} \ge 0$. Note that there is no dependence on any consistent world frame as only the relative distances are used in the construction. Subsequently, we define higher-order CBF (HO-CBF) functions as: 

\begin{equation}
    \mathbf{\nu}_{i} := \mathfrak{L}_f \mathbf{\nu}_{i,0} - \varsigma(\mathbf{\nu}_{i,0}),
\end{equation}
where $\mathfrak{L}_f$ is the Lie derivative along the drift dynamics and $\varsigma$ is a tunable class $\mathcal{K}_\infty$ function of the form 
\begin{equation}
    \varsigma(h)=\lambda\cdot h\,(h^2+\sigma^2)^{(p-1)/2}.
\end{equation}
The used function $\varsigma$ with parameters $\lambda>0, \sigma>0, p>0$ is displayed in Fig. \ref{fig:kappa function}. For $p<1$ the tuneable parameters allow for a more rapid convergence behavior than a for $p=1$.

We formulate the \ac{ccbf} as

\begin{equation}
    h(\vect{x}{}{}) := -\frac{\gamma}{\kappa}\log\left(\sum_{i=1}^N e^{-\kappa\,\tanh(\mathbf{\nu}_{i}(\vect{x}{}{})/\gamma)}\right),
\end{equation}
with saturation parameter $\gamma$ and temperature parameter $\kappa$.

To enforce invariance of the safe set $\mathcal{X}_\text{safe} = \{\vect{x}{}{} \mid h(\vect{x}{}{})\geq0 \}$, the condition for a \ac{rcbf} \cite{robust_cbf} reads

\begin{gather}\label{invariance}
    \mathfrak{L}_g h(\vect{x}{}{}) u \ge \vartheta(\vect{x}{}{}),\\
    \vartheta(\vect{x}{}{}) := - \mathfrak{L}_f h(\vect{x}{}{}) - \alpha h(\vect{x}{}{}) + \rho(||\mathfrak{L}_g h(\vect{x}{}{})||),
\end{gather}
for a scalar $\alpha \in \mathbb{R}^+$ and a \textit{robustness function} $\rho(y)=\rho_1 y+\rho_2 y^2$. For a valid \ac{ccbf} $h$, the satisfaction of condition \eqref{invariance} is sufficient to ensure forward invariance of the safe set $\mathcal{X}_\text{safe}$, even for inputs with unknown, additive disturbances $\vect{d}{}{}$ added to the undisturbed input $\vect{u}{\text{n}}{}$, e.g. $\vect{u}{}{} = \vect{u}{\text{n}}{} + \vect{d}{}{}$, accounting for tracking errors satisfying $||\vect{d}{}{}||\leq\inf_{y\in\mathbb{R}_{\ge0}}\frac{\rho(y)}{y}$ in the low-level controllers.
This scheme can thus be used to certify collision-free navigation in spite of imperfect tracking. %
We enforce the condition \eqref{invariance} for a nominal input $\vect{u}{\textrm{sp}}{}$  by means of a reactive safety filter \ac{qp} of the form

\begin{equation}\label{safety_qp}
\begin{split}
    \vect{u}{}{*} = & \argmin_{\vect{u}{}{} \in \mathbb{R}^3}
        \| \vect{u}{}{} - \vect{u}{\textrm{sp}}{} \|^2 \\
    & \textrm{s.t.}\quad \mathfrak{L}_{g}h(\vect{x}{}{})\vect{u}{}{}
        \ge \vartheta(\vect{x}{}{}).
\end{split}
\end{equation}
For the case of one single constraint, an analytical solution to \eqref{safety_qp} can be computed \cite{cbf_auto_vehicle} as

\begin{equation}
    {\vect{u}{}{*}} = {\vect{u}{\textrm{sp}}{}} + \max \left( 0, \eta(\vect{x}{}{}) \right) \mathfrak{L}_{g}h(\vect{x}{}{})^\top,
\end{equation}
where

\begin{equation}
\eta(\vect{x}{}{})=\begin{cases}
-\frac{\mathfrak{L}_{g}h(\vect{x}{}{}){\bm{u}_{\textrm{sp}}} - \vartheta(\vect{x}{}{})}{\|\mathfrak{L}_{g}h(\vect{x}{}{})\|^2} & \mathrm{if} \
    \mathfrak{L}_{g}h(\vect{x}{}{}) \neq \mathbf{0}\\
0 & \mathrm{if} \ \mathfrak{L}_{g}h(\vect{x}{}{}) = \mathbf{0}.
\end{cases}
\end{equation}

This computationally cheap and flexible scheme to reactively enforce collision avoidance of any higher level control policy. To further reduce chattering of the safety filter, we further apply \ac{ema} filtering to the output.

\subsubsection{Ablation Studies}\label{sec:ablation}
\begin{figure*}[htbp]
  \centering
  \subfloat[Procedurally generated test environments used in the ablation studies regarding the \navigation{}. Each environment is a rectangular corridor bounded above, below, and on the sides by walls, with the longitudinal direction left open for traversal. Spherical obstacles of \SI{1}{\meter} diameter are placed via Poisson-disc sampling with a parametric minimum separation $r_{\textrm{sep}}$ between obstacle centers, with lower values representing denser environments. The robot dimensions are visualized in each environment.\label{fig:world-densities}]{\includegraphics[width=0.95\textwidth]{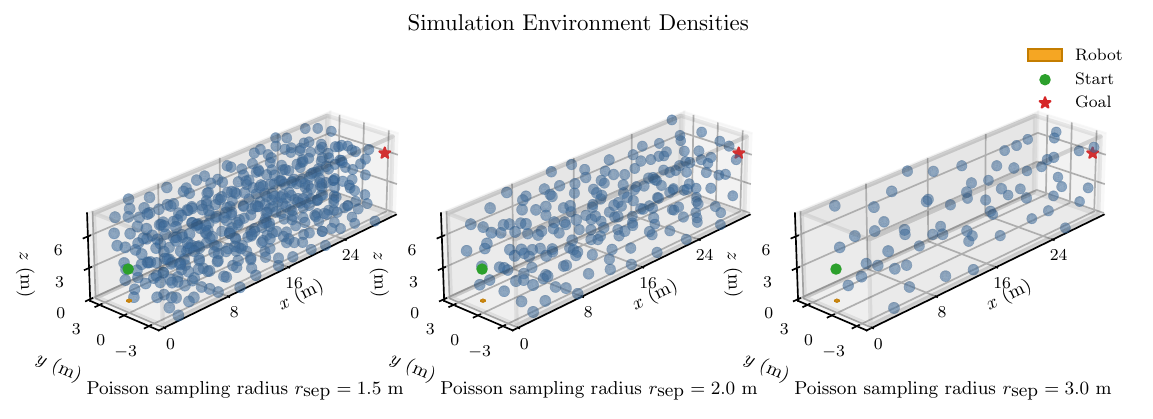}} \\
      \subfloat[Outcome rates across density (horizontal axis, $r_\textrm{sep}$) and command mismatch (columns, $\tau_d$). Rows show \textit{success}, \textit{crash}, and \textit{stagnation} rates over $20$ runs for each case. Solid lines: \ac{ccbf}-filtered; dashed: unfiltered. Blue: \ac{exteroceptivedrl}; green: \ac{neuralmpc}.\label{fig:ablations-combined}]{\includegraphics[width=0.95\textwidth]{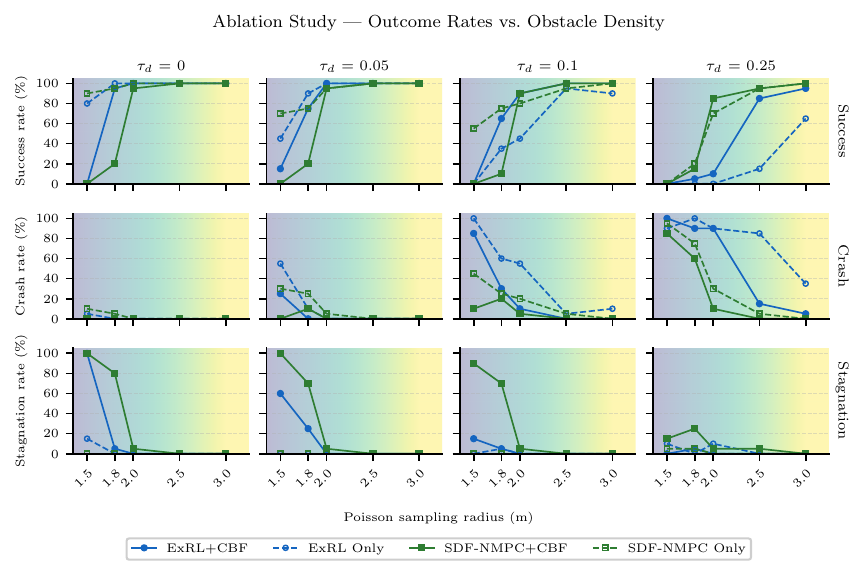}}
  \caption{Ablation study comparing \ac{exteroceptivedrl} and \ac{neuralmpc} navigation policies, with and without the \ac{ccbf} last-resort safety filter, across obstacle densities (\ref{fig:world-densities}) and a sweep of the time constant $\tau_d$ (\ref{fig:ablations-combined}). $\tau_d \leq \SI{0.10}{\second}$ approximately corresponds to the realistic operating regime; $\tau_d = \SI{0.25}{\second}$ is unrealistic. Since \textit{success} and \textit{stagnation} both correspond to collision-free behaviour, the crash rate is the primary safety metric.}
\end{figure*}

A set of evaluation studies are conducted using a simulated quadrotor with mass \SI{2.10}{\kilogram} and dimensions $0.3 \times 0.3 \times 0.1\si{\meter}$ in Gazebo~\cite{koenig2004gazebosim}. The robot is equipped with an acceleration and yaw-rate tracking controller and a hemispherical dome LiDAR sensor modeled after the RoboSense Airy (\ac{fov} \SI{180}{\degree}$\times$\SI{90}{\degree}). The performance of the \ac{exteroceptivedrl} and the \ac{neuralmpc} methods is compared both with and without the \ac{ccbf}-based safety filter, across environments of varying density, and under different levels of command mismatch induced by an independent first-order low-pass filter with time constant $\tau_d$ on each acceleration and yaw-rate command dimension. 
The slower dynamics introduced by the low-pass filter serve to practically emulate the effects of the closed-loop attitude response of the system, which presents non-instantaneous reference tracking and at times imperfect disturbance rejection. This is of particular interest especially as the actual time constant of the attitude subsystem differs between robots and delayed response poses a challenge to navigation in tight spaces. This allows us to characterize the failure modes of the navigation policies induced by increasingly disturbed system behavior.

Test environments, illustrated in Figure~\ref{fig:world-densities}, are procedurally generated as rectangular corridors of width and height \SI{8}{\meter} each, populated with spherical obstacles of \SI{1}{\meter} diameter placed using Poisson-disc sampling, guaranteeing a parametric minimum separation $r_{\textrm{sep}} \in \{ 1.5, 1.8, 2.0, 2.5, 3.0\}$~\textrm{m} between obstacle centers, i.e., surface-to-surface gaps of $\{ 0.5, 0.8, 1.0, 1.5, 2.0\}$~\textrm{m}. The corridor is bounded above, below and on the sides by walls, leaving the longitudinal direction open for traversal. The robot is initialized at the start of the corridor and tasked with following a path through the corridor to the other side.

For each environment, and for each value of the low-pass time constant $\tau_d$, $20$ independent runs are performed with each of the four configurations and the outcomes are visualized in Figure~\ref{fig:ablations-combined}. Every run terminates in one of three mutually exclusive states: \textit{success}, in which the robot reaches the goal; \textit{stagnation}, in which the robot halts before the goal but remains collision-free; or \textit{crash}, in which a collision with the environment is detected. The figure reports the rate of each outcome as a function of $r_\textrm{sep}$, with rows corresponding to the three outcome categories and columns to the four values of $\tau_d$. Smaller $r_\textrm{sep}$ values correspond to denser, more constrained environments, while larger $\tau_d$ values correspond to a more severe command mismatch. Solid lines indicate the \ac{ccbf}-filtered configurations and dashed lines the baselines without this last-resort filter, with colour distinguishing the upstream policy (\ac{exteroceptivedrl} in blue, \ac{neuralmpc} in green). Each marker aggregates the $20$ runs for the corresponding combination of parameters.

Two sets of comparisons are performed. Among the values of $\tau_d$ tested,
$\tau_d \leq \SI{0.10}{\second}$ corresponds to the realistic operating
regime that the studies are intended to characterize; $\tau_d =
\SI{0.25}{\second}$ is included as a deliberate stress test, beyond
realistic deployment conditions, in order to probe the limits of each
configuration and expose the regime in which the safety ceases to
hold. Because both \textit{success} and \textit{stagnation} outcomes
correspond to collision-free behaviour, the \textit{crash} rate is treated as the
primary safety metric throughout, while the success rate is reported as a
secondary but important measure of task completion.

The first comparison contrasts the two unfiltered baselines,
\ac{exteroceptivedrl} only and \ac{neuralmpc} only, which differ markedly
in how they degrade under increasing $\tau_d$. At $\tau_d = 0$ both reach the
goal on $80$--$\SI{100}{\percent}$ of runs across all densities, with the
\ac{exteroceptivedrl} policy crashing on \SI{5}{\percent} of the densest
layouts and \ac{neuralmpc} on \SI{10}{\percent}. As $\tau_d$ grows, the two
baselines diverge sharply. \ac{neuralmpc} degrades gracefully and
primarily along the density axis $r_\textrm{sep}$. At $\tau_d = \SI{0.25}{\second}$ it still
succeeds on \SI{100}{\percent} of runs at $r_\textrm{sep} = \SI{3.0}{\meter}$
and on \SI{95}{\percent} at $r_\textrm{sep} = \SI{2.5}{\meter}$, with
crashes concentrated in the dense environment setup. \ac{exteroceptivedrl},
in contrast, degrades globally: at the same $\tau_d$ it crashes on
$85$--$\SI{100}{\percent}$ of runs across the three densest settings and
even at $r_\textrm{sep} = \SI{3.0}{\meter}$ retains only \SI{65}{\percent}
success. This asymmetry may reflect \ac{neuralmpc}'s robustness against the induced mismatch, while the learned
policy's reliance on a specific dynamics model employed during training and the subsequent degradation wherever it departs from that model.

The second comparison contrasts each baseline against its
\ac{ccbf}-augmented counterpart. At $\tau_d = 0$, the safety filter
eliminates collisions entirely and converts results into stagnations rather than goal completions, preserving safety at the cost of task completion. The filter's protective effect persists across the
realistic regime. At $\tau_d = \SI{0.10}{\second}$, \ac{neuralmpc}+\ac{ccbf}
crashes on at most \SI{20}{\percent} of runs in any environment, always lower than \ac{neuralmpc} alone, and
\ac{exteroceptivedrl}+\ac{ccbf} crashes on $10$--$\SI{85}{\percent}$ in
the dense band, again lower than
\ac{exteroceptivedrl} alone. This indicates a substantial reduction in crash rate with the inclusion of the \ac{ccbf}. Under the unrealistic $\tau_d = \SI{0.25}{\second}$ conditions, the filter
loses authority in the densest environments and both augmented pipelines
crash on $85$--$\SI{100}{\percent}$ of runs at $r_\textrm{sep} =
\SI{1.5}{\meter}$, locating the failure boundary that the test is
designed to expose. The augmented versions inherit the characteristics of
their upstream policy: \ac{exteroceptivedrl}+\ac{ccbf} crashes earlier
and across a wider density range than \ac{neuralmpc}+\ac{ccbf} at every
$\tau_d > 0$, indicating that the \ac{ccbf} compensates for command
distortion but not for the upstream policy's own degradation under that
distortion. This analysis shows that imperfect control can occur in demanding environments and conditions, highlighting the need for navigation strategies with multiple layers of safety checking and avoidance. The \ac{ua} offers a configurable methodological plurality to support demanding deployments.

\subsubsection{Exteroceptive Overwrite}

When the map-based safety of \planning{} or the specific exteroceptive navigation methods are not desirable for, or compatible with, a given robot, the \ac{ua} allows the use of a conventional state-feedback controller. Examples include the Linear MPC methods in~\cite{flatness_linear_mpc,kamel_model_2017}, the underlying MPC in \acs{neuralmpc} with disabled collision constraints, or a solution for any other particular robot or through existing autopilot (e.g., PX4).

\subsection{Low-level Interfaces}\label{sec:low_level_interface}

The \ac{ua} interfaces diverse aerial and ground robots as follows. Although extendable in principle, the currently provided open-source interfaces are as follows:

\paragraph{Aerial Robots - Full Stack:} When the full stack with multi-layered safety is considered for flying systems, we command acceleration setpoints directly to compatible autopilots and low-level controllers such as PX4- and ArduPilot-based flight controllers. 

\paragraph{Aerial Robots - Without Multi-layered Safety:} If the multi-layered safe navigation is not necessary for flying systems, we provide options for commanding waypoints or 3D accelerations to existing controllers. Waypoints are straightforward for all systems that offer such control. However, this does not deliver the full stack functionality and collision avoidance is ensured only at the map/planning level.

\paragraph{Ground Robots:} For legged systems and other ground robots, we interface the \planning{} with a custom PID tracking controller to output velocity commands to follow the path given by the \planning{}. Velocity-setpoint commands are provided to the platforms for them to track. Those velocity references may be passed through other safety-mechanisms in case those are provided onboard the platforms.
    
Direct support for widely-adopted low-level control or autopilot interfaces is available out-of-the-box through MAVROS, or more broadly through the ROS framework. As the stack is structured in a Dockerized format, we support both ROS 1 and ROS 2-based robots out of the box.

\begin{table*}[h]
  \centering
  \caption{Overview of the different experiments composing the evaluation.}
  \label{tab:evaluation:overview}
  \newcommand{\vmone}{-0.675ex}
\newcommand{\vmtwo}{-1.35ex}
\newcommand{\vmthree}{-2.025ex}
\newcommand{\vmfour}{-2.7ex}
\newcommand{\na}{--}
\begin{threeparttable}
  \begin{tabular}{
    m{\widthof{Full-Stack}}
    l
    l
    l
    m{\widthof{Perceptual Challenge(s)}}
    m{\widthof{Geometric Challenge(s)}}
  }
    \toprule
    Purpose &Environment  &Robot  &\ac{ua} Module(s)\tnote{1}  &Perceptual Challenge(s) &Geometric Challenge(s)\\
    \midrule
    \multirow{4}{=}[\vmthree]{\ac{slam} Validation}
      &Fyllingsdal &\acs{ar1} &S &Self-similarity &\na\\ \cmidrule(l){2-6}
      &Runehamar &\acs{ar1} &S &Low-light &\na\\ \cmidrule(l){2-6}
      &Frozen Lake &\acs{ar1}   &S   &Self-Similarity   &\na\\ \cmidrule(l){2-6}
      &Campus &Handheld &S+V &Visual obscurants &\na\\
    \midrule
    \multirow{5}{=}[\vmone]{Safety Validation}
      &\multirow{3}{*}{Forest}
        &\acs{ar2} &S\textsubscript{LRI}+R+C &\multirow{3}{=}{Thin features}  &\multirow{3}{=}{Thin obstacles and cluttered}\\
      & &\acs{ar2} &S\textsubscript{LRI}+N\textsubscript{S}+C &  &\\
      & &\acs{ar2} &S\textsubscript{LRI}+N\textsubscript{U}+C &  &\\ \cmidrule(l){2-6}
      &\multirow{2}{*}{Campus}
        &\acs{ar2} &S\textsubscript{LRI}+P$_{\textrm{E}}$+R+C &\na  &\multirow{2}{=}{Moving obstacles}\\
      & &\acs{ar2} &S\textsubscript{LRI}+P$_{\textrm{E}}$+N\textsubscript{S}+C &\na &\\
    \midrule
    \multirow{7}{=}[\vmfour]{Full-Stack Validation}
      &\multirow{2}{*}{Forest} 
        &\acs{ar2} &S\textsubscript{LRI}+P$_{\textrm{E}}$+R+C &\multirow{2}{=}{Thin features}  &\multirow{2}{=}{Thin obstacles and cluttered}\\
      & &\acs{ar2} &S\textsubscript{LRI}+P$_{\textrm{E}}$+N\textsubscript{S}+C & &\\ \cmidrule(l){2-6}
      &\multirow{3}{*}{Mine} 
        &\acs{ar2} &S\textsubscript{LRI}+P$_{\textrm{E}}$+R+C &\multirow{3}{=}{Low-light} &\multirow{3}{=}{Narrow and multiple branches}\\
      & &\acs{ar2} &S\textsubscript{LRI}+P$_{\textrm{E}}$+N\textsubscript{S}+C & &\\ 
      & &\acs{gr1} &S\textsubscript{LI}+P$_{\textrm{E}}$ & &\\ \cmidrule(l){2-6}
      &Ship &\acs{ar2} &S\textsubscript{LI}+P$_{\textrm{E}}$+P$_{\textrm{I}}$ &Low-light &\na\\ \cmidrule(l){2-6}
      &Campus &\acs{gr1} &S\textsubscript{LRI}+P$_{\textrm{E}}$+V &\na &Narrow, varying size, and multiple branches\\
    \bottomrule
  \end{tabular}
  \begin{tablenotes}
  \item[1] Which modules of the \ac{ua} were used, including \ac{slam} (denoted by S, with a subscript denoting the configuration used in autonomous missions, where L: LiDAR, R: Radar, I: IMU), \ac{vlm} (denoted by V), Planning (denoted by P, while P$_{\textrm{E}}$ refers to exploration and P$_{\textrm{I}}$ to inspection), \ac{exteroceptivedrl} (denoted by R), \ac{neuralmpc} (denoted by N\textsubscript{S}), \ac{neuralmpc} without \ac{sdf} constraints (denoted by N\textsubscript{U}), and \ac{ccbf} (denoted by C).
  \end{tablenotes}
\end{threeparttable}

\end{table*}

\section{Evaluation Studies}
\label{sec:evaluation}

To validate the \ac{ua}, a comprehensive set of evaluation studies was conducted including (a) evaluation of the performance, accuracy and overall resilience of the \perception~in diverse environments and conditions, (b) evaluation of the performance and safety-inducing behaviors of the \navigation, as well as (c) full-stack results requiring the orchestrated operation of the \texttt{perception}, \texttt{planning} and \texttt{navigation modules}. Studies were conducted using both aerial and ground robots, alongside some handheld experiments (as part of the evaluation regime of the \perception). An overview of these experiments is shown in \cref{tab:evaluation:overview} and key parameters are listed in \cref{tab:evaluation:params}. All presented field experiments are included in the video files of the submission and can be found at \url{https://youtube.com/playlist?list=PLu70ME0whad9KCs5PHx-35-qpyK14yZYi&si=E-QJCTTyfhfFUqFo}.

\begin{table}[h!]
  \centering
  \caption{Key parameters used in the evaluations.} 
  \label{tab:evaluation:params}
  \newcommand{\unitless}{--}
\setlength{\tabcolsep}{4.5pt}
\newcommand{\vone}{-0.675ex}
\newcommand{\vtwo}{-1.35ex}
\begin{threeparttable}
\begin{tabular}{
    m{\widthof{SDF-NMPC}}
    m{\widthof{D-Opt. (nom.)}}
    c
    c
}
  \toprule
  Module    &Parameter &Value   &Unit\tnote{1}\\
  \midrule
  \multirow{7}{*}[\vtwo]{\ac{slam}}
    &$\eta_p$                      &0.2 &\si{\meter}\\
    &$n_p$                         &20 &\unitless\\
    &$\sigma_{\epsilon_{\mathcal{L}}}$ (nom.) & 0.07 & m \\
    &$\sigma_{\epsilon_{\mathcal{L}}}$ (lake) & 0.18 & m \\
    &Radar \ac{fov}             & $120 \times 120$  &\si{\degree}\\
    \cmidrule(l){2-4}
    &\multirow{2}{=}{Map update thresholds}
        &1   &\si{\meter}\\
    &   &10   &\si{\degree}\\
    \cmidrule(l){2-4}
    &D-Opt. (nom.) &0.02   &\unitless\\
    &D-Opt. (fog) &0.001   &\unitless\\
  \midrule
  \multirow{10}{*}{Planner}
    &$v_m$                         &0.2 &\si{\meter}\\
    &$v_h$                         &0.2 &\si{\meter}\\
    &$\bm{b}_R$ (\acs{ar2})               & $0.8 \times 0.8 \times 0.8$  &\si{\meter}\\
    &$\bm{b}_R$ (\acs{gr1})               & $1.0 \times 1.0 \times 0.4$  &\si{\meter}\\
    &$\bm{b}_L$ (\acs{ar2})               & $30.0 \times 30.0 \times 4.0$  &\si{\meter}\\
    &$\bm{b}_L$ (\acs{gr1})               & $40.0 \times 40.0 \times 4.0$  &\si{\meter}\\
    &\ac{fov} of $\depthsensor$    & $180 \times 90$    &\si{\degree}\\
    &\ac{fov} of $\camerasensor$   & $118 \times 94$    &\si{\degree}\\
    &$d_{view}$                    &1.25                &\si{\meter}\\
    &$d_{path}$                    &2.0                 &\si{\meter}\\
  \midrule
  \multirow{3}{*}{SDF-NMPC}
    &$T_{SDF}$                     &1.0    &\si{\meter}\\
    &$d_{\max}$                    &5.0   &\si{\meter}\\
    &$r$                           &0.5    &\si{\meter}\\
    \midrule
    \multirow{4}{*}{\ac{exteroceptivedrl}}
    &$u_n^x$    &2.0   &\si{\metre\per\second\squared}\\
    &$u_n^y$    &2.0   &\si{\metre\per\second\squared}\\
    &$u_n^z$    &1.5   &\si{\metre\per\second\squared}\\
    &$u_n^{\dot{\psi}}$            &$\frac{\pi}{3}$ &\si{\radian\per\second}\\
  \midrule
  \multirow{10}{*}{\ac{ccbf}} 
    &$\lambda$                     &3.0     &\unitless\\
    &$\sigma$                      &0.3     &\unitless\\
    &$p$                           &0.8     &\unitless\\
    &$\rho_1$                      &0.5     &\unitless\\
    &$\rho_2$                      &0.5     &\unitless\\
    &$\alpha$                      &4.0     &\unitless\\
    &$\kappa$                      &80.0    &\unitless\\
    &$\gamma$                      &40.0    &\unitless\\
    &$\varepsilon$                 &0.3     &\si{\meter}\\
    &$N$                           &256     &\si{\points}\\
  \bottomrule
\end{tabular}
\begin{tablenotes}
    \item[1] Unitless quantities denoted by \unitless.
\end{tablenotes}
\end{threeparttable}

\end{table}

\subsection{Verified Robot Morphologies}

In this paper, we evaluate the \ac{ua} on two multirotor robot configurations and a legged robot. In all robot missions, the \ac{ua} runs fully onboard in real-time, with module configurations as detailed in Table~\ref{tab:evaluation:overview}. The released open-source code involves simulation examples with additional morphologies such as ground rovers and helicopters. Verifying on diverse robots -- while further expanding the morphologies we support -- is aligned with our strategic goal to provide a generalist autonomy stack across broad morphological categories.

\subsubsection{Multirotors}

\paragraph{\acf{ar1}:} The first aerial robot, referred to as \ac{ar1}, is an improved version of the RMF-Owl~\cite{petris_rmf-owl_2022}. \ac{ar1} integrates a Khadas Vim4 computer and a sensing suite comprising of an Ouster OS0-128 LiDAR (\SI{10}{\hertz}), a Flir Blackfly S 0.4~MP color camera (\SI{20}{\hertz} normally, \SI{25}{\hertz} in Fyllingsdal), a Texas Instruments \si{\milli\meter}Wave IWR6843AOP radar sensor (\SI{10}{\hertz} normally, \SI{25}{\hertz} in Fyllingsdal), and a VectorNav VN-100 \ac{imu} (\SI{200}{\hertz}). The sensors and the onboard computer are time synchronized using a separate microcontroller as described in~\cite{nissov2025simultaneoustriggeringsynchronizationsensors}.

\paragraph{\acf{ar2}:} \ac{ar2} is a collision-tolerant quadrotor designed for autonomous operation in \ac{gnss}-denied confined environments. \acs{ar2} features a lightweight protective frame made from carbon-foam sandwich measuring approximately \qtyproduct{0.52x0.52x0.24}{\meter} (L $\times$ W $\times$ H), weighing \SI{2.3}{\kilo\gram}. The robot carries the UniPilot~\cite{unipilot} sensing and computing payload on which the entire \ac{ua} runs onboard. The module features an NVIDIA Jetson Orin NX as the compute module and a multi-modal sensing suite including a RoboSense Airy dome LiDAR (\SI{10}{\hertz}), $3\times$ MIPI Vision Components IMX296 color cameras (\SI{20}{\hertz}), a D3 Embedded RS-6843AOPU \ac{fmcw} radar (\SI{10}{\hertz}), and a VectorNav VN-100 \ac{imu} (\SI{200}{\hertz}). 
The onboard sensors are time-synchronized following the work in~\cite{unipilot}
\acs{ar2} further integrates a Pixracer Pro PX4-based flight control to track the acceleration commands given by the \ac{ua}.

\subsubsection{Legged Robots} 
\paragraph{\acf{gr1}}
On the ground, the \ac{ua} is evaluated using the ANYmal D legged robot -- hereafter referred to as \ac{gr1} -- with dimensions of \qtyproduct{0.93x0.53x0.80}{\meter} (L $\times$ W $\times$ H), and a mass of \SI{50}{\kilo\gram}. The stock standard robot features a Velodyne VLP-16 LiDAR and $6\times$ depth cameras for sensing and two 8th-generation Intel Core i7 CPUs (6 cores each) for compute. However, \ac{gr1} is also equipped with the UniPilot module which runs the \ac{ua} for all evaluations. This UniPilot carries the same sensing payload as in the \ac{ar2} platform and is interfaced with the onboard Intel computers which then run the locomotion and velocity tracking controllers. All the stack runs on the onboard UniPilot, considering the data of this module.

\subsection{Perception Module Evaluation}

We first evaluate the performance and resilience of the stack's \perception, focused around its multi-modal \ac{slam} functionalities. We subsequently evaluate downstream functionalities for \ac{vlm}-based scene reasoning. 

\begin{figure*}[h]
	\centering
	\includegraphics[clip, trim = 0cm 0cm 0cm 0cm, width=1\linewidth]{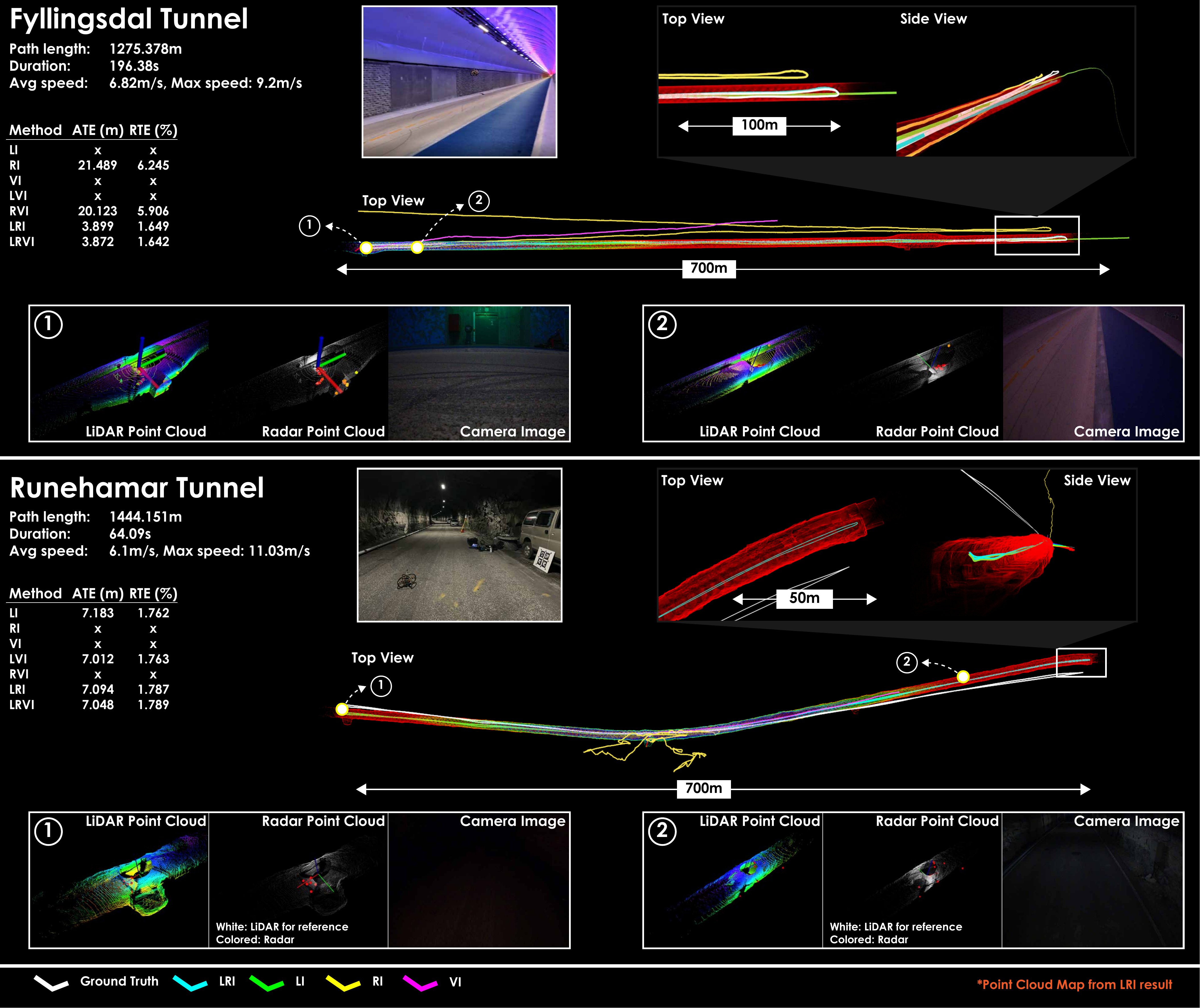}
	\caption{Overview of the \ac{slam} performance in the two tunnel environments. The Fyllingsdal tunnel presents an environment with geometric self-similarity, thus causing divergence in LiDAR-geometry-based methods. Multi-modal fusion demonstrates increased performance, by replacing the missing observability of the LiDAR measurements with information from either radar or vision. The Runehamar tunnel contains sections of near geometric self-similarity, too short to cause widespread divergence. This environment also has sections with low-lighting and poor radar returns, complicating state estimation. Again the multi-modal fusion demonstrates robust performance.}
    \vspace{-2ex}
	\label{fig:eval_slam_tunnels}
\end{figure*}

\begin{table*}[h]
  \centering
  \caption{Metrics for the multi-modal \ac{slam} evaluation, including \acs{ate} [\si{\meter}] and \acs{rte} [\si{\percent}].}
  \label{tab:evaluation:slam}
  \newcommand{\vmove}{-1.35ex}
\begin{threeparttable}
\begin{tabular}{llcccc}
  \toprule
  &  &Fyllingsdal Tunnel &Runehamar Tunnel & Frozen Lake & Campus Fog \\
  \midrule
  &Length [\si{\meter}]  &$1275.378$ &$1444.151$ & $826.054$ & $669.609$\\ \cmidrule{1-6}
  \multirow{13}{*}[\vmove]{\rotatebox{90}{\acs{ate} [\si{\meter}] / \acs{rte} [\si{\percent}]}}
    &FAST-LIO2  & $\color{red}\times$ & $6.578$ / $1.891$ & $\color{red}\times$ & $\color{red}\times$                                                                                      \\
    &FAST-LIVO2 & $4.930$ / $2.852$ & $7.881$ / $1.805$ & {\color{red}$\times$} & $\color{red}\times$  \\
    &GaRLIO &{\color{red}$-$}  &{\color{red}$\times$} & {\color{red}$\times$} & {\color{red}$\times$}\\
    &AF-RLIO & $\color{red}\times$ & $\color{red}\times$ & $\color{red}\times$ & $\color{red}\times$                                                              \\
    &ROVIO & $42.851$ / $20.295$ & $62.375$ / $29.806$ & $3.112$ / $7.177$ & $\color{red}\times$                                                                                \\
    &OpenVINS & $13.710$ / $6.271$ & $13.337$ / $5.443$ & $1.985$ / $3.700$ & $\color{red}\times$                                                                              \\
    \cmidrule{2-6}
    &\textbf{Ours - LI}  & $\color{red}\times$ & $7.183$ / $1.762$ & $18.661$ / $7.840$ & $\color{red}\times$                                                                               \\
    &\textbf{Ours - RI}  & $21.489$ / $6.245$ & $\color{red}\times$ & $10.189$ / $30.868$ & $14.075$ / $6.189$                                                                               \\
    &\textbf{Ours - VI} & $\color{red}\times$ & $\color{red}\times$ & $7.728$ / $7.989$ & $\color{red}\times$                                                                                                  \\
    &\textbf{Ours - LVI} & $\color{red}\times$ & $7.012$ / $1.763$ & $10.598$ / $4.633$ & $\color{red}\times$                                                                                               \\
    &\textbf{Ours - RVI} & $20.123$ / $5.906$ & $\color{red}\times$ & $9.141$ / $7.589$ & $17.737$ / $6.351$                                                                                               \\ \cmidrule{2-6}
    &\textbf{Ours - LRI} & $3.899$ / $1.649$ & $7.094$ / $1.787$ & $10.562$ / $6.900$ & $7.324$ / $4.240$                                                                               \\
    &\textbf{Ours - LRVI} & $3.872$ / $1.642$ & $7.048$ / $1.789$ & $10.030$ / $8.609$ & $8.345$ / $4.231$                                                                                              \\

  \bottomrule
\end{tabular}
\begin{tablenotes}
    \item Method failure due to ATE > $5\%$ is indicated by {\color{red}$\times$} and due to inability to generalize to sensor configuration (radar being at $25Hz$) is indicated by {\color{red}$-$}.
    \item For verifiability and reproducibility, the full implementation and the dataset involved in these studies are openly released.
\end{tablenotes}
\end{threeparttable}

\end{table*}

\begin{figure}[h]
	\centering
	\includegraphics[clip, trim = 0cm 0cm 0cm 0cm, width=1\columnwidth]{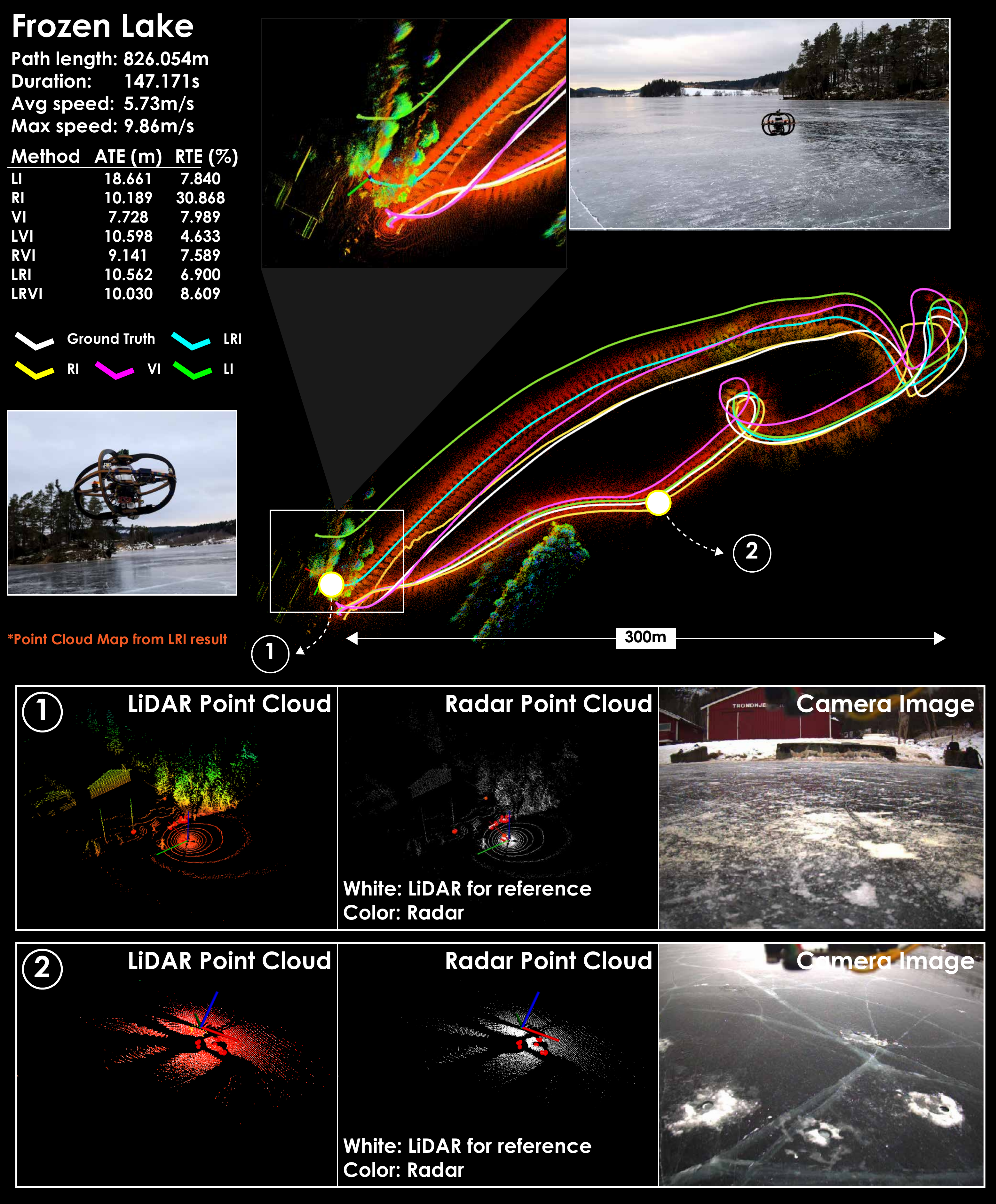}
	\caption{Overview of the \ac{slam} performance in the Frozen Lake environment, comparing different uni-modal approaches with the multi-modal LRI configuration. This environment presents difficulty for methods relying only on LiDAR or radar. For the former, the planar geometry of the environment can result in lacking observability in lateral position and yaw. For the radar, the limited number of returns results in increased drift.}
    \vspace{-5ex}
	\label{fig:eval_slam_lake}
\end{figure}

\begin{figure*}
    \centering
    \includegraphics[width=\linewidth]{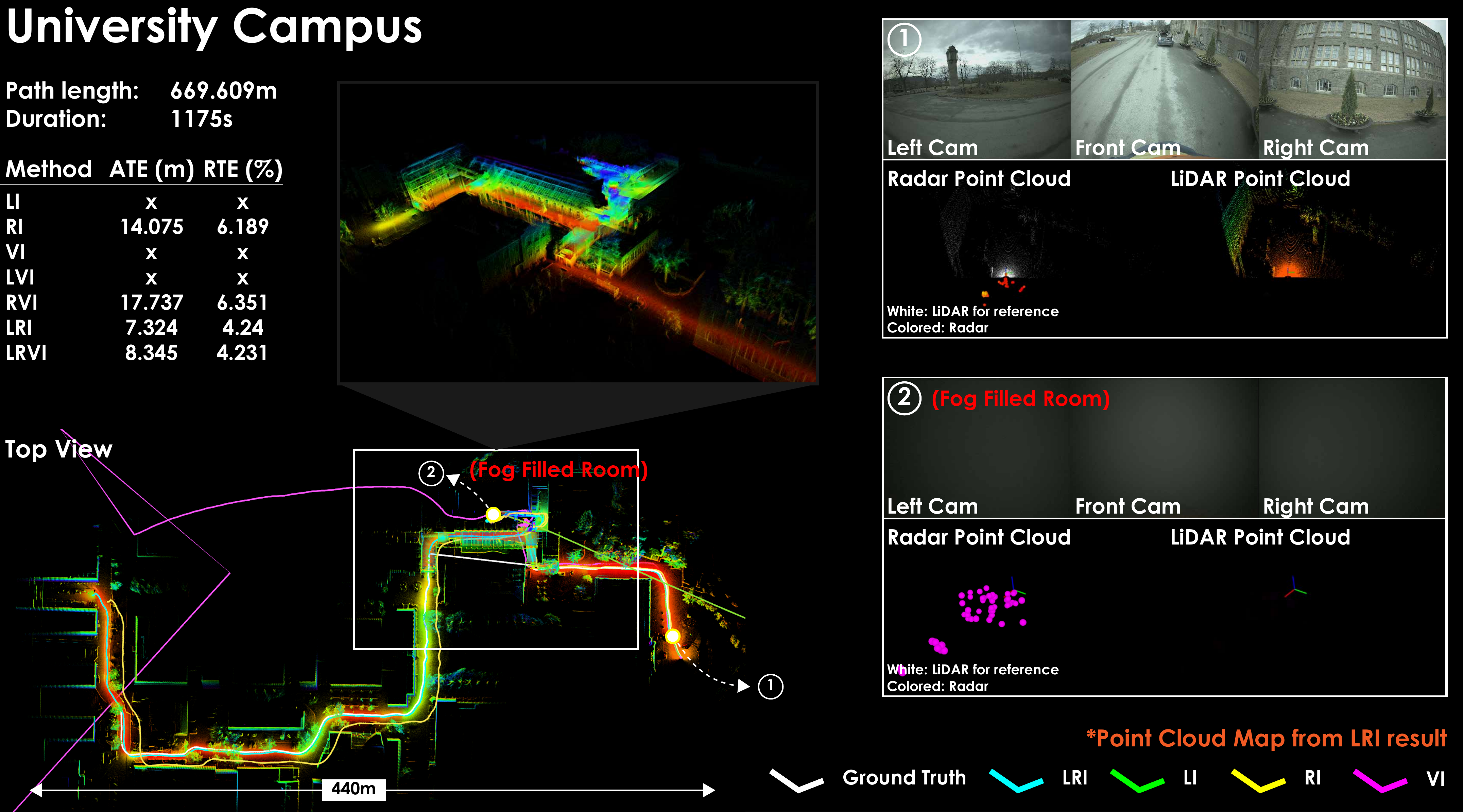}
    \caption{Overview of the \ac{slam} performance of the experiment in the campus environment with the handheld module. The experiment has the trajectory passing through a room filled with dense fog, causing large increase of noise present in the LiDAR and vision measurements, whereas the radar remains largely unaffected. The multi-modal fusion retains the accuracy of LiDAR-based methods in nominal conditions alongside the robustness of radar through degraded environments.}
    \vspace{-3ex}
	\label{fig:eval_slam_campus_fog}
\end{figure*}

\begin{figure*}[h]
	\centering
	\includegraphics[clip, trim = 0cm 0cm 0cm 0cm, width=0.9\linewidth]{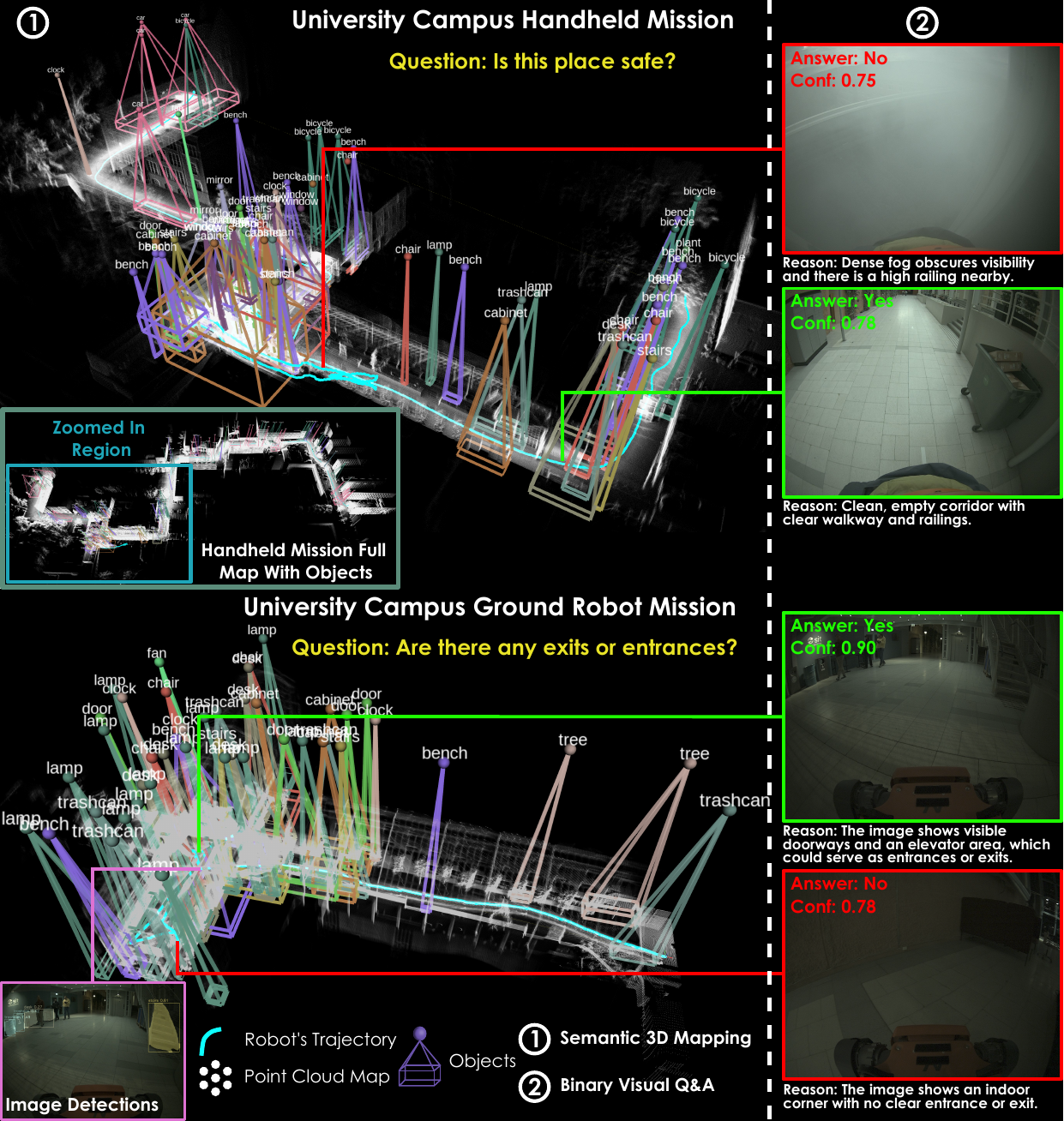}
	\caption{Qualitative results of the proposed \ac{vlm} reasoning system. Left: semantic 3D mapping with open-vocabulary object detections fused into a voxel grid, showing labeled objects and the robot trajectory over time. Right: binary visual question-answering examples, where the model provides ``Yes/No'' answers with confidence scores and explanations for high-level scene understanding tasks. The figure demonstrates the integration of spatial semantic mapping with high-level reasoning.}
    \vspace{-3ex}
	\label{fig:scene_reasoning}
\end{figure*}

\begin{figure*}[h]
	\centering
	\includegraphics[clip, trim = 0cm 0cm 0cm 0cm, width=1\linewidth]{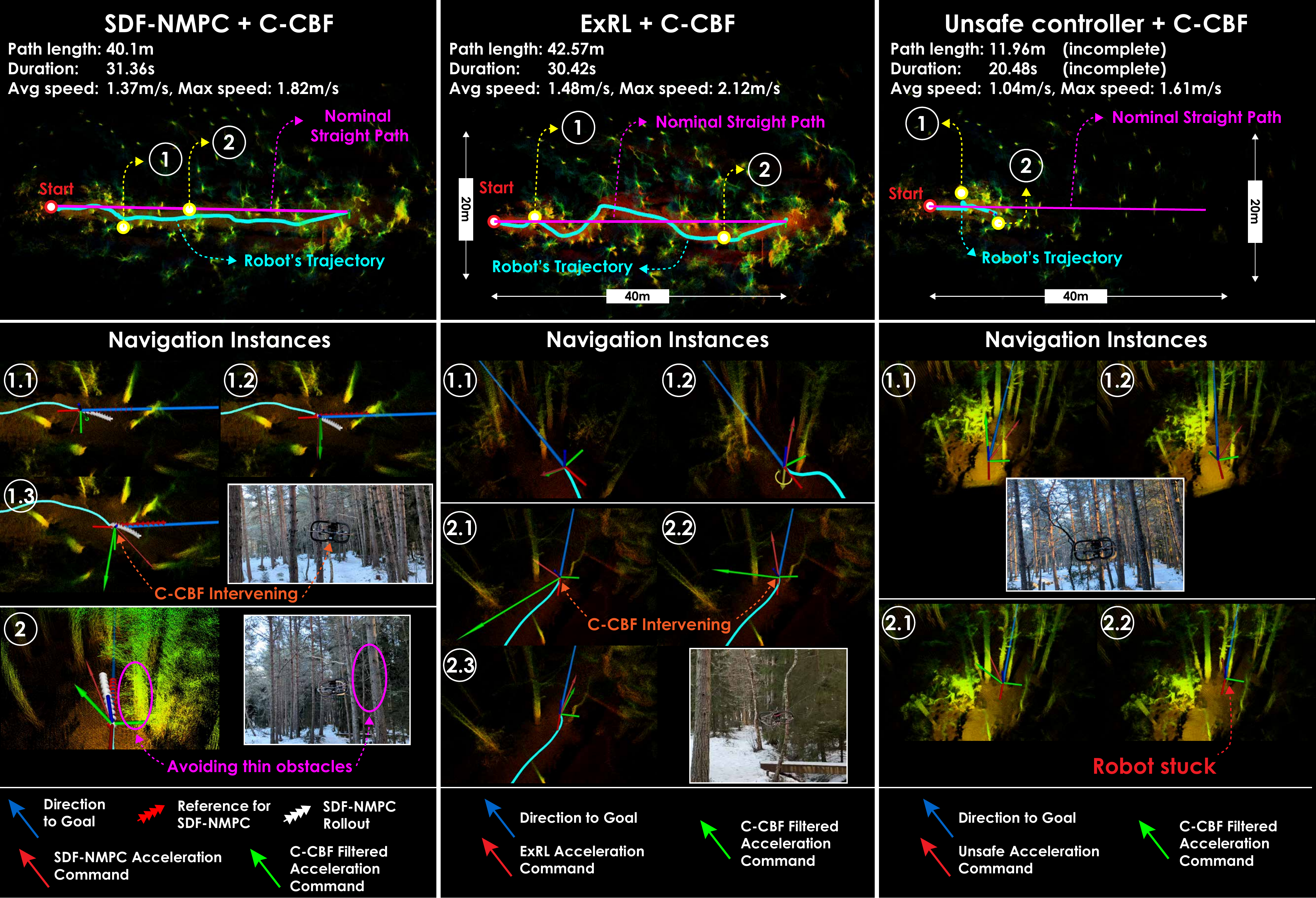}
	\caption{Navigation module evaluation is performed in the forest using \ac{ar2} with \ac{neuralmpc} + \ac{ccbf}, \ac{exteroceptivedrl} + \ac{ccbf}, and \ac{ccbf} paired with an unsafe policy (\ac{neuralmpc} but with its collision-avoidance constraints disabled) respectively. The robot was tasked to navigate to a waypoint \SI{38}{\meter} in front of it, with a reference path going through trees and branches. Individual experiments are shown, with specific instances highlighted. Both the \ac{neuralmpc} + \ac{ccbf} and the \ac{exteroceptivedrl} + \ac{ccbf} methods successfully navigated to the goal location, avoiding obstacles in the path. The \ac{ccbf} paired with an unsafe policy does not reach the goal and remains stuck, but remaining safe at all times. The full map is shown with specific instances highlighted.}
    \vspace{-3ex}
	\label{fig:eval_forest_safety}
\end{figure*}

\subsubsection{Multi-Modal SLAM}\label{subsec:multi-modal-slam}
To evaluate the multi-modal \ac{slam} solution, we assess its accuracy and robustness in perceptually-degraded conditions. Considering a set of diverse environments, specifically a bicycle tunnel (Fyllingsdal), a road tunnel (Runehamar), a frozen lake, and the \ac{ntnu} main campus, we present a modality-wise ablation as well as comparisons against state-of-the-art methods. The goal of this evaluation is to demonstrate the flexibility of the proposed \ac{slam} system, as well as the comparative advantages of the multi-modal fusion. These experiments are collected with ground truth, such that the estimation performance can be evaluated quantitatively. The same \ac{slam} system runs online to support the autonomous missions conducted in all the other experiments.
For these autonomous missions, the \ac{slam} module is operating in the \textbf{Ours - LRI} (fusing LiDAR, radar, \ac{imu}) configuration, except when noted otherwise (see \cref{tab:evaluation:overview}), as this has been found to be most robust across diverse environments and conditions.

The results of the evaluation are presented in Table~\ref{tab:evaluation:slam}. The table compares and ablates our multi-modal solution against state-of-the-art LiDAR-Inertial (FAST-LIO2~\cite{xu2022fastlio2}), Visual-Inertial (ROVIO~\cite{bloesch2017Rovio} and OpenVINS~\cite{geneva2020openvins}), LiDAR-Visual-Inertial (FAST-LIVO2~\cite{zheng2025fastlivo2}), and LiDAR-Radar-Inertial (GaRLIO~\cite{noh2025garlio} and AF-RLIO~\cite{qian2025afrlio}) works. GaRLIO and AF-RLIO are selected as a representative set of state-of-the-art methods for LiDAR-radar-inertial fusion which are both: likely to work with the small form-factor sensors considered in this work and with open-source implementations. The different permutations of our MIMOSA-X multi-modal \ac{slam} (LiDAR (\textbf{L}), Radar (\textbf{R}), Vision (\textbf{V}), and Inertial (\textbf{I})) are noted in the table as \textbf{Ours - XXXI}. As the \ac{imu} is an integral component of our method, permutations without \ac{imu} are omitted. The key parameters used in the evaluation of the \ac{slam} module are reported in 
\cref{tab:evaluation:params}. Note that all parameters are fixed across all tests, with the sole exception of the LiDAR point standard deviation (which is increased in the frozen lake environment), the \ac{imu} bias noise densities (decreased in the frozen lake environment), and the D-Optimality threshold for the vision fusion (which is different between environments with and without fog). In turn, the latter is one of the reasons that the robot experiments presented subsequently predominantly rely on the LRI solution. The table reports the \ac{ate} (in \SI{}{\meter}) and \ac{rte} (in \%), with \SI{10}{\meter} segment length, following~\cite{grupp2017evo}, calculated against the ground truth estimates. The ground truth for the tunnel trajectories (Fyllingsdal and Runehamar) is generated by fusing the tracking of a Leica GRZ101 mini-prism mounted on \ac{ar1} by a Leica MS60 MultiStation with the onboard \ac{imu} in an offline \ac{lm} optimization. In the campus and frozen lake datasets, where \ac{gnss} is available, ground truth is created using Pix4DMatic for a \ac{gnss}-augmented visual bundle adjustment optimization.

The \ac{slam} datasets were collected with (a) \ac{ar1} aerial robot and (b) a helmet-mounted modified version of the UniPilot~\cite{unipilot} module integrating a Hesai JT-128 LiDAR (\SI{10}{\hertz}), replacing the Robosense Airy, and a uRAD Industrial radar (\SI{10}{\hertz}), replacing the 3D Embedded RS-6843AOPU radar (both of which integrate the Texas Instruments IWR6843AOP chip). The full datasets including raw data and ground-truth are released to facilitate comparison and reproducibility.

\paragraph{Fyllingsdal Tunnel}
The \ac{ar1} aerial robot was manually piloted at an average speed of \SI{6.82}{\meter\per\second} (max speed \SI{9.2}{\meter\per\second}) in a \SI{\sim650}{\meter} section of the Fyllingsdal bicycle tunnel. The tunnel is composed primarily of long geometrically self-similar sections with sparse geometrically dissimilar rest areas. Ground truth for this environment is generated using the Leica-IMU fusion described above. Notably, the chirp configuration for the radar in this experiment is changed for better performance at high speeds. This results in a greater maximum Doppler as well as an increased measurement rate of \SI{25}{\hertz}. GaRLIO could not be evaluated on this sequence as the implementation requires the radar and LiDAR to be at the same rate. Due to the geometric self-similarity affecting the LiDAR optimization, FAST-LIO2 as well as \textbf{Ours - LI} failed. AF-RLIO fails similarly, due to the method relying on radar-based scan registration during periods of LiDAR degeneracy, which the small form-factor radar sensor used on the aerial platform is not well-suited for. The radar-inertial ablation (\textbf{Ours - RI}), while able to function, has a significant vertical and yaw drift due to the nature of the sensor. Furthermore, the high speeds created difficulties for vision-based methods, resulting in significant performance deterioration for ROVIO. As a result, \textbf{Ours - VI} fails, and \textbf{Ours - RVI} does not see large performance improvements over \textbf{Ours - RI}. OpenVINS performed better than ROVIO, \textbf{Ours - VI}, \textbf{Ours - RVI}, and \textbf{Ours - LVI}, however still not as well as the nominal multi-modal configuration \textbf{Ours - LRI} and \textbf{Ours - LRVI}. The proposed method, by fusing the multiple exteroceptive modalities, demonstrates performance robust both to the challenging conditions, but also to the asynchronous measurements, leading to the improved results of \textbf{Ours - LRI}, and \textbf{Ours - LRVI}. Similarly, FAST-LIVO2 is able to leverage the vision information to outperform most of the other methods, achieving performance similar to, but slightly worse than, \textbf{Ours - LRI} and \textbf{Ours - LRVI}. The estimated trajectories, ground truth and the accumulated LiDAR point cloud map from the \textbf{Ours - LRI} configuration is visualized in~\cref{fig:eval_slam_tunnels}.

\paragraph{Runehamar Tunnel}
The \ac{ar1} aerial robot was manually flown through a section of the Runehamar tunnel, for a total trajectory length of \SI{\sim1.4}{\kilo\meter}. Ground truth for this environment is generated using the Leica-\ac{imu} method described above. The tunnel has a rough interior, which allows for LiDAR-based methods (FAST-LIO2 and \textbf{Ours - LI}) to function well despite the otherwise minor geometric self-similarity. However, regions of the tunnel are not illuminated. Despite the aerial platform carrying onboard lighting, the scale of the environment is such that the images captured by the camera are dark and hence challenging for visual-based methods, leading to very poor performance from ROVIO (and hence \textbf{Ours - VI}). OpenVINS, by doing a histogram equalization of the image is able to better extract features from the darkest regions and as a result, demonstrates improved performance over ROVIO, however, still worse than the LiDAR-based fusions. Impacted by the poor visual measurement quality, FAST-LIVO2 performs slightly worse than FAST-LIO2. Radar measurement quality in this environment was also quite poor, in part due to the speed of the trajectory and in part due to low number of reflections. As a result \textbf{Ours - RI}, \textbf{Ours - RVI}, and the baseline radar-fusion methods (GaRLIO and AF-RLIO) do not perform well either. However, in combination with the LiDAR (in \textbf{Ours - LVI}, \textbf{Ours - LRI}, or \textbf{Ours - LRVI}) the proposed multi-modal fusion demonstrates robust performance. Some of the aforementioned results are visualized in \cref{fig:eval_slam_tunnels}, alongside the previous tunnel experiment. Note here the sparsity of the instantaneous radar point cloud in this particular environment.

\paragraph{Frozen Lake}
The \ac{ar1} aerial robot was manually piloted on top of a frozen lake, starting from close to a bank, flying out into the middle, and returning close to the starting location. The ground truth was generated using the Pix4DMatic bundle adjustment result. Once the robot has traveled further than the range of the LiDAR away from any bank of the lake, the point cloud is geometrically self-similar and resembles a large plane. Notably, the LiDAR point cloud is also affected by the ice such that rays with a large incidence angle on the ice return invalid points and valid points have increased noise standard deviation. The radar after takeoff and before landing mostly returns less than three points per point cloud with frequently empty point clouds. The returns are primarily from the surface directly below the robot as it is flying. 

GaRLIO fails in this dataset as it calculates a least-squares estimate of velocity from the radar point cloud for outlier rejection, which requires a minimum of three points. AF-RLIO while functioning initially, quickly breaks once far enough away from the bank due to the sparse radar point clouds. Both FAST-LIO2 and FAST-LIVO2 initially function well, however, their accuracy deteriorates rapidly when the system performs an aggressive yaw maneuver and accumulates significant error. Despite also based on LiDAR-inertial fusion, \textbf{Ours - LI} does not fail in this environment due to parametric differences in \textbf{Ours - LI}, that were not replicable in FAST-LIO2, which results in more robust performance. 
Due to the visually feature-full environment and good outdoor lighting conditions, OpenVINS provides the best baseline performance (followed by ROVIO). Our solution retains performance similar to these baselines across most vision-involving configurations (i.e., \textbf{Ours - VI}, \textbf{Ours - LVI}, \textbf{Ours - RVI}, and \textbf{Ours - LRVI}), while it is notable that \textbf{Ours - LI}, \textbf{Ours - RI}, and \textbf{Ours - LRI} remain functional. The performance of the proposed method's ablation is shown in \cref{fig:eval_slam_lake}, where the geometric self-similarity of the frozen lake is clear.

\paragraph{Campus Fog}
A handheld UniPilot module is carried through a typical university campus environment, for a total trajectory length of \SI{670}{\meter}. The trajectory starts outdoors but proceeds indoors into a fog-filled room before returning outdoors. As a result, the experiment features large variation in the local environment scale as well as dense visual obscurants, which are known to cause problems for LiDAR- and vision-based estimation. The ground truth for this experiment is created using the Pix4DMatic bundle adjustment-based method. 

As expected, both the vision- and LiDAR-based methods (i.e., FAST-LIO2, FAST-LIVO2, ROVIO, OpenVINS, \textbf{Ours - LI}, \textbf{Ours - VI}, and \textbf{Ours - LVI}) perform well until the room with visual obscurants (fog), wherein the aforementioned methods accumulate significant error resulting from the extended duration of unusable measurements. The \textbf{Ours - RI} ablation functions regardless as the radar is unaffected by such phenomena, however still accumulates drift due to the long mission duration. GaRLIO and AF-RLIO, despite including the radar, are challenged here as well. For the former, the sparseness of the radar point clouds result in challenges with respect to \ac{ransac}-based outlier rejection, and as a result the method fails. For the latter, when LiDAR point cloud degeneracy is detected, the method switches to radar-based registration. This is generally difficult with small form-factor radar sensors, again leading to failure. This also indicates the strengths and inherent robustness of our radar Doppler factor formulation. The proposed multi-modal ablations which include the radar (\textbf{Ours - RI}, \textbf{Ours - RVI}, \textbf{Ours - LRI}, and \textbf{LRVI}) perform robustly, as the radar's invariance to visual obscurants is able to be complementarily fused with the accuracy associated with vision- and LiDAR-based methods in suitable conditions. As expected, the ablations which include the LiDAR together with the radar perform best. The results for this experiment are visualized in \cref{fig:eval_slam_campus_fog}, note in particular the challenge posed on LiDAR- and vision-based sensing by the visual obscurants in the fog-filled room. Here, the LiDAR returns nearly no points and the camera is completely blinded.

\subsubsection{Scene Reasoning}\label{subsec:eval_vlm}

Provided the multi-modal \ac{slam} capabilities of the \ac{ua}, we further assess downstream functionality for scene reasoning. \cref{fig:scene_reasoning} presents two real-world examples of the proposed \ac{vlm} reasoning system. As a straightforward extension to the core \perception~capabilities, our open-vocabulary semantic mapping system and 3D object detection are shown on the left side of the figure, where the reconstructed point cloud, together with the bounding boxes of the detected objects, is illustrated. Our semantic mapping pipeline, based on the open-vocabulary object detector, operates at 1 Hz. This result demonstrates the ability of the \perception~to consistently maintain semantic understanding over time. 

On the right side of~\cref{fig:scene_reasoning}, we show instances of the visual question/answering module, which provides contextual reasoning beyond geometric perception. As shown in the examples, the system correctly identifies potentially unsafe conditions in fog with high confidence, as well as determines whether there are exits or entrances in the current view. In our current implementation, GPT-5 is queried through the OpenAI API every 50 s. The end-to-end inference latency across our evaluations was $5.66 \pm 1.55$ s, including both the API latency and the model inference time. Overall, these results demonstrate that the combination of semantic 3D mapping and binary visual Q\&A can enable more robust scene understanding and reasoning. 

\vspace{-1ex}

\begin{figure*}[h]
	\centering
	\includegraphics[clip, trim = 0cm 0cm 0cm 0cm, width=1\linewidth]{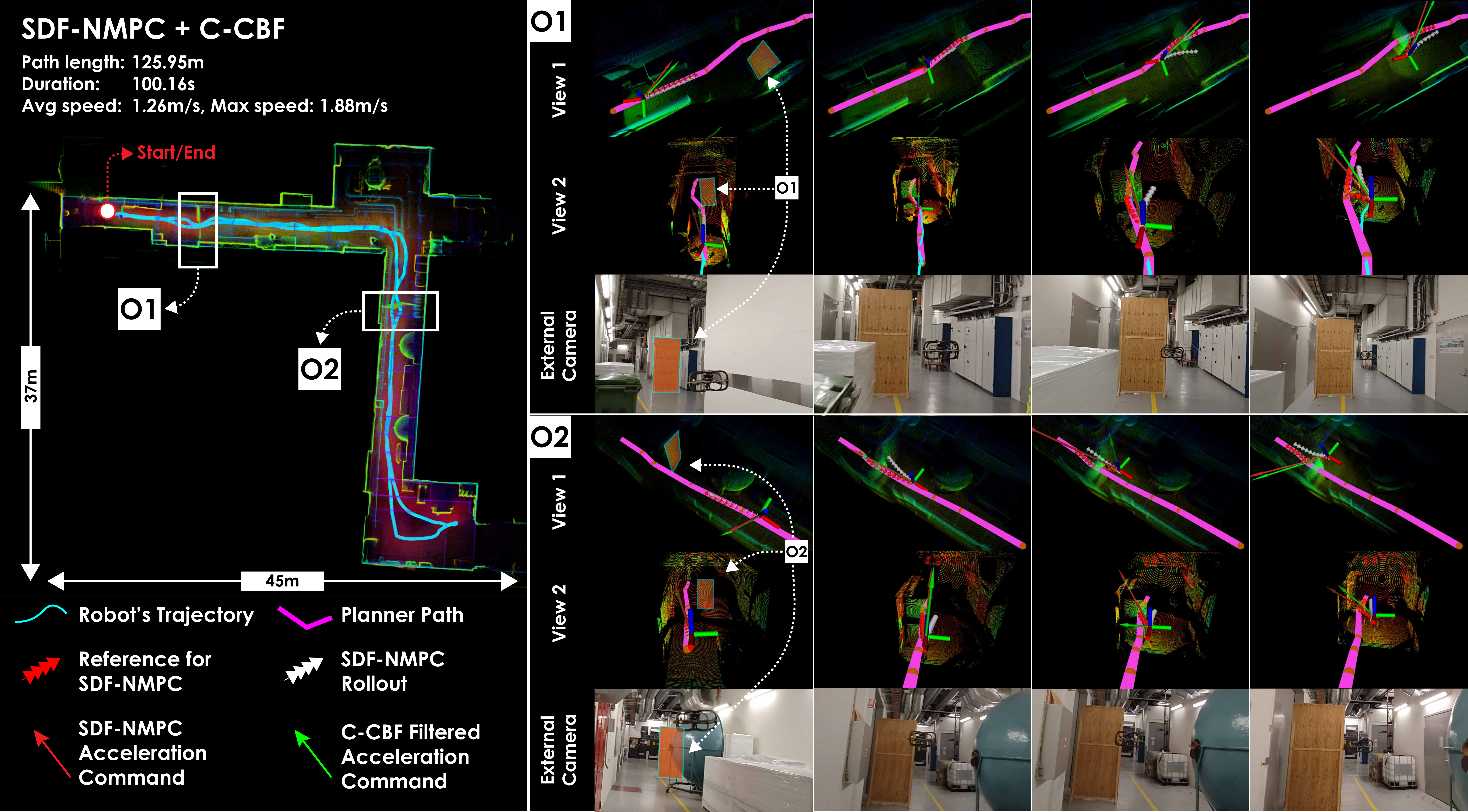}
	\caption{Navigation module evaluation is performed in the campus using \ac{ar2} with \ac{neuralmpc} + \ac{ccbf}. The robot was tasked to explore the basement autonomously and return home. At two instances, an obstacle was moved into the robot's planned path during the exploration and the return phases respectively, forcing the \ac{neuralmpc} + \ac{ccbf} to avoid these previously-absent obstacles in the robot's path. The two specific instances in the mission are highlighted to demonstrate reactive collision-avoidance. The full map is shown alongside the specific instances, with the regions of the specific obstacles highlighted.}
	\label{fig:eval_safety_basement_nmpc}
\end{figure*}

\begin{figure*}[h]
	\centering
	\includegraphics[clip, trim = 0cm 0cm 0cm 0cm, width=1\linewidth]{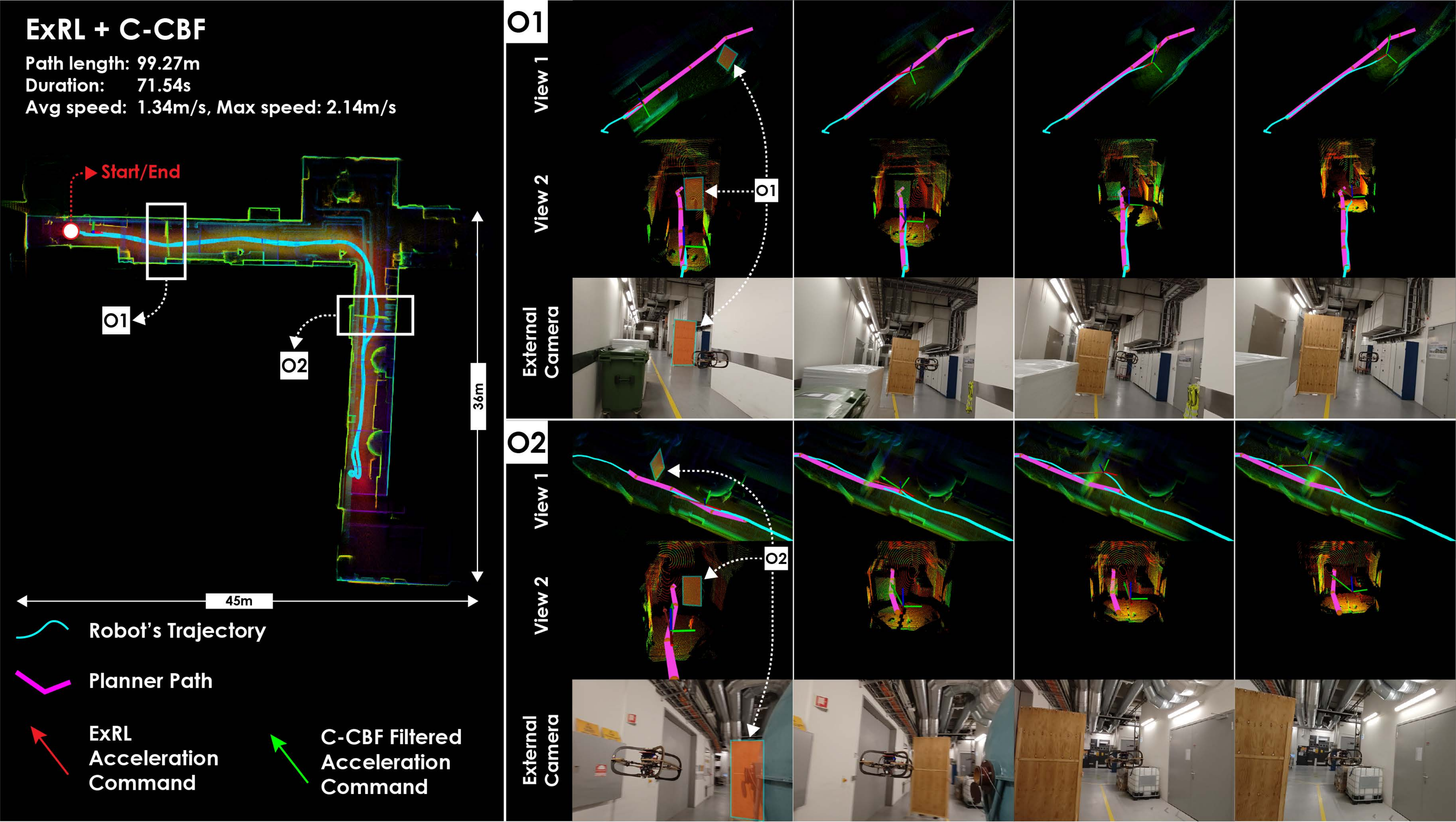}
	\caption{Navigation module evaluation is performed in the campus using \ac{ar2} with \ac{exteroceptivedrl} + \ac{ccbf}. The robot was tasked to explore the basement autonomously and return home. At two instances, an obstacle was moved into the robot's planned path during the exploration and the return phases respectively, forcing the \ac{exteroceptivedrl} + \ac{ccbf} to avoid these previously-absent obstacles in the robot's path. The two specific instances in the mission are highlighted to demonstrated reactive collision-avoidance. The full map is shown alongside the specific instances, with the regions of the specific obstacles highlighted.}
	\label{fig:eval_safety_basement_rl}
\end{figure*}

\subsection{Navigation Module Evaluation}\label{subsec:eval_navigation}
We conduct experiments to thoroughly evaluate the \navigation{}. Specifically, two studies are conducted. The first, evaluates the ability of the \navigation{} to navigate to a waypoint without the presence of a guiding path from the \planning{}. In the second, the ability of the \navigation{} to handle the sudden appearance of unmapped obstacles in the planned path is studied. The aim of these experiments is to a) evaluate the performance of the \navigation{} in the \acs{neuralmpc} + \ac{ccbf}, \acs{exteroceptivedrl} + \ac{ccbf}, and unsafe controller + \ac{ccbf} (where applicable) configurations, b) contrast the behaviors of these collision avoidance methods, and c) evaluate the benefits of the multi-layered safety approach. In all these missions, the \ac{ar2} platform was used, with the \ac{slam} module running the graph optimization online after receiving each exteroceptive measurement, considering radar measurements at \SI{10}{\hertz} and LiDAR measurements at \SI{10}{\hertz}. 
In each experiment (including the evaluations in~\cref{subsec:eval_fullstack}) the \ac{neuralmpc} runs at $\SI{40}{\hertz}$, \ac{exteroceptivedrl} at $\SI{30}{\hertz}$, and \ac{ccbf} runs at $\SI{50}{\hertz}$. The last received exteroceptive sensor measurements and state estimates are used to populate the state and inputs for each method while it executes at the desired frequency.

\begin{figure*}[h]
	\centering
	\includegraphics[clip, trim = 0cm 0cm 0cm 0cm, width=1\linewidth]{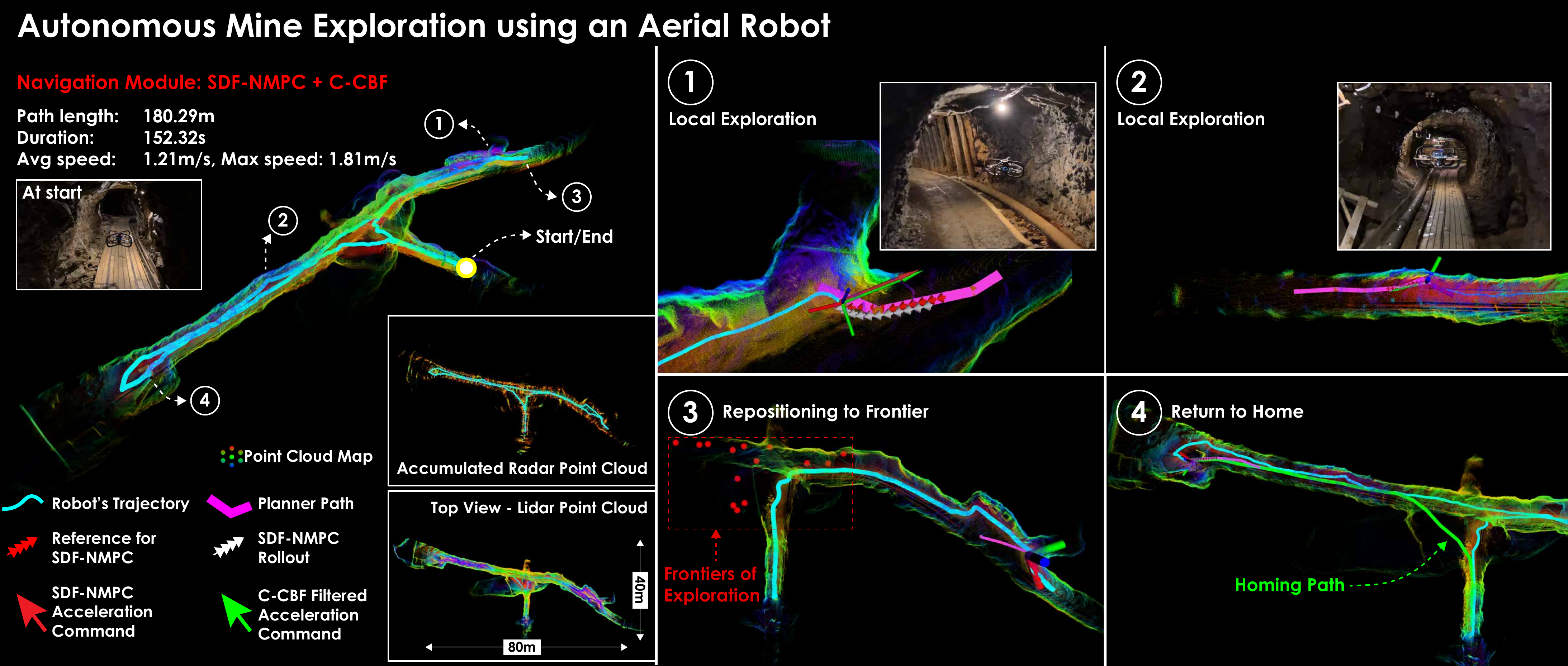}
	\caption{Full-stack evaluation in a multi-branched section of an underground mine using \ac{ar2} with \ac{neuralmpc} as the navigation method. The robot started in one branch and explored all three branches, repositioning when needed. The \ac{neuralmpc} tracks the planned path closely, resulting in next to zero interventions from the \ac{neuralmpc} or \ac{ccbf}. The figure shows the full map of the environment along with key instances in the mission. Additionally, the mission statistics are shown at the top.}
	\label{fig:eval_aerial_mine_nmpc}
\end{figure*}

\paragraph{Navigation to Waypoint}
The first evaluation studies the ability of the \navigation{} to navigate to a waypoint without a guiding path from the map-based \planning{}. The study is conducted in a forest environment. Three experiments are conducted in which the collision avoidance methods used are \ac{neuralmpc} + \ac{ccbf}, \ac{exteroceptivedrl} + \ac{ccbf}, and an unsafe policy (\ac{neuralmpc} but with its collision-avoidance constraints disabled) + \ac{ccbf}. In each experiment, the robot starts from the same location, and the same waypoint is given to the \navigation{}. The results of the study are shown in Figure~\ref{fig:eval_forest_safety}. Both configurations, \ac{neuralmpc} + \ac{ccbf} and \ac{exteroceptivedrl} + \ac{ccbf}, are able to reach the waypoint, avoiding the obstacles. The collision avoidance is predominantly carried out by the \ac{neuralmpc} or \ac{exteroceptivedrl}, with the \ac{ccbf} intervening only in a few instances (Figures~\ref{fig:eval_forest_safety}.1.3, \ref{fig:eval_forest_safety}.2.1, and \ref{fig:eval_forest_safety}.2.2). Engagement of the \ac{ccbf} is not a proof that \ac{neuralmpc} or \ac{exteroceptivedrl} would necessarily lead to a collision but indicates that the tuning of this last-resort method was such that it triggers it to adjust the reference commands. In turn, close evaluation of the few instances when \ac{ccbf} was engaged in these experiments indicates that the robot centroid was on average \SI{0.65}{\meter} from the obstacles (which given the robot dimensions, entails less than \SI{0.5}{\meter} clearance). On the other hand, the combination of the unsafe version of \ac{neuralmpc} with the \ac{ccbf} (which then becomes the only obstacle-avoidance mechanism) is not able to reach the goal and gets stuck, but remains safe at all times. As the \ac{ccbf} only aims to remain in the safe set, navigating to the goal is not per se an objective for this method. It is noted that the third configuration (unsafe controller + \ac{ccbf}) is not a recommended configuration of the \ac{ua} and is only evaluated to present the different roles of the navigation layers.

Qualitatively, this study further shows the clear difference in the robot's behavior when using \ac{neuralmpc} vs \acs{exteroceptivedrl}. The \ac{neuralmpc} follows the straight line from the start to the goal more closely, while the \acs{exteroceptivedrl} policy deviates significantly. On the other hand, the \acs{exteroceptivedrl} achieves higher speeds throughout the trajectory than \ac{neuralmpc}. Due to this, the resulting mission time for both is comparable. Hence, the selection between \ac{neuralmpc} and \acs{exteroceptivedrl} depends on the requirements of the task. The different behaviors manifested between \ac{neuralmpc} and \acs{exteroceptivedrl} is among the reasons why the release of the \ac{ua} contains both methods.

\paragraph{Moving Obstacles}
The second study conducted to evaluate the \navigation{} aims to evaluate its performance in the presence of obstacles appearing in the planned path. Specifically, the following scenario was constructed. The robot was tasked to explore a section of a university building at NTNU. In two separate instances, after the \planning{} plans a path based on the online map, an obstacle is placed to block this path. The planner is not re-triggered and thus the reference path shall be in collision. The ability of the \navigation{} to handle this scenario is tested. 
Two experiments are conducted with the configurations \ac{neuralmpc} + \ac{ccbf} and \acs{exteroceptivedrl} + \ac{ccbf}. Figures~\ref{fig:eval_safety_basement_nmpc} and \ref{fig:eval_safety_basement_rl} show the result of the respective experiments. As can be seen, both policies are able to successfully avoid the unseen obstacle. The figures also show that the \ac{neuralmpc} tends to avoid the obstacle with less, yet sufficiently safe, clearance as compared to \acs{exteroceptivedrl} as its formulation requires it to have minimal deviation from the reference path. \acs{exteroceptivedrl} does not have this constraint and only aims to reach the end of the planned path safely.

\begin{figure*}[h]
	\centering
	\includegraphics[clip, trim = 0cm 0cm 0cm 0cm, width=1\linewidth]{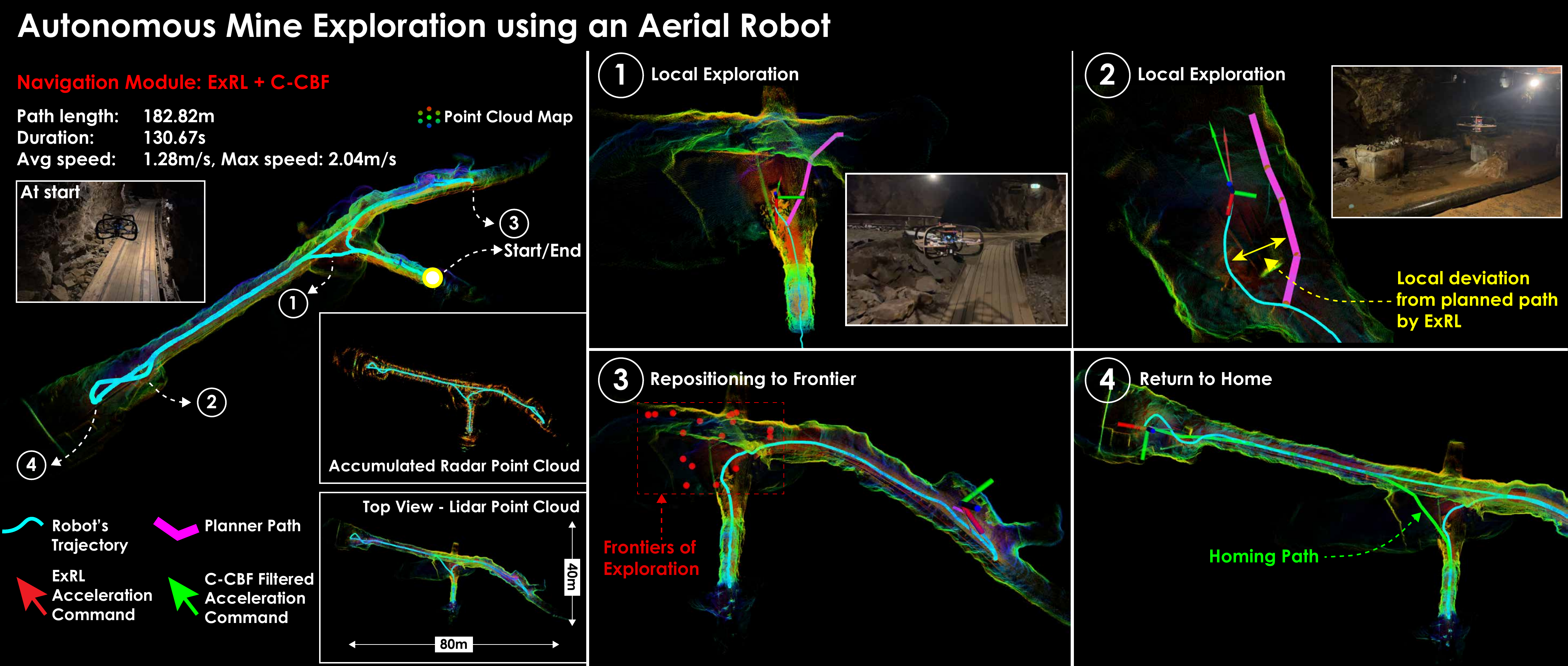}
	\caption{Full-stack evaluation in a multi-branched section of an underground mine using \ac{ar2} with \ac{exteroceptivedrl} as the navigation method. The robot started in one branch and explored all three branches, repositioning when needed. The \ac{exteroceptivedrl} policy only aims to reach the end of each planned path, presenting larger local deviations, but remaining safe at all times. The figure shows the full map of the environment along with key instances in the mission. Additionally, the mission statistics are shown at the top.}
	\label{fig:eval_aerial_mine_rl}
\end{figure*}

\subsection{Evaluation of the Full Stack}\label{subsec:eval_fullstack}

The full \ac{ua} is evaluated using both aerial and ground robots. When evaluating the full stack, we examine the result of the coordinated interaction between the \texttt{perception}, \texttt{planning} and \texttt{navigation modules}. The \perception{} is the foundation of the demonstrated autonomy, the \planning{} drives the behaviors manifested by the robots, while the \navigation{} provides control and reinforces safety for the autonomous systems. Specifically, the Exploration and Inspection objectives are tested with the Planning to Target being implicitly evaluated through these. Additional insights regarding the Planning to Target behavior can be found in~\cite{zacharia2026omniplannerarxiv}. Similarly to \cref{subsec:eval_navigation} the \ac{slam} module graph optimization is calculated online, with update rate matching what was previously described.

\subsubsection{Evaluations with an Aerial Robot}\label{subsec:eval_fullstack_aerial}

\paragraph{} We evaluate the \ac{ua} on the \ac{ar2} in three distinct environments namely a) an underground mine, b) a forest, and c) a ship cargo hold.

\paragraph{Underground Mine}
The first experiment is conducted in a section of the L{\o}kken mine in Norway. We demonstrate results both when otherwise using the \ac{neuralmpc} and \ac{exteroceptivedrl}. The selected section of the mine is a $3$-way intersection with the robot starting at one end of the narrowest branch (\SI{1.5}{\meter} wide). The mine has low lighting conditions, and due to the dome \ac{fov}, the LiDAR data can quickly become degenerate in narrow branches if the sensor is facing a wall. In both missions, the robot explored the first branch it started in, continued to one of the other branches, repositioned to the next upon exploring it, before finally returning to the start location when the allotted area was fully explored. 
The robot successfully explored the environment while remaining safe at all times. Figures~\ref{fig:eval_aerial_mine_nmpc} and \ref{fig:eval_aerial_mine_rl} show the maps and planning instances in the missions corresponding to \ac{neuralmpc} and \ac{exteroceptivedrl} respectively. As the environment topology does not allow much deviation from the planned path, the total path length in both missions is comparable, however, the \ac{exteroceptivedrl} policy finishes faster reaching higher speeds.
It was observed that at some instances in the narrowest part near the starting area (marked as Start/End in the figures), the \ac{neuralmpc}/\ac{exteroceptivedrl} and the \acs{ccbf} objectives are competing resulting in transient oscillations. The \ac{neuralmpc}/\ac{exteroceptivedrl} are tasked to both make progress along the planned path and maintain safety, while the \acs{ccbf} only aims for robot safety thus leading to competing actions in situations that have tight safety margins as here. Nevertheless, the system was always able to continue and this behavior was short-lived.

\begin{figure*}[h]
	\centering
	\includegraphics[clip, trim = 0cm 0cm 0cm 0cm, width=0.98\linewidth]{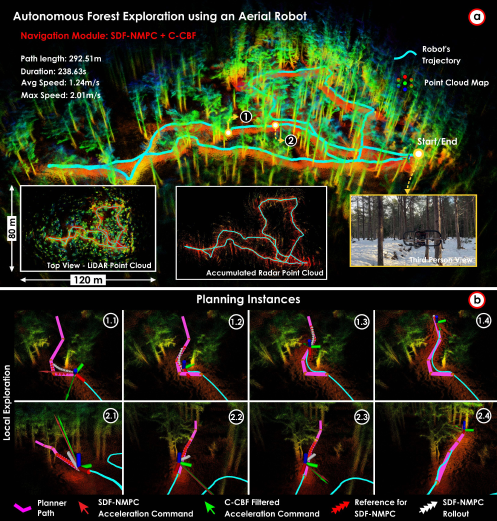}
	\caption{Full-stack evaluation in the forest using \ac{ar2} with \ac{neuralmpc} as the navigation method. The robot was tasked to explore a given area autonomously and return home. The figure shows the full map and planning instances in the mission. As the \ac{neuralmpc} follows the planned path closely, it needs to intervene infrequently.}
	\label{fig:eval_aerial_forest_nmpc}
\end{figure*}

\begin{figure*}[h]
	\centering
	\includegraphics[clip, trim = 0cm 0cm 0cm 0cm, width=0.98\linewidth]{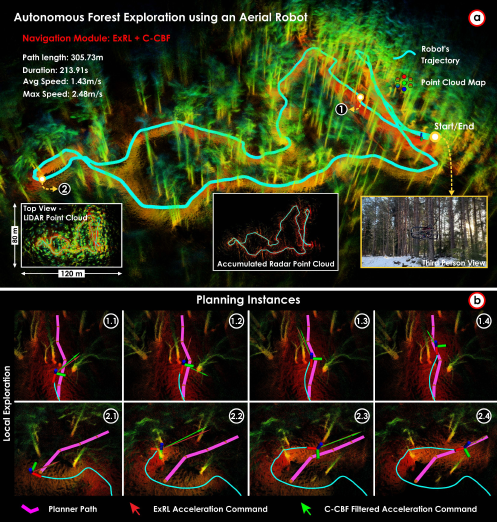}
	\caption{Full-stack evaluation in the forest using \ac{ar2} with \ac{exteroceptivedrl} as the navigation method. The robot was tasked to explore a given area autonomously and return home. The figure shows the full map and planning instances in the mission. As can be seen, the \ac{exteroceptivedrl} is not formulated to follow the planned path strictly, but successfully navigates towards the end of the path, avoiding obstacles.}
	\label{fig:eval_aerial_forest_rl}
\end{figure*}

\paragraph{Forest}
The \ac{ar2} was deployed in a forest in Trondheim, Norway to explore an area of \qtyproduct{120x80}{\meter} with height limited to $\SI{2.5}{\meter}$. The forest area contains trees at varying densities with thin branches and foliage in some places. The ground was covered in snow at the time of testing. 
In this test, two missions were conducted, one each with the \ac{neuralmpc} and the \ac{exteroceptivedrl} as the core navigation policy feeding into the \acs{ccbf}. Figures~\ref{fig:eval_aerial_forest_nmpc} and \ref{fig:eval_aerial_forest_rl} show the results for the respective missions. In both missions, the robot started at the same location with identical mission and robot parameters. The robot first performed exploration of the given space and returned back to the start location. As can be seen from Figures~\ref{fig:eval_aerial_forest_nmpc} and \ref{fig:eval_aerial_forest_rl}, the \ac{neuralmpc} is designed to follow the path given by the \planning{} more accurately than \ac{exteroceptivedrl}. Hence, the robot can take longer trajectories to reach the end of the same path when \ac{exteroceptivedrl} is used as compared to \ac{neuralmpc}. However, as shown by the average and max speed, the \ac{exteroceptivedrl} policy generates smoother and faster trajectories than \ac{neuralmpc} (partially as a result of the non-smooth paths given by the \planning{}), thus resulting in similar mission times. It is thus a decision point for the user of the \ac{ua} to select among these two core navigation policies with the \ac{neuralmpc} being a very reasonable choice when following the planner plans closely is desired, while \ac{exteroceptivedrl} is particularly relevant when a more loose tracking of these references combined with agile maneuvering is preferred. Figures~\ref{fig:eval_aerial_forest_nmpc} and \ref{fig:eval_aerial_forest_rl} show the full map and planning instances from the respective missions, along with the robot in the environment. In both missions, the robot was successfully able to explore the allotted area, avoiding collisions even in the presence of thin obstacles due to the multi-layered safety. Figures~\ref{fig:eval_aerial_forest_nmpc}.2.1-\ref{fig:eval_aerial_forest_nmpc}.2.4 show one such instance where the \ac{neuralmpc} deviates from the path planned by the \planning{}. Similarly, the \ac{exteroceptivedrl} policy successfully guides the robot to the end of the path. At one instance in the mission, shown in Figure~\ref{fig:eval_aerial_forest_rl}.1.2, the \acs{ccbf} can be seen intervening and correcting the command of the \ac{exteroceptivedrl} policy as the robot passes through a narrow opening, thus highlighting the importance of the multi-layered safety approach. As when the \navigation{} was evaluated separately, it is worth mentioning that when the \ac{ccbf} was engaged the distance of the robot centroid from the obstacles was \SI{0.68}{\meter} and the component of the velocity towards the obstacle was~\SI{1.10}{\meter\per\second}. Although the engagement of the \ac{ccbf} does not strictly imply that the core method would lead to a collision, it indicates the role of such a last-resort safety method to assure what proximity and maneuvering towards the obstacles is considered as acceptable.

\begin{figure*}[h]
	\centering
	\includegraphics[clip, trim = 0cm 0cm 0cm 0cm, width=1\linewidth]{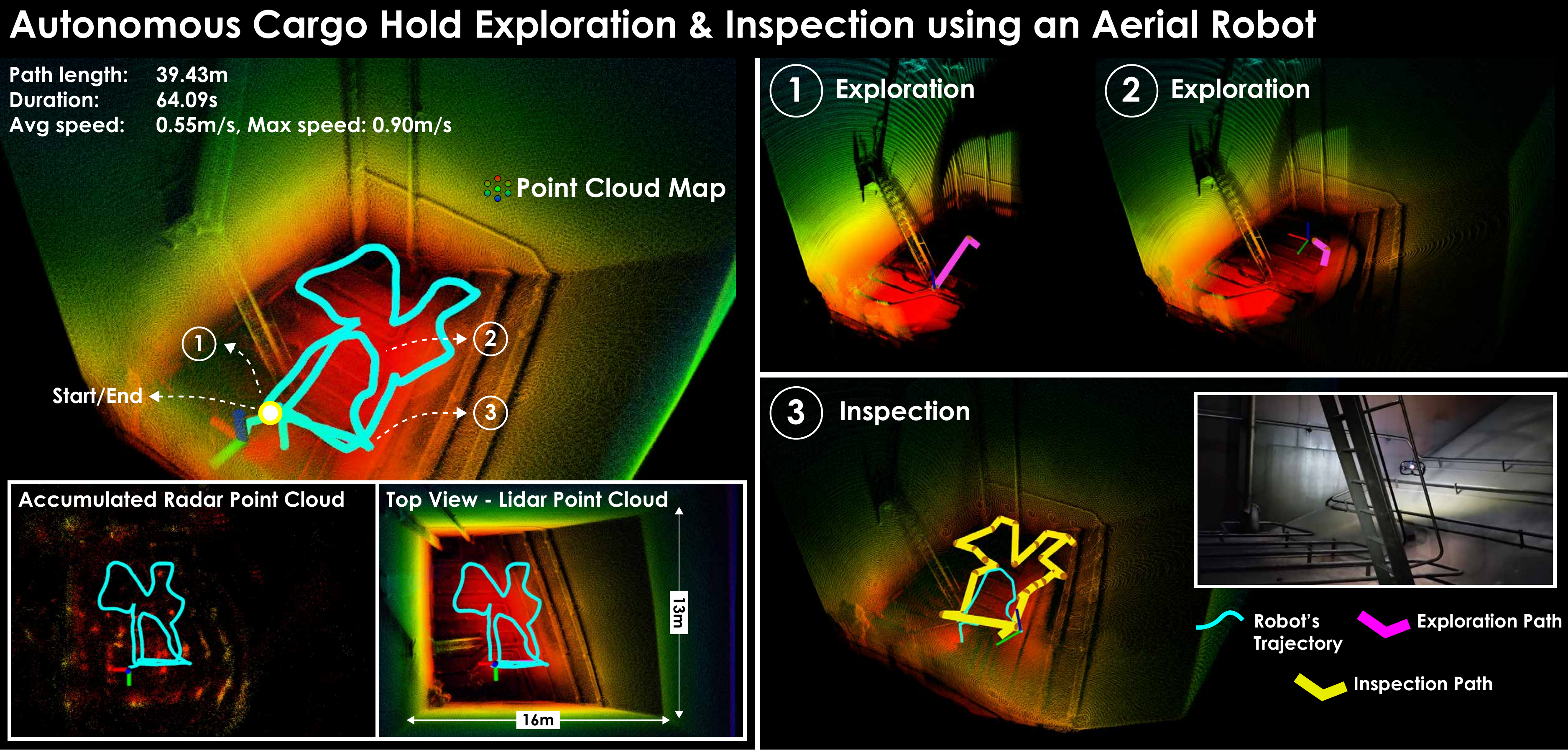}
	\caption{Deployment of \ac{ar2} in the ship cargo hold. The robot started with no prior information of the environment. It first performed exploration to map the tank. Upon completion, it switched to the inspection behavior, where it viewed the mapped surfaces with the camera sensor at the desired viewing distance.}
	\label{fig:eval_aerial_cargo}
\end{figure*}

\paragraph{Ship Cargo Hold}
In the third experiment, the \ac{ar2} was deployed in a cargo hold of an oil tanker ship. In contrast to the previous missions, here the inspection behavior of the \planning{} is engaged. The dimensions of the cargo hold were \qtyproduct{16x13x15}{\meter}, however, the mission height was limited to \SI{3}{\meter} for safety considerations. The robot was tasked to explore the cargo hold and inspect the mapped surfaces. The robot started inside the cargo hold with no prior knowledge, performed exploration, and upon completion switched to the inspection behavior. The complete map and instances of the mission along with an image of the robot in the environment is shown in Figure~\ref{fig:eval_aerial_cargo}. As an exception, it is noted that in this experiment the safety policies in the \navigation{} were disabled as (a) the environment is not demanding in terms of collision avoidance (one large room with no obstacles inside), and (b) this allows the robot to more flexibly travel outside the depth sensor's \ac{fov} in the inspection phase.

Through these experiments, we demonstrate the importance and the role of the three layers of safety namely a) map-based collision-avoidance, b) depth-driven \ac{neuralmpc} or \ac{exteroceptivedrl} policies, and c) the last resort \ac{ccbf}. The map-based safety allows longer horizon planning enabling more complex behavior, and is the safety layer doing the majority of the collision-avoidance throughout the missions. The depth-driven navigation policies add an additional safeguard against challenges to map-based safety as documented in this work. Finally, the \ac{ccbf} provides formal safety guarantees ensuring that the robot remains safe at all times. 

\subsubsection{Legged}\label{subsec:eval_fullstack_ground}
\paragraph{} To demonstrate the performance of the \ac{ua} on ground robots, we deployed \ac{gr1} in two distinct settings a) in an underground mine, and b) inside a university building at NTNU.

\begin{figure*}[h]
	\centering
	\includegraphics[clip, trim = 0cm 0cm 0cm 0cm, width=0.99\linewidth]{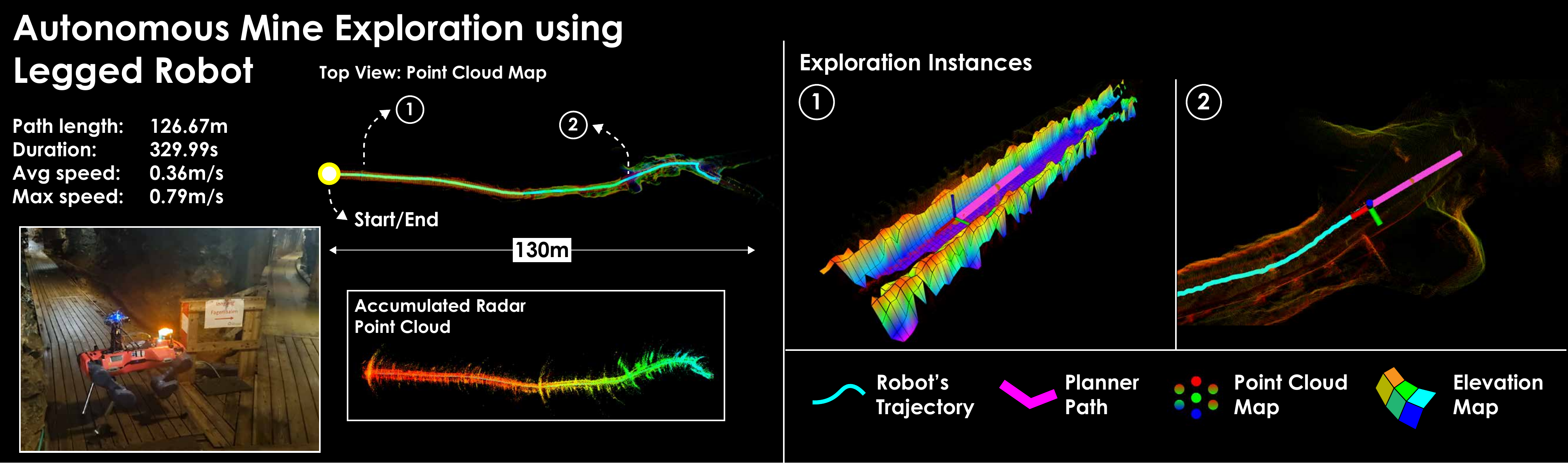}
	\caption{Full-stack evaluation in the L\o{}kken underground mine using the \ac{gr1} legged robot. The mission was conducted in one of the mine shafts, presenting narrow cross-section at times, and gaps on the side. Due to the dual map representation (volumetric and elevation maps), the robot was successfully able to handle these challenges completing the mission. }
	\label{fig:eval_legged_mine}
\end{figure*}

\paragraph{Underground Mine}
In the first mission, \ac{gr1} was deployed in another section of the L{\o}kken mine. This section consisted of one mine shaft having narrow passages and areas with gaps on the side, requiring careful planning and locomotion. Figure~\ref{fig:eval_legged_mine} shows the map and planning instances of the mission. Using the dual map representation (volumetric and elevation map), the \ac{ua} is able to successfully complete the mission. 
It is noted that here the additional safety layers of the \navigation{} are not utilized as the commercial ANYmal robot already provides the partially analogous feature of ``perceptive locomotion'' fusing short-range depth from its all-around depth cameras for traversability-aware near-term navigation.

\begin{figure*}[h]
	\centering
	\includegraphics[clip, trim = 0cm 0cm 0cm 0cm, width=1\linewidth]{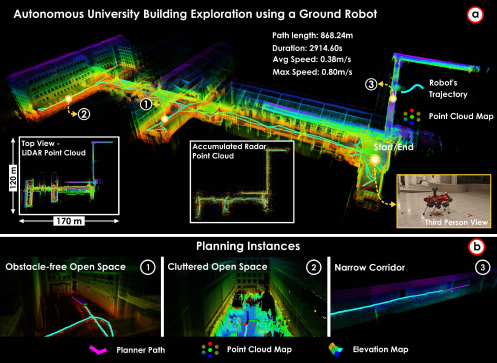}
	\caption{Full-stack evaluation was conducted in a university building using the \ac{gr1} legged robot. The robot was tasked with exploring the entire ground floor, which featured both open spaces and narrow corridors. The figure shows the complete map along with the planning instances from the mission.}
	\label{fig:eval_legged_campus}
\end{figure*}

\paragraph{University Building}
The second mission was conducted inside a building at NTNU. The building consists of two sections a) a large open hall with side offshoots, and b) a section with a network of narrow (width \SI{<1.5}{\meter}) corridors, as can be seen in Figure~\ref{fig:eval_legged_campus}. The robot started in the open hall, and explored it along with the offshoots. Upon completion, the robot repositioned towards the narrow corridor section, explored those, and returned to the start location. This environment presents several challenges to the entire stack, including a) large scale, b) branching corridors, and c) varying environment size. However, the multi-modal \ac{slam} solution provided resilient odometry and consistent maps throughout the mission, as well as the bifurcated architecture of the \planning{} with traversability-aware planning (exploiting both the volumetric and the elevation maps) lead to successful mission completion.

\section{Conclusion \& Future Work}
\label{sec:conclusion}

The \ac{ua} is openly released with the aim of serving as a foundation for a common autonomy blueprint across diverse robot configurations operating in the air, on land, and at sea. Currently, the \ac{ua} supports a wide variety of aerial and ground robot morphologies and enables resilient \ac{gnss}-denied, perceptually-degraded localization, mapping, and scene reasoning within target reach, exploration and inspection missions with a key focus on assured safety through multi-layered navigation. Extensive field evaluation results, alongside openly released datasets, allow for its comprehensive evaluation.

We seek to collaborate with the research community towards enhancing the reliability and resilient performance of the stack, alongside its extension to different robot morphologies and the incorporation of new behaviors. Future development plans specifically include (a) the support of further morphologies, including highly non-holonomic platforms such as fixed-wing uncrewed aerial vehicles, (b) increased emphasis on navigation within dynamic environments, (c) development of further object-centric behaviors, especially guided by natural language, (d) the fusion of additional modalities and specifically infrared vision, (e) improving the vision fusion into MIMOSA-X to be more tightly coupled analogous to what is already done for LiDAR and radar, (f) enhancing the \ac{exteroceptivedrl} toward improved long-horizon capabilities, and (g) extend multi-layered safety with traversability-aware reactive modules for ground systems.

Last but not least, we aim to document how lessons learned from the deployment of the stack can lead to certain improvements and adaptations. This currently includes investigations for (a) how to best handle the trade-off between \ac{ccbf} and the exteroceptive \ac{neuralmpc} and \ac{exteroceptivedrl} methods, (b) computationally-efficient ways to directly fuse vision features in the \ac{slam} solution of the \ac{ua}, alongside (c) refining the rewards of \ac{exteroceptivedrl} to better balance between the ability of the method to negotiate complex environments, and how energetically-efficient the trajectories are.

\begin{acks}
We would like to acknowledge Statens Vegvesen for enabling us to perform tests in Runehamar, Vestland Fylkeskommune for allowing us to perform experiments in the Fyllingsdal sykkeltunnel, Leica Geosystems for providing the \ac{ros} compatible MS60 and AP20 setup for collection of ground truth, as well as Orkla Industrimuseum for facilitating the tests in the L{\o}kken Mine.
\end{acks}

\section*{Author contributions}
\begin{itemize}
    \item Mihir Dharmadhikari: Contributed in the formulation of the idea, the \ac{ua} architecture, and all the evaluations. Furthermore, he is the core developer of the \planning{}.
    \item Nikhil Khedekar: Contributed in the formulation of the idea, the \ac{ua} architecture, and all the evaluations. Furthermore, he is a core developer of the multi-modal \ac{slam}, specifically the LiDAR and Vision modalities.
    \item Mihir Kulkarni: Contributed in the formulation of the idea, the \ac{ua} architecture, and all the evaluations. Furthermore, he is the core developer of Exteroceptive Deep RL navigation policy.
    \item Morten Nissov: Contributed in the formulation of the idea, the \ac{ua} architecture, and all the evaluations. Furthermore, he is a core developer of the multi-modal \ac{slam}, specifically the Radar modality.
    \item Martin Jacquet: He is the core developer of the Neural SDF-NMPC and co-developer of the Composite CBF-based Safety Filter, and contributed towards their integration in the \ac{ua}.
    \item Angelos Zacharia: He is the co-developer of the \planning{}. He contributed towards its integration in the \ac{ua} and the legged robot experiments.
    \item Marvin Harms: He is the core developer of the Composite CBF-based Safety Filter. He contributed towards its integration in the \ac{ua} and the evaluations of the safety policies.
    \item Albert Gassol Puigjaner: He is the core developer of the VLM-based reasoning part of the \ac{ua} and contributed towards its integration in the \ac{ua}.
    \item Philipp Weiss: Contributed towards the development of the hardware setup and conducting the field experiments.
    \item Kostas Alexis: Contributed in the formulation of the idea, the \ac{ua} architecture, and planning for all evaluations. Furthermore, he contributed to the planning, problem formulation, and algorithmic approach of each module in the \ac{ua}.
\end{itemize}
All authors contributed to the writing of this manuscript. 

\section*{Statements and declarations}
\subsection*{Ethical considerations}
This article does not contain any studies with human or animal participants.
\subsection*{Consent to participate}
Not applicable.
\subsection*{Consent for publication}
Not applicable.

\begin{dci}
The author(s) declared no potential conflicts of interest with respect to the research, authorship, and/or publication of this article.
\end{dci}

\begin{funding}
The author(s) disclosed receipt of the following financial support for the research, authorship, and/or publication of this article: This work was supported by European Commission Horizon Europe grant agreements a) SPEAR (EC 101119774), b) DIGIFOREST (EC 101070405), c) SYNERGISE (EC 101121321), and d) AUTOASSESS (EC 101120732).
\end{funding}

\bibliographystyle{SageH}
\bibliography{bibliography/Library,bibliography/ua_papers,bibliography/estimation,bibliography/verification_works,bibliography/autonomy_stacks,bibliography/planning, bibliography/vlms}

\clearpage
\appendix
\section{Index to multimedia Extensions}\label{app:multimedia}
\small
\begin{table}[h]
\centering
    \newlength{\descwidth}
    \setlength{\descwidth}{\dimexpr\textwidth - 4cm\relax}
    \begin{tabular}{ccp{\descwidth}}
        \toprule
            \textbf{Ext.} & \textbf{\makecell{Media \\ type}} & \textbf{Description} \\
        \midrule
            1 & Video & Filename: \path{00_unified_autonomy_stack_main_video.mp4}. This video provides an overview of this paper including a) the \ac{ua} architecture, b) the details of each core modules, \texttt{perception}, \texttt{planning}, and \texttt{navigation}, and c) videos of indicative field deployments presented in the paper. \\
            2 & Video & Filename: \path{01_full_stack_evaluation_aerial_robot.mp4}. This video shows the recording and visualizations of the field experiments evaluating the full stack on the aerial robot as presented in \cref{subsec:eval_fullstack_aerial}. The robot is deployed in an underground mine, a forest, and a ship cargo hold, evaluating the exploration and inspection behaviors. \\
            3 & Video & Filename: \path{02_full_stack_evaluation_ground_robot.mp4}. This video shows the recording and visualizations of the field experiments evaluating the full stack on the ground robot as presented in \cref{subsec:eval_fullstack_ground}. The robot is deployed in an underground mine and a university building, evaluating the exploration behavior in narrow and large scale environments. \\
            4 & Video & Filename: \path{03_navigation_evaluation.mp4}. This video shows the recording and visualizations of the field experiments evaluating the \navigation{} as presented in \cref{subsec:eval_navigation}. The navigation module is tested in a waypoint navigation task and in the presence of moving obstacles demonstrating the usefulness of the multi-layered safety approach.\\
            5 & Video & Filename: \path{04_slam_evaluation.mp4}. This video shows the recording and visualizations of the field experiments evaluating the multi-modal \ac{slam} component of the \perception{} as presented in \cref{subsec:multi-modal-slam}. The multi-modal \ac{slam} is evaluated in datasets including a) flying in self-similar and dimly lit tunnels, b) flying over a self-similar frozen lake, and c) handheld dataset in a university campus containing area filled with fog. \\
            6 & Video & Filename: \path{05_reasoning_evaluation.mp4}. This video shows the recording and visualizations of the field experiments evaluating the VLM-based reasoning component of the \perception{} as presented in \cref{subsec:eval_vlm}. The module is tested on two datasets a) the campus dataset from the \ac{slam} evaluation, and b) the ground robot mission in the university building. The video shows the open vocabulary object detection and binary Q\&A capabilities of the module.\\
        \bottomrule
    \end{tabular}
    \caption{Index of multimedia extensions.}
\label{tab:extensions}
\end{table}
\normalsize

\end{document}